\documentclass{article}

\usepackage{microtype}
\usepackage{graphicx}
\usepackage{subcaption} 
\usepackage{booktabs} 
\usepackage{placeins}

\usepackage{hyperref}

\usepackage[accepted]{icml2024}

\usepackage{amsmath}
\usepackage{amssymb}
\usepackage{mathtools}
\usepackage{amsthm}
\usepackage{multirow}

\usepackage[capitalize,noabbrev]{cleveref}

\theoremstyle{plain}

\theoremstyle{definition}

\theoremstyle{remark}

\usepackage[textsize=tiny]{todonotes}

\newcommand{\answerYes}[1][]{\textcolor{blue}{[Yes] #1}}

\newcommand{\answerNA}[1][]{\textcolor{gray}{[N/A] #1}}
\newcommand{\answerTODO}[1][]{\textcolor{red}{\bf [TODO]}}

\icmltitlerunning{Surprisingly Strong Performance Prediction with Neural Graph Features}

\begin{document}

\twocolumn[
\icmltitle{Surprisingly Strong Performance Prediction with Neural Graph Features}



\icmlsetsymbol{equal}{*}

\begin{icmlauthorlist}
\icmlauthor{Gabriela Kadlecová}{mff,ui}
\icmlauthor{Jovita Lukasik}{sieg}
\icmlauthor{Martin Pilát}{mff}
\icmlauthor{Petra Vidnerová}{ui}
\icmlauthor{Mahmoud Safari}{fr}
\icmlauthor{Roman Neruda}{ui}
\icmlauthor{Frank Hutter}{fr,el}
\end{icmlauthorlist}

\icmlaffiliation{mff}{Charles University, Faculty of Mathematics and Physics}
\icmlaffiliation{ui}{The Czech Academy of Sciences, Institute of Computer Science}
\icmlaffiliation{fr}{University of Freiburg}
\icmlaffiliation{sieg}{University of Siegen}
\icmlaffiliation{el}{ELLIS Institute Tübingen}

\icmlcorrespondingauthor{Gabriela Kadlecová}{gabi.kadlecova@outlook.com}

\icmlkeywords{Machine Learning, ICML}

\vskip 0.3in
]



\printAffiliationsAndNotice{}  

\begin{abstract}
Performance prediction has been a key part of the neural architecture search (NAS) process, allowing to speed up NAS algorithms by avoiding resource-consuming network training. Although many performance predictors correlate well with ground truth performance, they require training data in the form of trained networks.
Recently, zero-cost proxies have been proposed as an efficient method to estimate network performance without any training. However, they are still poorly understood, exhibit biases with network properties, and their performance is limited.
Inspired by the drawbacks of zero-cost proxies, we propose neural graph features (GRAF), simple to compute properties of architectural graphs. GRAF offers fast and interpretable performance prediction while outperforming zero-cost proxies and other common encodings. In combination with other zero-cost proxies, GRAF outperforms most existing performance predictors at a fraction of the cost.
\end{abstract}

\section{Introduction}
\label{intro}
With the rising popularity of deep learning with applications across different domains, finding a well-performing neural network architecture for a given application has become a key problem. The field of Neural Architecture Search (NAS) automatizes the process and has gained noticeable attention in the past years \citep{ElskenMH19, insight1000}.
Due to the high costs of neural network training, speedup techniques are a key component of the NAS process. One of these techniques is based on performance predictors -- models that help us to reduce the number of trained architectures by predicting their performance based on a small number of sampled and trained networks \citep{naslib-predictors}.
Since these prediction models still require some networks to be trained and in some cases even include an overhead when fitting on the sampled train set, other, even more efficient techniques were needed.
As an answer, \emph{zero-cost proxies} have been recently proposed as a variant that does not require any network training \citep{lightweight, TanakaKYG20, MellorTSC21}. For every network, these proxies need mostly a single mini-batch of data to compute a score that correlates with the true performance of the network after full training on the downstream task of interest.

However, the reason for the correlation is still underexplored. Furthermore, \citet{nbsuitezero} even discovered biases between these proxies and network properties like the number of skip-connections. 
On the large number of various tasks that a NAS method should be able to solve nowadays \citep{tnb101}, zero-cost proxies show inconsistent performances and most of these proxies even have a lower average correlation with the network performance than simple metrics, such as \texttt{flops} \citep{colin2022adeeperlook, nbsuitezero}. In addition, most performance prediction methods lack interpretability. This is in particular an important property that would eventually allow us to determine what architectural properties drive the performance for a given task.

To address these limitations of zero-cost proxies, we first examine in an in-depth analysis the reason behind the zero-cost proxy biases, showing that it is in some cases a direct dependency rather than an actual correlation. Inspired by that, we take advantage of these biases and propose in this paper \emph{neural graph features} -- simple to compute network properties like operation counts or node degrees. We show that using the proposed neural graph features as an input to several (simple) prediction models provides strong and interpretable performance prediction on several tasks, ranging from the common accuracy prediction to hardware metrics prediction and lastly to (multi-objective) robustness prediction.

Our main contributions are summarized as follows:
\begin{enumerate}
    \item We examine biases in zero-cost proxies (ZCP) and show that some of them directly depend on the number of convolutions. We also demonstrate that they are poor at distinguishing structurally similar networks.
    \item We propose \emph{neural graph features} (GRAF), interpretable features that outperform ZCP in accuracy prediction tasks using a random forest, and yield an even better performance when combined with ZCP.
    \item 
    Using GRAF's interpretability, we demonstrate that different tasks favor diverse network properties.
    \item We evaluate GRAF on tasks beyond accuracy prediction, and compare with different encodings and predictors. The combination of using ZCP and GRAF as prediction input outperforms most existing methods at a fraction of the compute. 
\end{enumerate}

We make the code of our contributions available publicly\footnote{\href{https://github.com/gabikadlecova/zc\_combine}{https://github.com/gabikadlecova/zc\_combine}}

\section{Related work}
\label{relatedwork}

Recent works have proposed different kinds of performance predictors, with the most popular being model-based methods, learning curve extrapolation methods, zero-cost methods, and weight-sharing methods. Using the library NASLib, a wide variety of predictors were evaluated with unified fit and query time settings across common benchmarks, as well as in search settings \citep{naslib-predictors}. In this work, we focus on model-based predictors and zero-cost proxies, as they are directly comparable to our method. 

Model-based predictors include tree-based models like random forest \citep{RFcite}, or graph neural networks like BRP-NAS \citep{brpnas}. An extended overview of model-based predictors is provided in the appendix (Section \ref{app-rel-work}).

Zero-cost proxies can be used as direct predictors of the target metric, or also as input to model-based predictors \citep{lightweight, nbsuitezero}. Notably, Multi-Predict has used zero-cost proxies with an MLP accuracy predictor that enables transfer to other search spaces \citep{multipredict}. Another work used a random forest predictor and zero-cost proxies to estimate robust accuracy of NAS-Bench201
\citep{robustness, lukasik2023evaluation}.

As for interpretable predictors, NASBOWL enables the extraction of so-called motifs -- network subgraphs that are found in high or low-performing networks \citep{nasbowl}. Other non-predictor works have derived a dependence of network performance on depth \citep{nofreelunch}, or that limiting search spaces to a specific subset of operations entails high-performing networks \citep{LOPES2023119695}.

Many of the well-performing predictors take several minutes to train for larger sample sizes -- a limitation in budget-restricted search settings, as it introduces a computational overhead. Another drawback is, with the notable exception of NASBOWL, a lack of interpretability. Even NASBOWL does not consider properties like network depth or operation count. In our work, we discover that these properties are highly influential on the prediction quality.

The drawback of zero-cost proxies is that they often need access to training data, and in some settings, this might be impossible, e.g. due to privacy reasons. Most proxies are computed by providing a mini-batch to networks, thus requiring us to compile the whole model. For very large architectures, or for networks with proprietary training routines, it might be advantageous to work only with encodings without any overhead. Additionally, although they are labeled zero-cost, some of them can take minutes per architecture to compute \citep{nbsuitezero}. The advantage of our proposed graph features is that they are completely data-independent.

\section{Introducing Neural Graph Features}

\begin{figure*}[ht]
\begin{center}
    \begin{subfigure}[t]{0.325\textwidth}
        \centering
        \includegraphics[width=0.93\textwidth]{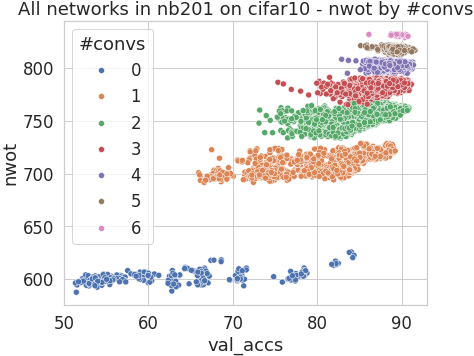}
        \caption{\texttt{nwot} against \texttt{cifar-10} val. acc.}
        \label{fig-nb201-nwot}
    \end{subfigure}%
    \begin{subfigure}[t]{0.325\textwidth}
        \centering
        \includegraphics[width=0.955\textwidth]{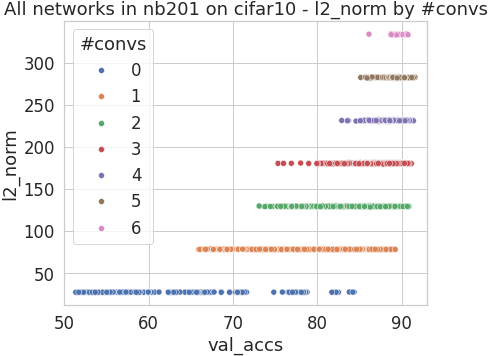}
        \caption{\texttt{l2\_norm} against \texttt{cifar-10} val. acc.}
        \label{fig-nb201-l2}
    \end{subfigure}%
    \begin{subfigure}[t]{0.34\textwidth}
        \centering
        \includegraphics[width=0.95\textwidth]{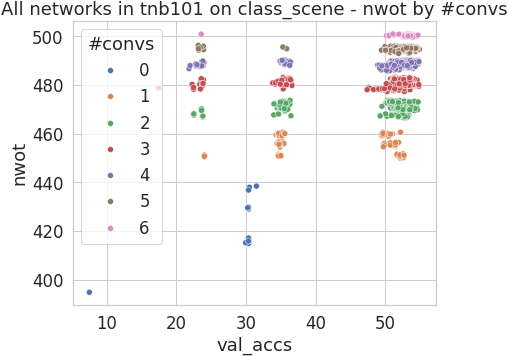}
        \caption{\texttt{nwot} against \texttt{class\_scene} val. acc.}
        \label{fig-tnb101-nwot}
    \end{subfigure}

\caption{ZCP score of all networks from NB201 against the validation accuracy - colors indicate the number of conv3x3 and 1x1}
\label{fig-scatter}
\end{center}
\end{figure*}

In our work, we will use precomputed zero-cost proxy scores from NAS-Bench-Suite-Zero~\citep{nbsuitezero}. Refer to Section \ref{app-zcp} in the appendix for more details about the provided zero-cost proxies, and Section \ref{app-benches} for the used benchmarks, datasets, and abbreviations. For NB101 and NB301, zero-cost proxies were computed only for a fraction of the search space, and all subsequent experiments are evaluated on these samples. For NB201 and TNB101-micro, we also exclude networks with unreachable branches due to zero operations. Detailed information about the process is in Section \ref{appendix-unreachable} in the appendix.

To motivate our approach, we first examine the limitation of many zero-cost proxies, namely their correlation with the layer count. Based on these insights, we then propose our simple-to-compute graph properties which we name neural graph features (GRAF).

\subsection{Limitations of zero-cost proxies}
As shown in NB-Suite-Zero, many existing zero-cost proxies have a good correlation with validation accuracy on NB201 and \texttt{cifar-10}. In this section, we investigate the reason for the good results.

From all ZCP available in NB-Suite-Zero, \texttt{nwot} has the best spearman correlation with validation accuracy. However, there is a hidden property of the proxy that brings about the good correlation. Figure \ref{fig-nb201-nwot} shows the \texttt{nwot} score of all NB201 networks plotted against validation accuracy, where the color of a point indicates the number of \texttt{conv1x1} plus \texttt{conv3x3} (i.e. all convolutions) in the architecture. We can see that each cluster corresponds to a specific number of convolutions, and the \texttt{nwot} score increases with the number of convolutions. Figure \ref{fig-nb201-l2} depicts the same behavior for \texttt{l2\_norm}, where the score is constant for every cluster. Similar results were shown with NB-Suite-Zero, however, they discovered only the correlation with the number of convolutions, not direct dependence.

\begin{figure*}[h]
\vskip 0.2in
\begin{center}
    \begin{subfigure}[t]{0.33\textwidth}
        \centering
        \includegraphics[width=\textwidth]{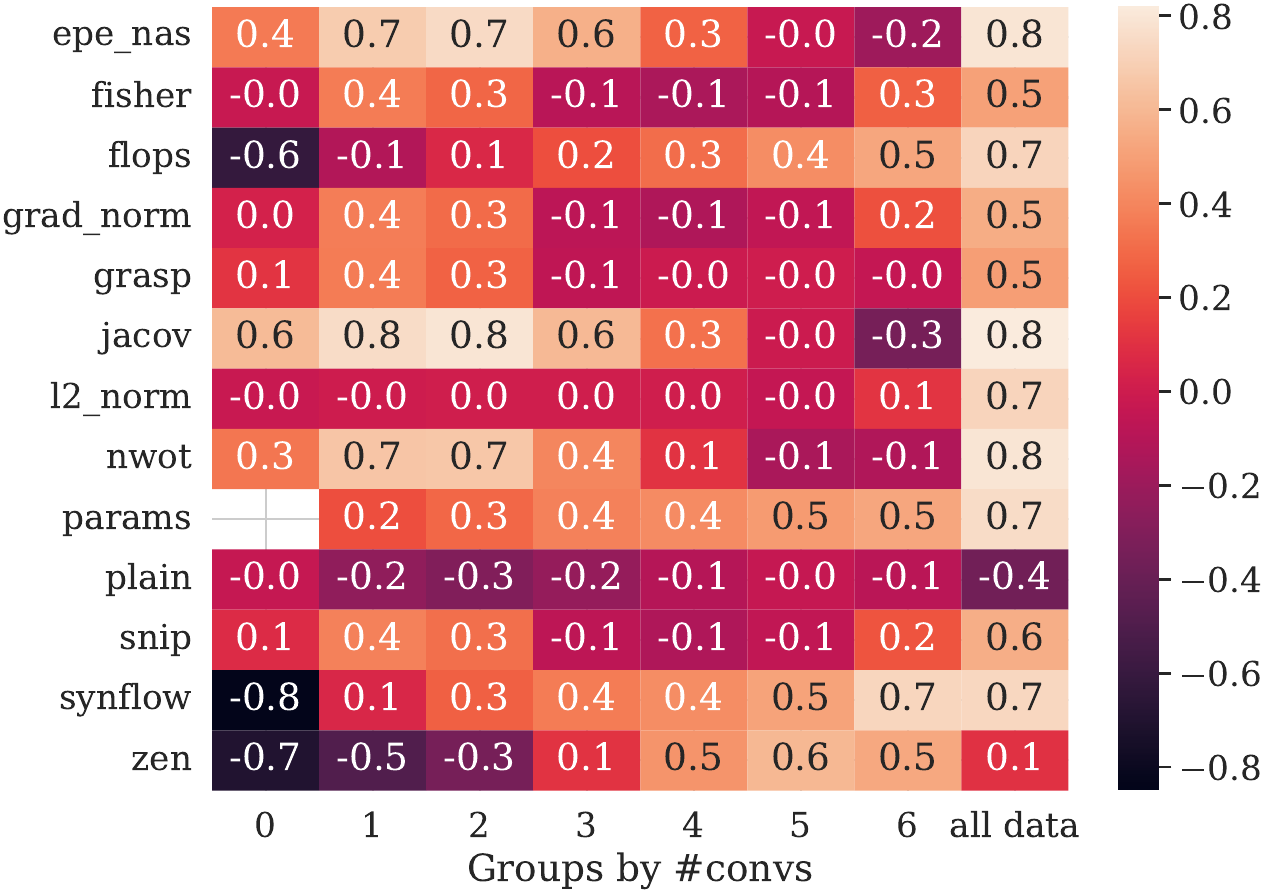}
        \caption{NB201 \texttt{cifar10}}
        \label{fig-heatmaps-nb201}
    \end{subfigure}%
    \begin{subfigure}[t]{0.33\textwidth}
        \centering
        \includegraphics[width=\textwidth]{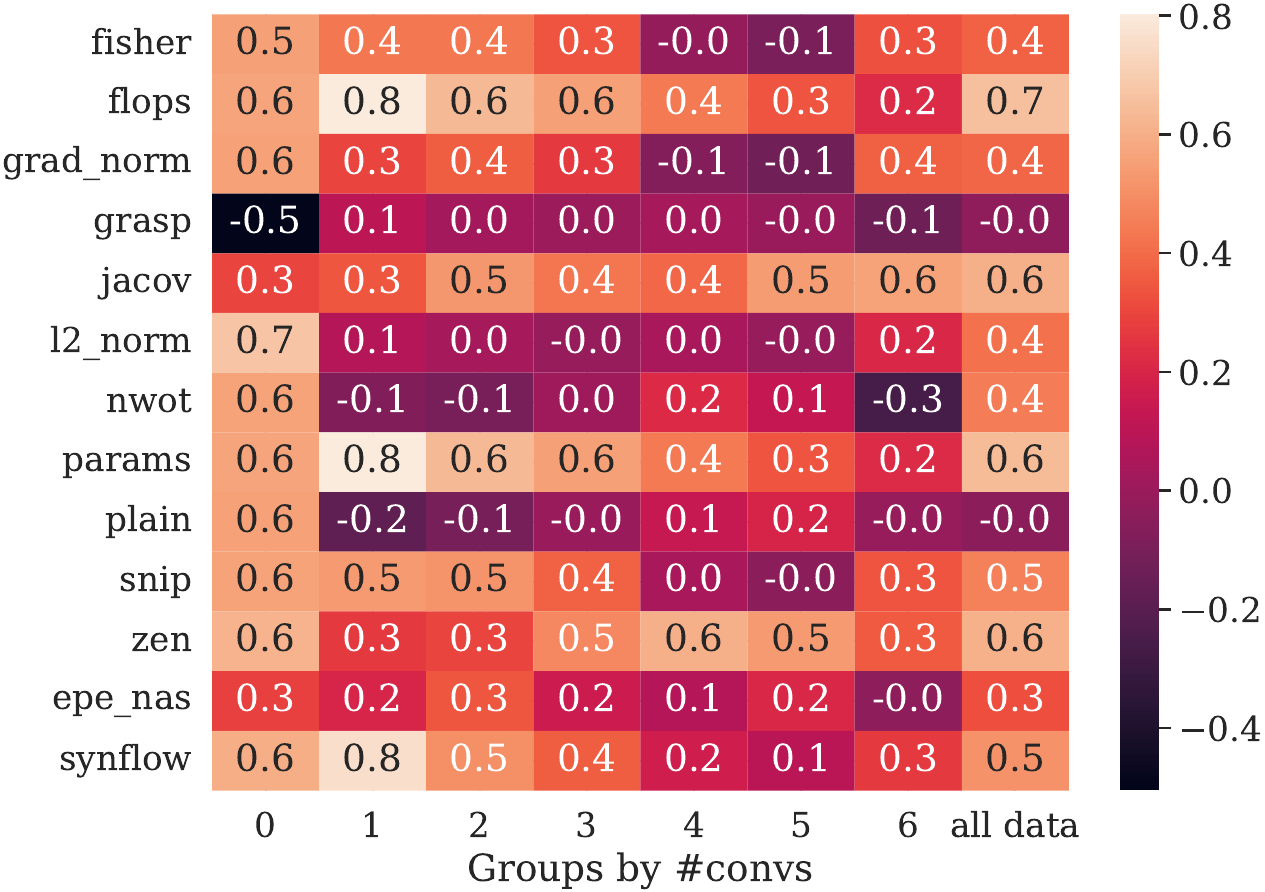}
        \caption{TNB101-micro \texttt{class\_scene}}
    \end{subfigure}%
    \begin{subfigure}[t]{0.33\textwidth}
        \centering
        \includegraphics[width=\textwidth]{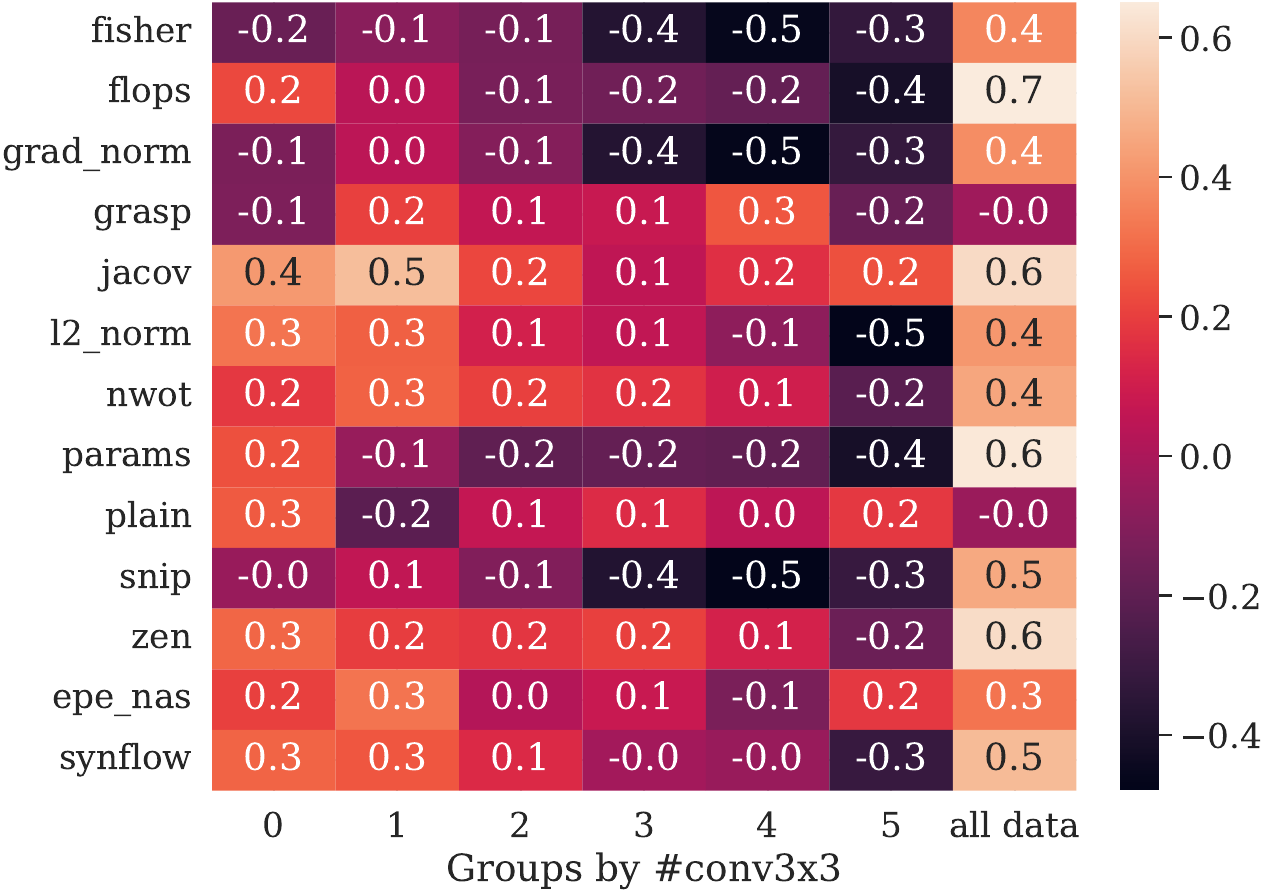}
        \caption{TNB101-micro \texttt{class\_scene}}
    \end{subfigure}

\caption{Spearman correlation of zero-cost proxies with validation accuracy by clusters of architectures with the same number of 1x1 and 3x3 convolutions (a,b), and \textbf{only} 3x3 convolutions (c). \textit{0-6} represents clusters with corresponding convolution counts, \textit{all data} is the correlation over all available networks (same as in NB-Suite-Zero \citep{nbsuitezero}).}
\label{fig-heatmaps}
\end{center}
\vskip -0.2in
\end{figure*}

In Figure \ref{fig-heatmaps-nb201}, we can see that although most proxies have a good correlation over the whole search space, many have trouble distinguishing networks with the same number of convolutions, and most have trouble distinguishing networks with a high number of convolutions. For example, the row with \texttt{jacov} has a correlation equal to 0.8 for the cluster with 1 convolution, but in the cluster with 5 convolutions, the correlation is 0.

The same property applies for \texttt{cifar-100} and \texttt{ImageNet16-120} (Figure \ref{fig-imgnet-ncovns} in the appendix).  Contrary to NB201, the proxies had a fairly low correlation with TNB101-micro tasks. Figure \ref{fig-tnb101-nwot} explains why -- unlike \texttt{cifar-10}, the task performance does not correlate with the number of convolutions, and \texttt{nwot} does not capture some property that would be important for the task.
Figure \ref{fig-heatmaps} also shows, that for \texttt{class\_scene}, the proxies can distinguish networks with the same number of convolutions. However, for clusters with the same number of \texttt{conv3x3}, the proxies have a negative correlation with validation accuracy. On NB101 and NB301, the dependence on the number of convolutions is not direct, but there are other dependencies (Section \ref{moar-biases} in the appendix).

To summarize, zero-cost proxies capture the number of convolutions, an important property for NB201 tasks like \texttt{cifar-10} classification, but most of them cannot distinguish structurally similar networks -- they lack some other important properties. 
These findings have inspired us to examine the potential of using operation counts and other graph properties as input for performance predictors. 

\subsection{Neural graph features}
In this section, we describe the \emph{neural graph features} (GRAF). Given operation set $O$ and an architecture graph $G=(L, V, E)$, where $V$ is the vertex set, $E$ is the edge set, and $L$ is the set of labels (associated with edges or vertices based on the search space type, see Section \ref{app-benches}), we define the following features:

\begin{itemize}
    \item Number of times the operation $o\in O$ is used in $L$
    \item Minimum path length from the input node to output node going only over operations $O' \subseteq O$
    \item Maximum path length from the input node to the output node going only over operations $O' \subseteq O$
    \item Output degree of the input node counting only operations $O' \subseteq O$
    \item Input degree of the output node counting only operations $O' \subseteq O$
    \item Mean input/output degree of intermediate nodes counting only operations $O' \subseteq O$
\end{itemize}

\begin{figure*}[h]
\vskip 0.2in
\centering
    \begin{subfigure}{0.24\textwidth}
        \centering
        \includegraphics[height=1.2cm]{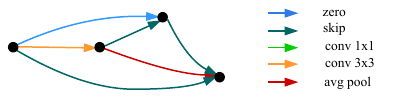}
            \vskip 0.1in
        \caption{Exemplary cell}
    \end{subfigure}
    \begin{subfigure}{0.24\textwidth}
        \centering
        \includegraphics[height=1.2cm]{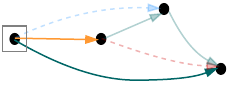}
    \vskip 0.1in
        \caption{Input node degree in allowed \newline [conv 3x3, skip] = 2}
        \label{fig-GRAF-node}
    \end{subfigure}
    \begin{subfigure}{0.24\textwidth}
        \centering
        \includegraphics[height=1.2cm]{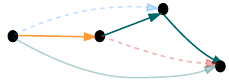}
            \vskip 0.1in
        \caption{Maximal path length over \newline  [conv 3x3, skip] = 3}
        \label{fig-GRAF-max}
    \end{subfigure}
    \begin{subfigure}{0.24\textwidth}
        \centering
        \includegraphics[height=1.2cm]{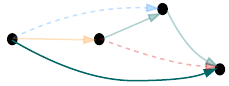}
            \vskip 0.1in
        \caption{Maximal path length over [skip]=1}
        \label{fig-GRAF-max2}
    \end{subfigure}
\caption{Exemplary overview of neural graph features in the NAS-Bench-201 search space.}
\label{fig-GRAF}
\end{figure*}
These features are computed for all possible subsets $O' \subseteq O$. They are the same for all cell-based benchmarks with one exception -- for NB301, min path length is not computed, since all nodes are directly connected to the output. Figure \ref{fig-GRAF} visualizes three different example features computed for an architecture. For node degree (\ref{fig-GRAF-node}), the computed value is 2, since we count the conv3x3 and skip edges going from the input node, but not the zero edge. For the first max path length (\ref{fig-GRAF-max}), the computed value is 3, as we can take a long way over 1 skip edge and 2 conv3x3 edges. However, for the second max path length (\ref{fig-GRAF-max2}), the value is only 1, since we use only the skip edge.

We designed GRAF to reflect findings in recent work as well as zero-cost proxy properties. Operation counts stem from the analysis from the previous section. Maximum path length represents the depth of the network, and minimum path length tells us whether there are shortcuts from the input to the output -- for example, whether there is a skip connection. Node degree features can be compared to motifs from NASBOWL -- some best-performing motifs included the input node with a specific operation pattern \citep{nasbowl}.

In the appendix, we include Table~\ref{tab:countsandtimes} that shows GRAF counts with computation time per benchmark (always well below one second per network). We also include macro features in Section \ref{app-GRAF-macro}; these are used for TNB101-macro and are based on channel and stride counts. 

\begin{figure*}[ht]
\includegraphics{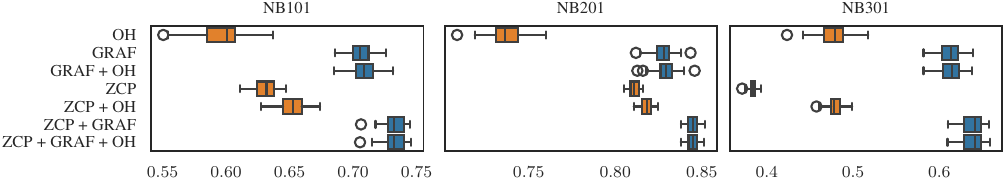}
\vskip -0.05in
\caption{Validation accuracy prediction (\texttt{cifar-10}) across 3 benchmarks with 1024 sampled networks. Comparison of 3 encodings (ZCP, onehot, and GRAF) and their combinations. Blue boxes denote runs including GRAF.}
\label{fig-valacc-1024}
\end{figure*}
\subsection{Using Graph Features in Prediction}
\label{sec:graf}

In this section, we use GRAF as input data for validation accuracy prediction with a random forest predictor. We also include zero-cost proxies (ZCP), one-hot encoding (onehot) and combinations of these encodings. We examine variants where we include only \texttt{flops} and \texttt{params} (FP) instead of all ZCP, since they can be calculated without any batches passed through the networks. We evaluate the different settings on all available benchmarks and datasets, for 3 train sample sizes (32, 128 and 1024) and across 50 seeds. We report Kendall tau for every run. Full results are available in Section \ref{appendix-val-accs} in the appendix.

\begin{figure*}[ht]
\includegraphics{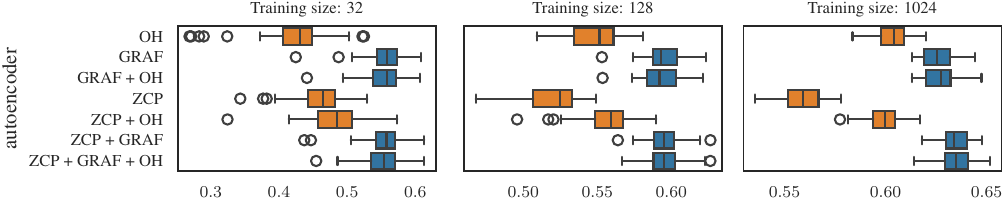}
\vskip -0.05in
\caption{TNB101\_micro \texttt{autoencoder} task across 3 train sample sizes. Comparison of 3 encodings (ZCP, onehot, and GRAF) and their combinations. Blue boxes denote runs including GRAF.}
\label{fig-tnb101-autoencoder-body}
\end{figure*}

Figure \ref{fig-valacc-1024} shows results for 1024 sampled train networks on \texttt{cifar-10} and NB101, NB201, and NB301. GRAF performs better than ZCP and onehot in all cases, and ZCP + GRAF yields the overall best results. Figure \ref{fig-tnb101-autoencoder-body} shows results for the \texttt{autoencoder} task and TNB101-micro. We can see that using GRAF leads to better results than ZCP or OH across all sample sizes. For other TNB-micro tasks (Figures \ref{app-tnbmicro-1}, \ref{app-tnbmicro-2}) and TNB-macro (Figures \ref{app-tnbmacro-1}, \ref{app-tnbmacro-2}), the best configurations always include GRAF.
Overall, GRAF performs the same or better than ZCP even on the smaller sample sizes (32, 128) across all benchmarks and tasks (Section \ref{appendix-val-accs} in the appendix) except for NB201, where it is slightly worse (Table \ref{app-nb201-boxes}). 

These results show that while zero-cost proxies have some flaws, they still capture important network properties that help the prediction. The strong performance of ZCP + GRAF suggests that GRAF does not capture all network properties, and ZCP and GRAF complement each other well for prediction. It is important to note that for some tasks beyond image classification, using GRAF on its own is sufficient.

\subsection{Feature importance}
\label{sec:featimp}

\begin{table*}[h]
\caption{The ten most important features on NB201 \texttt{cifar-10} and TNB101-micro - \texttt{autoencoder}. Ranking is based on Shapley values computed from random forest trained on 1024 training samples with ZCP+GRAF.}
\label{featimp-nb201}
\vskip 0.1in
\small
\centering
\begin{tabular}{lrllr}
\hline
  \multicolumn{2}{c}{NB201 - \texttt{cifar-10}} && \multicolumn{2}{c}{TNB101-micro - \texttt{autoencoder}} \\
Feature name &  Mean rank & $\quad$ &                   Feature name &  Mean rank \\
\hline
                         jacov &       0.00 &  &       min path over skip &       0.00       \\
                          nwot &       1.12 &  &                    jacov &       1.00       \\
                         flops &       3.62 &  &                   fisher &       2.00       \\
                       synflow &       4.08 &  &  min path over [skip,C3x3] &       5.50     \\
min path over [skip,C3x3,C1x1] &       4.78 &  &                     snip &       5.58       \\
                        params &       5.04 &  &  min path over [skip,C1x1] &       5.64     \\
                       epe\_nas &       6.04 & &                grad\_norm &       6.64      \\
                           zen &       6.36 &  &                      zen &       8.08       \\
     min path over [skip,C3x3] &      11.08 &  &                    grasp &       9.34       \\
            min path over skip &      11.88 &  &                  l2\_norm &       9.74      \\
\hline
\end{tabular}
\end{table*}

We analyze the feature importance of random forest fit on ZCP + GRAF using SHAP~\citep{shap}. For each of the 50 runs, we compute the mean absolute Shapley value for all features. Then, we sort the values, obtain ranks, compute mean rank across the 50 runs and list the 10 features with the lowest (best) ranks.

Table \ref{featimp-nb201} shows the most important features for NB201 and \texttt{cifar-10}, and TNB101-micro and \texttt{autoencoder}. For \texttt{cifar-10}, many ZCP have low ranks, and from GRAF, skip-connection and convolution min path length (shortcuts) are the most important. However, on TNB101-micro, the situation is different. For the \texttt{autoencoder} task, shortcuts with skip-connection and average pooling are important features (confirming findings of \citet{LOPES2023119695} -- they found that \texttt{autoencoder} had skip-connections in well-performing networks).
In the appendix, Table~\ref{featimp-tnb101img} shows that while \texttt{ImageNet16-120} important features overlap with \texttt{cifar-10}, for \texttt{class\_scene}, node degree features are important.

In Section~\ref{app-featimp} in the appendix, we list feature importances for the other benchmarks and tasks. Although NB101 and NB301 also evaluate architectures on \texttt{cifar-10}, the important features are different -- notably, \texttt{jacov} was highly ranked on NB201 but is not in the top 10 features for these benchmarks. Similarly, node degree features are important for NB101 and NB301, but not present in NB201. This indicates that search space design has a great impact on performance prediction -- models may need to learn orthogonal architectural properties. 

\subsection{Analysing feature redundancy}

\begin{figure}[h]
\includegraphics{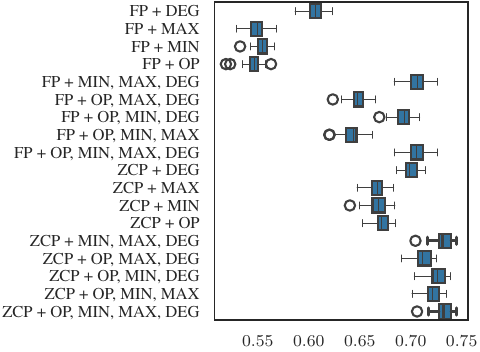}
\vskip -0.05in
\caption{Ablation study on GRAF groups -- analyzing cases when only 1 group is present vs when keeping all but 1 group.}
\label{fig-nb201-abla}
\end{figure}

\paragraph{Group level}
First, we analyze the contribution of the following feature groups: max path length (MAX), min path length (MIN), operation count (OPS), and node degree (DEG). We look at performance of ZCP + one feature group, and ZCP + all but one feature group. We compare two cases: with all ZCP, and only with \texttt{flops} and \texttt{params} (FP). 
Figure \ref{fig-nb201-abla} shows random forest prediction across 50 seeds on 1024 networks (NB201, \texttt{cifar-10}). We see that DEG has the highest individual contribution. However, when using all but one group, leaving out MIN leads to the greatest drop in performance. It also seems that OP do not have a large influence when other feature groups are present, indicating they might possibly be computed from the other features or from FP. However, we still use them in our models for better interpretability. Figures \ref{app-abl-nb101} and \ref{app-abl-nb201} in the appendix show similar results for NB101 and NB201 + \texttt{ImageNet16-120} --- all groups except OP are needed, but the most influential groups differ.

\paragraph{Feature level}

Since GRAF is designed for all subsets of the operation set, some features might be redundant. To identify redundancies, we examine the linear dependence of features. Using linear regression on the GRAF feature set, we predict one feature using all other features. Then, if the prediction for a feature is perfect, we remove it from the dataset. Doing so, we iteratively eliminate all redundant features until we get a linearly independent set of features.

This process resulted in 30 features for NB101, 72 for NB201, and 546 for NB301. However, when we compare the results of a random forest on the smaller feature set versus the original set, the results with the smaller set are worse if using only the GRAF subset (refer to Table~\ref{ap:tab:redundancy} in the appendix). When combined with ZCP, the performance is the same as the full GRAF + ZCP on NB101 and NB201, yet still worse on NB301. A possible explanation could be that the models are not able to construct the removed features from the subset of features during learning. ZCP cover some of the performance loss, but fail for a more complex search space like NB301.

\section{Empirical Evaluation on Diverse Tasks}

\begin{figure*}[ht!]
\includegraphics{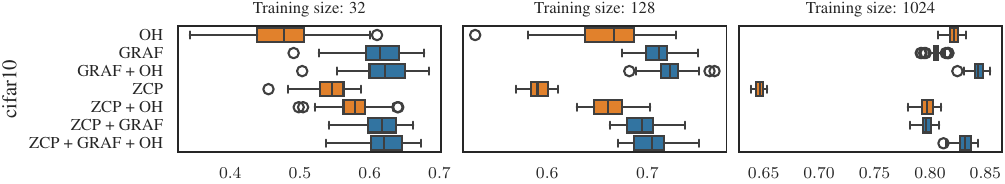}
\caption{Random forest \texttt{edgegpu\_energy} prediction  on NB201 \texttt{cifar-10} (different training sample sizes). }
\label{fig-nb201-edgegpu}
\end{figure*}
\begin{figure}[h!]
\begin{center}
    \includegraphics{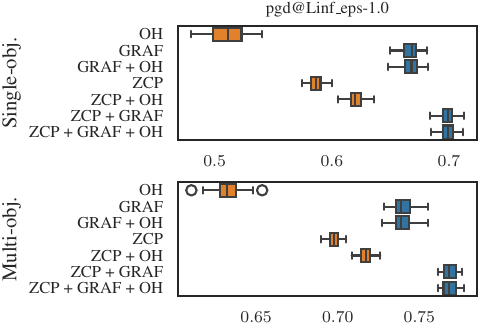}
\caption{Robust accuracy prediction for the PGD adversarial attack with 1024 training data on the attack perturbations $\epsilon=1.0/255$ for the objectives, only robust accuracy (top), and both, validation and robust accuracy (bottom). NB201 and \texttt{cifar-10}.}
\label{fig-nb201-pgd}
\end{center}
\end{figure}

In this section, we evaluate GRAF on tasks beyond validation accuracy prediction -- namely hardware metrics and robustness tasks. Then, we examine how ZCP and GRAF can be included into BRP-NAS.
We also compare GRAF with existing encodings and performance predictors. Lastly, we use the ZCP + GRAF predictor in a search setting on NB201. Hardware, robustness and BRP-NAS experiments have the same experiment settings as in Section~\ref{sec:graf}.

\subsection{Other Prediction Tasks}
\paragraph{Hardware tasks}
Our proposed GRAF features were also evaluated on hardware metrics from HW-NAS-Bench~\citep{hwnasbench}. This includes mainly energy and latency on different devices, in total 10 metrics for 3 datasets.  The prediction tasks are of varying difficulty, ranging from easy tasks to difficult ones. Onehot is generally a good predictor in HW-NAS-Bench tasks as shown by~\citet{hwpredictors}.
Figure \ref{fig-nb201-edgegpu} shows results on the \texttt{edgegpu\_energy} task. GRAF is better than ZCP on all sample sizes, and better than onehot on smaller sample sizes. GRAF + onehot yields the overall best results. 
Additional results can be found in the appendix~\ref{ap:hwexp}, they show that the best setting contains GRAF among predictors on the vast majority of tasks.

\paragraph{Robustness tasks}
In the following section, we evaluate how GRAF improves the results on the task of predicting the robustness of the architectures. We use the evaluated robustness of the NAS-Bench-201 \citep{nb201} from the robustness dataset \citep{robustness}. This dataset evaluated the architectures from NAS-Bench-201 on different adversarial attacks and different perturbation strength; we consider in this paper two white-box attacks (FGSM \citep{fgsm}, and PGD \citep{pgd}. 
Important to note that the adversarial robustness is highly correlated with the validation accuracy since the former can never improve over the latter. Here, we will differentiate the robustness evaluation into two different tasks: (i) predicting only the robustness accuracy, and (ii) predicting both, the validation accuracy \textbf{and} the robustness accuracy jointly (\textit{Multi\_Obj}). In more detail, we evaluate the prediction ability on both attacks for three different perturbation strengths ($\epsilon =\{0.1/255, 0.5/255, 1.0/255 \} $) on \texttt{cifar-10} using three different training sizes for learning the random forest prediction model. For the first prediction type (i), the combination of zero-cost proxies, the proposed graph features, and the additional onehot encoding shows the highest Kendall tau in all considered tasks (see Section \ref{app_robustness} in the appendix for a detailed overview).  Furthermore, with increasing perturbation strength, using only the graph features shows a better prediction ability than using only the zero-cost proxies from the literature.  
The same behavior is visible for the joint objective prediction task (ii). The latter task of predicting both accuracies (validation and robust) at once shows also a higher Kendall tau. Thus, this seems to be the easier task for the prediction model, which was also shown in \citet{lukasik2023evaluation}.
We present the boxplots for the PGD attack for perturbation strength $\epsilon = 1.0/255 $ for both evaluation tasks in \ref{fig-nb201-pgd} and more results on both adversarial attacks in both settings in Section \ref{app_robustness} in the appendix.

\subsection{Other Types of Encoding}
Apart from the GRAF features and the zero-cost proxies, we also experimented with other types of network architecture encodings -- path encoding, arch2vec \citep{arch2vec}, and NASBOWL's Weisfeiler-Lehman features (WL) for the number of iterations equal to $1$. \citep{nasbowl}. Detailed results of these experiments are included in the appendix (Tables \ref{app-nasbowl-nb201} and \ref{app-nasbowl-nb101}). Here, we give just a brief overview.

Adding path encoding to the GRAF features does not improve the results, with the exception of \texttt{cifar-10} (NB101, NB201 and NB301) and \texttt{cifar-100} (NB201) on the largest sample size. However, the increase from ZCP + GRAF is small and is not observable on \texttt{ImageNet16-120} or TNB101-micro tasks. When GRAF is not used, ZCP + path encoding outperforms ZCP. This indicates that the GRAF features already contain most of the useful information from path encoding. The same behavior can be observed for the WL features and arch2vec, as they improve the results without GRAF, but when GRAF is present, the improvement is only small.

\subsection{Existing performance predictors}
\paragraph{BRP-NAS}

We compare our results with BRP-NAS~\citep{brpnas}, also including variants when BRP-NAS takes additional feedback from ZCP, GRAF, or onehot. This additional feedback is concatenated with the output from the BRP-NAS GCN and used in the output layer. A detailed description of this process and of the results of the experiments can be found in Section~\ref{sec:appbrpnas} in the appendix.

On most tasks, the best combination of BRP-NAS with the ZCP + GRAF features typically does not have better results than the best random forest combination. However, both ZCP + GRAF improve the BRP-NAS performance over the pure BRP-NAS in most cases.

On the accuracy task (cf. Table~\ref{tab:brp:acc}) on NB201 BRP-NAS is slightly better than the RF-based models only with the largest training size for the ImageNet16-120 target. On the NB301 benchmark, however, BRP-NAS has the same or better results for all training sizes, which may indicate that the GCN may be able to extract additional important information from the graph structure. On the hardware tasks (Table~\ref{ap:brp:hw1})
, the original BRP-NAS has a similar performance to the best combination of features with RF-based predictors. Including the other information improves performance in all cases, with BRP-NAS + GRAF + OH being among the best configurations. Interestingly, very often the performance of BRP-NAS + ZCP is not much better than the performance of BRP-NAS alone. On the other hand, on the robustness tasks (Tables~\ref{ap:brp:advs},\ref{ap:brp:adve}), BRP-NAS has slightly worse results than RF-based models with GRAF and onehot features. Additionally, unlike the other tasks, additional information does not improve the results, except for the smallest perturbation and the smallest training set. An explanation could be that the properties relevant for prediction are captured both by BRP-NAS and ZCP + GRAF.

\paragraph{NASLib}
\label{naslib-perfpreds}
In this section, we compare the graph features with other predictors using the same setting as in the performance predictor survey by \citet{naslib-predictors}. We chose ZCP + GRAF for evaluation, as it performed the best across various settings. We use subsets of NB301 and NB101, as ZCP from NB-Suite-Zero are available only for a limited number of architectures. Figure \ref{fig-predictors} shows results for NB101, and detailed results are in Section \ref{naslib-predictors} in the appendix, along with more information about the different predictors and ZCP + GRAF model variants.

\begin{figure}[h]
\centering
\includegraphics[width=\columnwidth]{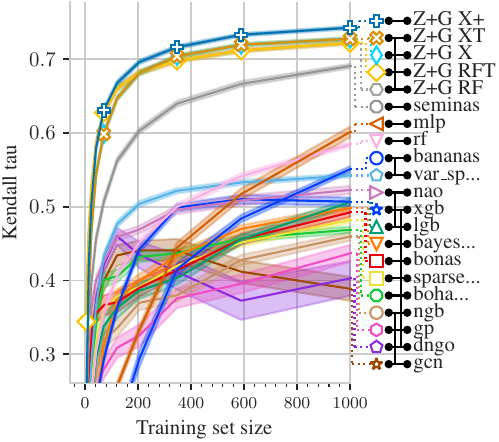}
\vskip -0.1in
\caption{Evaluation of NASLib performance predictors on a subset of NB101, \texttt{cifar-10}.}
\label{fig-predictors}
\end{figure}

In all cases, ZCP + GRAF models are the best-performing predictors. The predictor results match the original results from the survey except for NB101 (Figure \ref{fig-predictors}), where BANANAS \citep{white2019bananas} and GCN \citep{gcnpred} are much worse in our case. An explanation could be that the sampled NB101 set is too diverse, and when sampling a train set from it, all remaining networks are too dissimilar for both predictors. This claim is supported by the encoding study by \citet{encodings}, where path encoding had poor results on networks outside of a train set limited to a subspace of the search space.

\paragraph{TA-GATES}
Next, we compared GRAF with TA-GATES, a well-performing graph neural network predictor. We used the same models as in the previous section (without tuning). On small sample sizes, ZCP + GRAF performs better than TA-GATES. On larger sample sizes, only XGB+ has a similar performance. Although predictors with ZCP + GRAF are faster and more interpretable, using well-performing graph neural networks might be a promising direction in achieving better performance prediction. Full details are provided in the Section \ref{app-tagates}.

\subsection{NASLib search run}
Lastly, we evaluate ZCP + GRAF in a search setting.
As in Section \ref{naslib-perfpreds}, we repeat the experiments from the predictor survey and use the predictor as a surrogate in Bayesian optimization. We again use the same models for ZCP + GRAF except for the tuning. More details are included in Section \ref{app-naslib-search} in the appendix.

Figure \ref{fig-search-img} shows the results of the search on \texttt{ImageNet16-120}. ZCP + GRAF is on average more sample efficient than the other predictors. On \texttt{cifar-10} (Figure \ref{fig:cifarsearch}), SemiNAS \citep{seminas} has on average the same performance as GRAF in later search stages. Thus, a more extensive search evaluation would be needed as a part of future work, ideally including various tasks and optimizers.

\begin{figure}[h]
\includegraphics[width=\columnwidth]{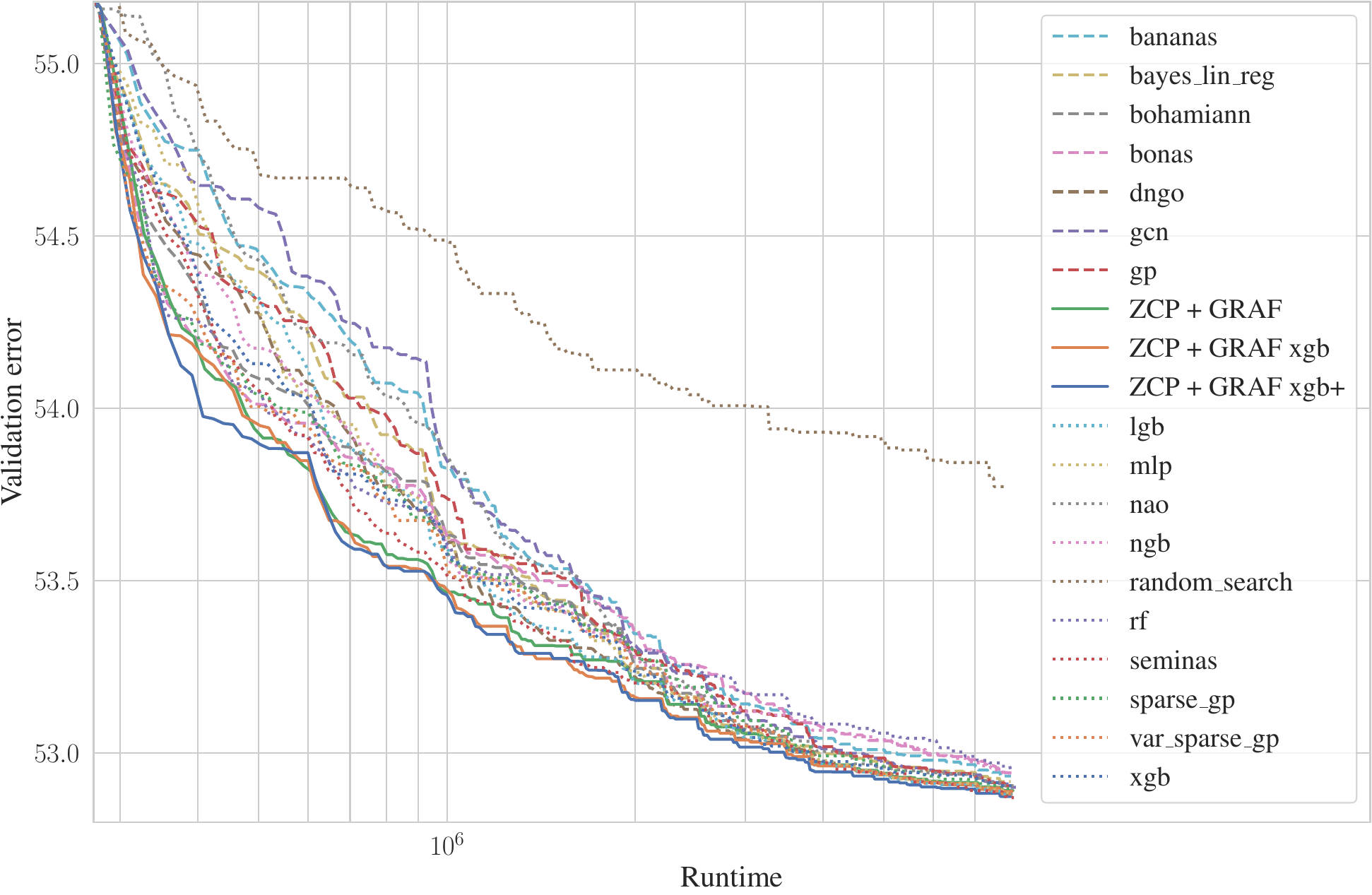}
\vskip -0.1in
\caption{Bayesian optimization -- search on \texttt{ImageNet16-120} and NB201.}
\label{fig-search-img}
\end{figure}

\section{Conclusion}
To summarize, we introduced GRAF, simple-to-compute graph features inspired by the shortcomings of zero-cost proxies. When used as the input to a random forest predictor, they outperform ZCP and other common encodings.
Their interpretability enabled us to highlight that different tasks favor different network properties.
We evaluated GRAF extensively across a variety of tasks (including metrics beyond validation accuracy), showing very strong results overall. We demonstrated that they can also improve existing models such as the graph neural network BRP-NAS. Lastly, we evaluated them in prediction and search settings, where they outperformed all available predictors. 

\section{Discussion}

The potential of GRAF is to bring a better understanding of existing search spaces, tasks, and existing performance predictors. In fact, due to the simple design, GRAF could be a good baseline for more complex predictors -- especially since many predictors for NASLib had worse results while consuming more resources. GRAF could also inspire new zero-cost proxies that capture specific network properties and correlate better with tasks beyond \texttt{cifar-10} classification. Lastly, since GRAF macro had better results than proxies and the onehot encoding, it could be easily extended to new macro search spaces and transformer search spaces. An interesting question is whether we could scale up the networks according to the most important features while keeping a good performance.

When compared with other predictors, TA-GATES in performance prediction and SemiNAS in \texttt{cifar-10} search matched XGBoost \citep{xgboost} trained on ZCP + GRAF for the respective tasks. Also, BRP-NAS combined with GRAF had the best results on NB301 compared to the random forest results. This suggests that more complex models are still promising and combination with additional information like ZCP or GRAF may be the key to better results across search spaces.

A disadvantage of GRAF is that if a search space differs significantly from the cell-based search spaces and macro search space used in this work, new graph features need to be designed. Also, even for similar search spaces, transferability is limited due to different operation sets.
However, due to their simplicity, designing new graph features should not be hard. Transfer between search spaces might even be an ill-posed problem due to different architecture training pipelines -- also supported by the fact, that important features for \texttt{cifar-10} were different across search spaces.

A major drawback is that for the best results, zero-cost proxies are still needed. One possible explanation could be that there might be some graph properties that we have not used. Another hypothesis could be that while network performance depends on the neural graph, some part of it also depends on some properties of the information flow or training dynamics that cannot be captured by analyzing the architecture only. This problem opens an interesting direction for future research. 

\section*{Acknowledgements}
This research was funded by the Deutsche Forschungsgemeinschaft (DFG, German Research Foundation) under grant number 417962828.
Computational resources were provided by the e-INFRA CZ project (ID:90254),
supported by the Ministry of Education, Youth and Sports of the Czech Republic.
This research was partially supported by the SVV project number~260~698.
G. Kadlecov\'a was supported by Charles University Grant Agency project no.~246322.
P. Vidnerov\'a was supported by the project 22-02067S of the Czech Science Foundation.
J. Lukasik acknowledges support by the DFG research unit 5336 - Learning to Sense.
Frank Hutter is a Hector Endowed Fellow at the ELLIS Institute Tübingen.
We would like to thank Arjun Krishnakumar for the valuable discussions that helped to shape the initial ideas of this paper.

\section*{Impact Statement}
This paper presents work whose goal is to advance the field of Machine Learning. Specifically, our work is in the field of performance prediction, which aims to reduce the costs of network evaluation. We believe we contributed to this goal, as our predictors outperform existing methods while being cheap to fit. This enables energy savings in future NAS applications.

Since our method is interpretable and has a good performance across various tasks, we could gain insight into how network components influence other important metrics, namely fairness. This could get us closer to designing more fair deep learning applications. The interpretability could also make NAS more attractive for fields where it is important -- for example medical or decision-making fields.

\bibliography{GRAF}
\bibliographystyle{icml2024}

\newpage
\appendix
\onecolumn

\section{NAS Best Practice Checklist}\label{app:nas_best_practice_checklist}
We now describe how we addressed the individual points of the NAS best practice checklist~\citep{lindauer2020best}.

\begin{enumerate}

\item \textbf{Best Practices for Releasing Code}\\[0.2cm]
For all experiments you report: 
\begin{enumerate}
  \item Did you release code for the training pipeline used to evaluate the final architectures?
    \answerNA{We query architectures from NAS benchmarks (via NASLib)}
  \item Did you release code for the search space
  \answerNA{We use search spaces from NASLib}
  \item Did you release the hyperparameters used for the final evaluation pipeline, as well as random seeds?
  \answerNA{}
  \item Did you release code for your NAS method?
  \answerYes{Link to a public github repository will be provided upon acceptance. Code is a part of the submission.}
  \item Did you release hyperparameters for your NAS method, as well as random seeds?
  \answerYes{We provide hyperparameters to the XGB+ model and the BRP-NAS model. Other models have default hyperparameters or are the same as in previous work (NASLib, TA-GATES).}   
\end{enumerate}

\item \textbf{Best practices for comparing NAS methods}
\begin{enumerate}
  \item For all NAS methods you compare, did you use exactly the same NAS benchmark, including the same dataset (with the same training-test split), search space and code for training the architectures and hyperparameters for that code?
    \answerYes{For BRP-NAS, we used different splits than in RF experiments, but we
    compute the average over 50 seeds}
  \item Did you control for confounding factors (different hardware, versions of DL libraries, different runtimes for the different methods)?
    \answerYes{For experiments where predictors are compared in terms of runtime, we ran the predictors on the same hardware.}
	\item Did you run ablation studies?
    \answerYes{}
	\item Did you use the same evaluation protocol for the methods being compared?
    \answerYes{}
	\item Did you compare performance over time?
    \answerYes{}
	\item Did you compare to random search?
    \answerYes{}
	\item Did you perform multiple runs of your experiments and report seeds?
    \answerYes{Seeds are provided in the codebase, default values were used.}
	\item Did you use tabular or surrogate benchmarks for in-depth evaluations?
    \answerYes{}

\end{enumerate}

\item \textbf{Best practices for reporting important details}
\begin{enumerate}
  \item Did you report how you tuned hyperparameters, and what time and resources
this required?
    \answerYes{}
  \item Did you report the time for the entire end-to-end NAS method
(rather than, e.g., only for the search phase)?
    \answerYes{Applies mostly to NASLib experiments}
  \item Did you report all the details of your experimental setup?
    \answerYes{}

\end{enumerate}

\end{enumerate}

\clearpage
\section{Related Work (extended)}
\label{app-rel-work}
We extend Section \ref{relatedwork} and provide more information and examples of model-based predictors, as well as works on interpretability.
Given a train set of architectures and their performance $\{a, f(a)\}$, a model-based predictor $f'$ solves a regression task by learning to estimate $f(a)$ from the train set.
Many model-based predictors are based on graph neural networks that work directly with the graph structure of architectures. One example is BRP-NAS, a graph neural network latency and accuracy predictor. Its authors also demonstrated that in search, predicting binary relations between networks proved to be more sample efficient for accuracy prediction \citep{brpnas}.

Other predictors use models like Gaussian processes, tree-based methods, or neural networks that take as input architectures encoded as vectors \citep{naslib-predictors}. A study on neural encodings showed that the success of different encodings (one-hot encoding, path-encoding, and their variants) depends on the context in which they are used -- good encodings for search may fare worse in performance prediction \citep{encodings}. Another possibility how to encode the architectures is through architectural representation learning. \citet{arch2vec} have demonstrated that unsupervised representation learning (named arch2vec) leads to better embedding quality compared to the embedding extracted in the supervised accuracy prediction task \citep{arch2vec}.

In terms of interpretability, recent work analyzed the ``no free lunch'' theorem for architectures, where given a fixed budget, it is impossible to maximize the expressivity, convergence and generalization of an architecture \citep{nofreelunch}. The authors discovered that while expressive networks tend to be deep and narrow, convergence and generalization are biased toward wide and shallow topologies. Interestingly, previous work showed that NAS optimizers like DARTS \citep{darts} or ENAS \citep{enas} favor wide and shallow architectures \citep{Shu2020Understanding}.

A recent analysis of cell-based search spaces has shown that only a subset of the operation set is needed to generate high-performing architectures \citep{LOPES2023119695}. In fact, for \texttt{cifar10}, \texttt{cifar100}, and \texttt{ImageNet16-120} -- the most popular datasets for NAS evaluation -- the number of \texttt{conv3x3} is a crucial factor in network performance. The authors demonstrated more variability on TransNAS-Bench-101 tasks, encouraging the evaluation on various datasets and tasks.

\FloatBarrier
\section{Benchmarks, Zero-cost Proxies and Encodings}

\subsection{NAS Benchmarks}
\label{app-benches}

In this work, we used the same benchmarks as in NAS-Bench-Suite-Zero (NB-Suite-Zero) \citep{nbsuitezero} -- NAS-Bench-101 \citep{nb101}, NAS-Bench-201 \citep{nb201}, NAS-Bench-301 \citep{nb301} and TransNAS-Bench micro and macro \cite{tnb101}. Out of these benchmarks, only NB101 has operation labels on vertices, the other have operation labels on edges. It is important to note that TNB101-micro is a subset of NB201, and contains all networks without max pooling operations. All except for the macro search space TNB101-macro are cell-based search spaces.
Table \ref{tab:nasbenches} lists all benchmarks with abbreviations used throughout the paper, the number of sampled architectures to be used in experiments, and the total number of architectures in the search space. We used only the subsets of the search spaces for which pre-evaluated zero-cost proxies are available. For NB201 and TNB101-micro, we also decreased the number of networks due to a problem described in Section \ref{appendix-unreachable}.

Table \ref{tab:datasets} lists the datasets used for each benchmark, with non-classification tasks marked.

\begin{table}
  \caption{List of benchmarks used in our experiments with abbreviations and the number of sampled architectures. Total search space size is included for reference. \label{tab:nasbenches}}
   
\vskip 0.1in
\tabcolsep=4pt
\centering
\small
\begin{tabular}{lrrrrr}
\toprule
&  NAS-Bench-101 &  NAS-Bench-201 &   NAS-Bench-301 &  TransNAS-Bench-101 &  TransNAS-Bench-101 \\
&     &        &         &   -micro & - macro \\ 
\midrule
abbreviation &   NB101  &  NB201 &  NB301  &          TNB101-micro &         TNB101-macro\\
\# arch (sampled)  &    10~000 &   9~445 &   11~221 &           2~128 &          3~256 \\
\# arch (total)  &    423~624 &   9~445 &    $10^{18}$\scriptsize$^*$ &           4~096 &      3~256\\
\bottomrule
\scriptsize$^*$surrogate benchmark
\end{tabular}
\vskip -0.2 in
\end{table}

\begin{table}
  \caption{List of datasets used in our experiments. Task that are \emph{not} classification tasks are marked (*).\label{tab:datasets}}
\vskip 0.1in
\tabcolsep=4pt
\centering
\small
\begin{tabular}{ll}
\toprule
benchmark &  dataset  \\
\midrule
NB101 & \texttt{cifar-10} \\
NB201 & \texttt{cifar-10}, \texttt{cifar-100}, \texttt{ImageNet16-120} \\
NB301 & \texttt{cifar-10} \\
\hline 
TNB101 & \texttt{class\_scene}, \texttt{class\_object}, \texttt{autoencoder} (*), \texttt{jigsaw}, \\
(micro and macro) &  \texttt{normal} (*), \texttt{room\_layout} (*), \texttt{segmentsemantic} (*) \\

\bottomrule
\end{tabular}
\vskip -0.2 in
\end{table}

\subsection{Unreachable Branches in NB201 and TNB101-micro}
\label{appendix-unreachable}
\begin{figure*}[ht]
\begin{center}
\includegraphics[width=0.8\textwidth]{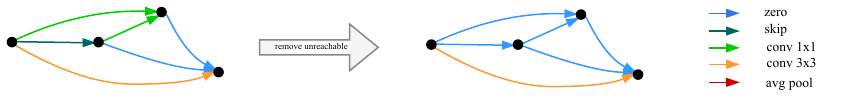}
\caption{Visualization of unreachable nodes in the NAS-Bench-201 search space and our adaptation.}
\label{fig-nb201-unreachables}
\end{center}
\end{figure*}

For NB201, due to the presence of zero operation, some networks have edges (non-zero operations) that do not receive any non-zero input, and in other networks, some operations are not connected to the output. Figure \ref{fig-nb201-unreachables} illustrates the problem, where multiple operations do not contribute to the output. 

In our work, we keep only networks without unreachable branches (Figure \ref{fig-nb201-unreachables}) for all NB201 and TNB101-micro experiments. The main reason is that some zero-cost proxies have scores influenced by the unreachable operations -- for example, \texttt{params} includes parameters of these operations. For validation accuracy prediction, this effect would be harmful, since \texttt{params} does not correspond to the true information flow. However, for hardware tasks, the unreachable operations might contribute to higher energies and latencies, and including unreachable parameters makes sense.

All in all, we believe removing unreachable operations in NAS experiments is good practice -- these networks should not be used in practice (due to higher energy costs), and removing the operations requires just a DFS run, i.e. $\mathcal{O(V + E)}$ for $V$ nodes and $E$ edges. In related work, NAS-Bench-ASR includes removing unreachable branches in the same manner as we do for NB201 \citep{nbasr}.

\FloatBarrier
\subsection{Zero-cost proxies}
\label{app-zcp}
Here, we will provide an overview of the used zero-cost proxies. We use 13 zero-cost proxies from NB-Suite-Zero in our work: \texttt{epe\_nas}, \texttt{fisher}, \texttt{flops}, \texttt{grad\_norm}, \texttt{grasp}, \texttt{jacov}, \texttt{l2\_norm}, \texttt{nwot}, \texttt{params}, \texttt{plain}, \texttt{snip}, \texttt{synflow}, \texttt{zen}.
In general, these proxies can be differentiated into two distinct groups: data-independent, and data-dependent. Within these groups, there are different types of proxies, e.g., jacobian-based zero-cost proxies. 

The data-independent group contains proxies that ignore the downstream dataset (for example \texttt{CIFAR-10}) entirely. Especially, so-called \textit{baseline} proxies, fall into that group. These proxies are based on basic network information, such as the number of parameters (\texttt{params}) \cite{lightweight}, or the sum of the weight's L2-norm \texttt{l2-norm} \cite{NingTLZLYW21}. In addition to these two baseline proxies, \texttt{synflow} \cite{TanakaKYG20}, a pruning-at-initialization based score multiplying all weights in the network, was also successfully used as a zero-cost proxy \cite{lightweight}.
The last data-independent zero-cost proxy is the \texttt{zen-score} \cite{LinWSCS00021}, which approximates the neural network by piecewise linear functions conditioned on activation patterns. 

The majority of zero-cost proxies however is data-dependent; note, the data are not used to update the weights but only for score calculation using mostly one mini-batch. \citet{MellorTSC21} used heuristics based on the Jacobian of the network to calculate zero-cost proxies, resulting in \texttt{jacov} and \texttt{nwot}. The former measures the covariance of the Jacobian, whereas the latter calculates the number of active linear regions in the network. Based on that, \texttt{epe-nas} was introduced \cite{LopesAA21}, which measures ability to distinguish different classes based on the correlation matrix of the Jacobian.  Alternatively, \texttt{grad-norm} \cite{lightweight} sums the Euclidean norm of the gradients. 
In addition to these Jacobian-based scores, there exists several different techniques based on the pruning-at-initialization literature using data. \citet{LeeAT19} introduced \texttt{snip}, which approximates the change in loss. \citet{WangZG20} propose a technique. \texttt{grasp}, which approximates the change in gradient norm. The last pruning-at-initialization based technique \texttt{fisher} \cite{TurnerCOSG20}, which is defined by the sum of the gradients of the network's activation. These techniques were used in \cite{lightweight} as zero-cost proxies. 
Also natural network baselines can be data-dependent; \texttt{plain}, which is the multiplication of the weights and its gradients, and \texttt{flops} are both used as zero-cost proxies in \cite{lightweight}.

\FloatBarrier
\subsection{Encodings}
\label{app-enc-list}
Throughout the paper, we work with these additional encodings: one-hot encoding (onehot), path encoding, arch2vec, NASBOWL's Weisfeiler-Lehman features, and ZCP.

\paragraph{One-hot and path encoding}
One-hot encoding is composed from the one-hot encoding of operations and a flattened adjacency matrix. Path encoding has the size of all possible paths in a cell from the search space -- if the path is present, the corresponding index is 1, otherwise 0. For example, the cell on the right in Figure \ref{fig-nb201-unreachables} would have 1 at positions corresponding to a) conv3x3, b) zero, and c) zero-zero. Both encodings were studied in an encoding study by \citet{encodings}. The path-encoding combined with BANANAS was shown to have a very good performance on NB101 \citep{white2019bananas}.

\paragraph{arch2vec}
Arch2vec is a graph neural network autoencoder model \citep{arch2vec}. It contains a GIN encoder \citep{gincite} and inner product matrix and MLP operations decoder. It enables us to extract latent representations of the neural graphs.

\paragraph{NASBOWL}
We use NASBOWL's Weisfeiler-Lehman kernel features -- features based on isomorphism test of graphs \citep{nasbowl, wlkernel}. NASBOWL uses them as a part of a Gaussian process predictor, which we do not use in this work.

\FloatBarrier
\section{ZCP Biases -- More Results}
\label{moar-biases}
We present additional results on ZCP dependence on the number of convolutions

\begin{figure}[h]
\begin{center}
    \begin{subfigure}[t]{0.33\textwidth}
        \centering
        \includegraphics[width=\textwidth]{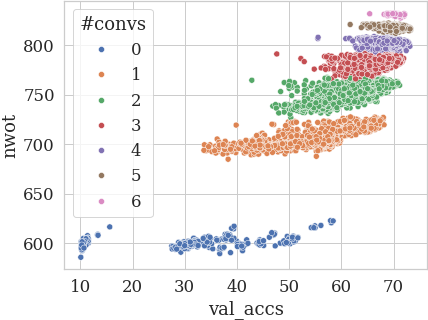}
        \caption{NB201 \texttt{cifar-100} -- nwot}
        \label{fig-c100-nwot}
    \end{subfigure}%
    \begin{subfigure}[t]{0.33\textwidth}
        \centering
        \includegraphics[width=\textwidth]{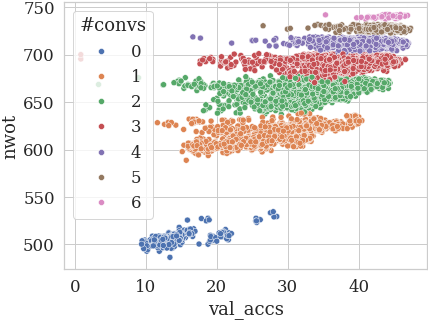}
        \caption{NB201 \texttt{ImageNet16-120} -- nwot}
    \end{subfigure}%
    \begin{subfigure}[t]{0.33\textwidth}
        \centering
        \includegraphics[width=\textwidth]{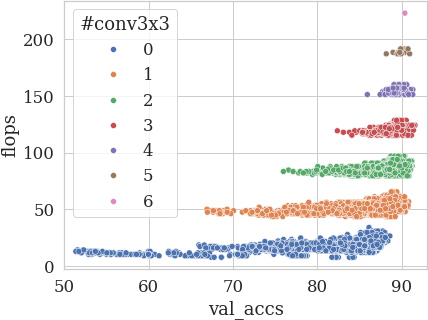}
        \caption{NB201 \texttt{cifar-10} -- \texttt{flops} by \#conv3x3}
    \end{subfigure}

\vskip -0.1in
\caption{Additional results of proxy dependence on the number of layers (left, middle) -- \# convs, right -- \#conv3x3.}
\label{fig-imgnet-ncovns}
\end{center}
\end{figure}

Figure \ref{fig-imgnet-ncovns} shows more results for NB201 -- for \texttt{cifar-100} and \texttt{ImageNet16-120}, we see a similar dependence of \texttt{nwot} on the number of convolutions. For \texttt{flops}, we instead see a dependence on the number of conv3x3.

\begin{figure}[h]
\begin{center}
    \begin{subfigure}[t]{0.5\textwidth}
        \centering
        \includegraphics[width=0.7\textwidth]{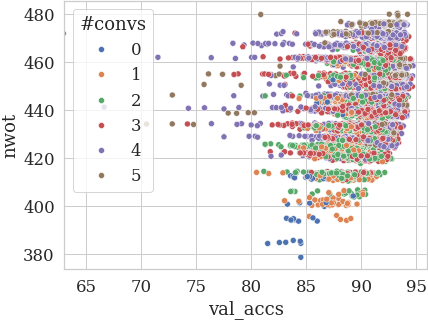}
        \caption{NB101 \texttt{cifar-10} -- nwot}
        \label{fig-nb101-nwot}
    \end{subfigure}%
    \begin{subfigure}[t]{0.5\textwidth}
        \centering
        \includegraphics[width=0.73\textwidth]{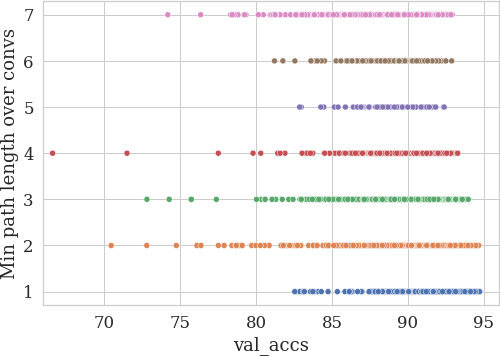}
        \caption{Min path over convs}
        \label{fig-nb101-convs}
    \end{subfigure}

\vskip -0.1in
\caption{For \texttt{NB101}, the dependence on the number of convolutions is not so simple as in NB201 (left). From the features, min path length over convolutions has all best-performing networks with lengths 1 or 2.}
\label{fig-nb101-proxcorr}
\end{center}
\end{figure}

For NB101, the dependence on the number of convolution for NB101 is not simply observable, although we could possibly find some dependence based on other features and the networks size (Figure \ref{fig-nb101-nwot}).
We can observe a different interesting property -- NB101 has all the best-performing networks with min path length over conv1x1 and conv3x3 equal to 1 or 2 (Figure \ref{fig-nb101-convs}) -- and in fact, Table \ref{tab:featimp1} shows that it is the most performing feature on \texttt{cifar-10}.

\begin{figure}[h]
\begin{center}
    \begin{subfigure}[t]{0.28\textwidth}
        \centering
        \includegraphics[width=\textwidth]{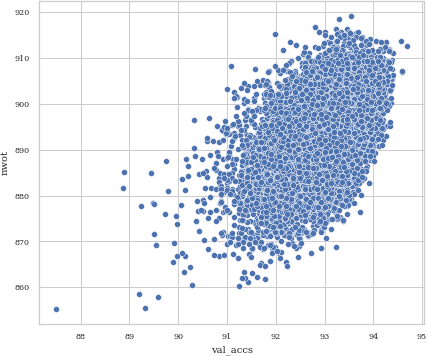}
        \caption{NB301 \texttt{cifar-10} -- nwot}
        \label{fig-nb301-nwott}
    \end{subfigure}%
    \begin{subfigure}[t]{0.28\textwidth}
        \centering
        \includegraphics[width=\textwidth]{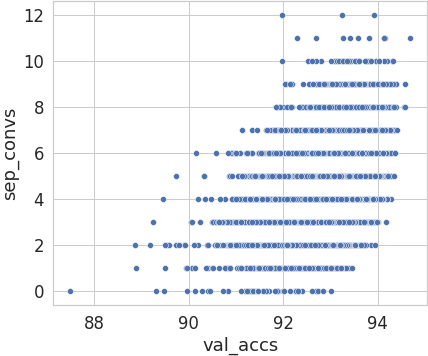}
        \caption{NB301 \#sep\_conv}
    \end{subfigure}%
    \begin{subfigure}[t]{0.36\textwidth}
        \centering
        \includegraphics[width=\textwidth]{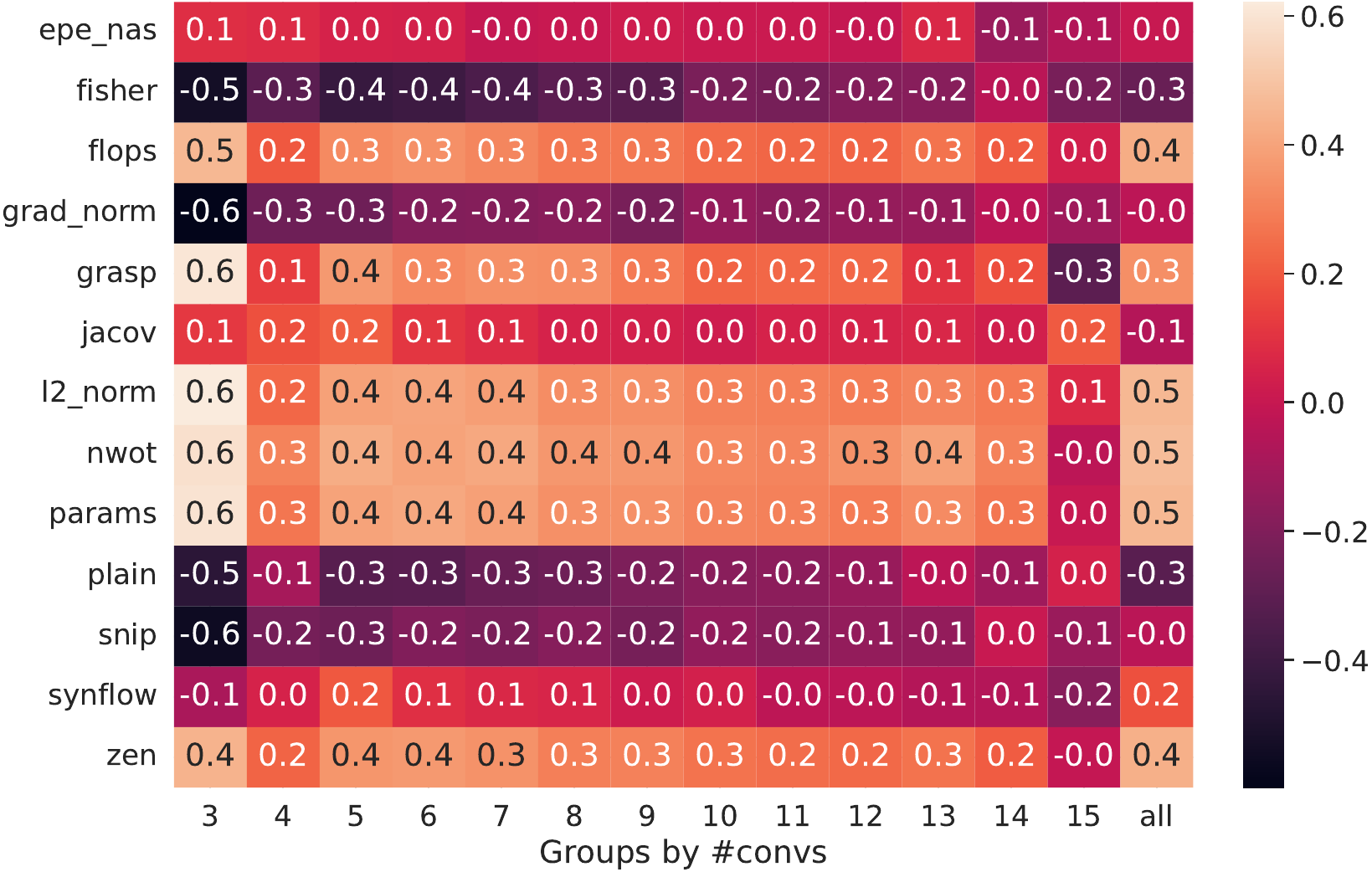}
        \caption{NB201 \texttt{cifar-10} -- \texttt{flops} by \#conv3x3}
    \end{subfigure}

\vskip -0.1in
\caption{NB301 does not show direct dependence on \#convs, however, \texttt{nwot} (left) has only a slightly better correlation than \#sep\_conv (middle), and it is the only proxy with a better correlation (right)}
\label{fig-nb301-proxcorr}
\end{center}
\end{figure}

NB301 does not have a direct dependence on the number of convolutions, but \#conv (sum of all sep\_convs and dil\_convs) is a stronger proxy than all ZCP except \texttt{nwot}, which is slightly better (Figure \ref{fig-nb301-proxcorr}).

\FloatBarrier
\section{Details about GRAF and Models}
\label{app-grafdetails}
We list the total feature count and GRAF computation time for each of the benchmarks in Table \ref{tab:countsandtimes}.

\begin{table}[h]
  \caption{Numbers of GRAF features and time needed for GRAF computation (average
    of 10 evaluations, across all available networks from the benchmark samples) on AMD Ryzen 7 3800X.}
    \label{tab:countsandtimes}
\tabcolsep=4pt
\centering
\small
\begin{tabular}{lrrrrr}
\toprule
&  NB101 &  NB201 &   NB301 &  TNB101 &  TNB101 \\
&        &        &         &   -micro & - macro \\ 
\midrule
count &   47  &  191 &  1540  &          95 &         16 \\
time  &    6.1~s &   27.3~s &   427.1~s &           3.6~s &          0.65~s \\
\bottomrule
\end{tabular}
\vskip -0.1 in
\end{table}

For min and max path lengths, it can happen that no path exists, e.g. when none of the network operations are in the set of allowed operations for a feature. Then, the value would be defined as infinity. To avoid too large numbers in feature columns, we set the value to $|V'| + 1$, where $|V'|$ is the maximum number of nodes in a cell from the search space. The other features are defined in all cases (when the set of allowed nodes is restrictive, the feature value is 0).

For most tasks, we use a random forest model with default hyperparameters from scikit-learn \citep{sklearn}. When comparing with other predictors, we also include an XGBoost \citep{xgboost} with non-default hyperparameters (AutoGluon \citep{agtabular} default parameters with slight tuning), denoted XGB+, see Table \ref{xgb+params}.

\begin{table}[h]
    \centering
    \caption{XGB+ hyperparameters\label{xgb+params}}
    \label{tab:my_label}
    \begin{tabular}{l|r}
    tree\_method & hist \\
    subsample & 0.9 \\
    n\_estimators & 10000 \\
    learning\_rate & 0.01 \\
    \end{tabular}
\end{table}

\FloatBarrier
\subsection{Macro features}
\label{app-GRAF-macro}
TNB101-macro is a linear search space of 4-6 modules with four different module types -- normal, downsampling (strided), channels increasing, and strided + channel increasing. We introduce two classes of features:

\begin{itemize}
    \item Total number of channel increases/strided convolutions until position $i$.
    \item Total number of a module type in the architecture. 
\end{itemize}

The motivation is that since the four types are only one-hot encoded, the model would need to learn which index is strided and which index increases channels. Including the information removes this need. Similarly, the number of increases/strides until a specific position reflects the behavior of input image processing in the network.

\clearpage
\section{GRAF Validation Accuracy -- Full Results}
\label{appendix-val-accs}

For the experiments, we have chosen the random forest predictor (with default scikit-learn hyperparameters) due to its fast fitting time. For all benchmarks and sample sizes, we used 18 CPU hours on Intel Xeon CPU E5-2620.

\begin{figure}[h]
\includegraphics{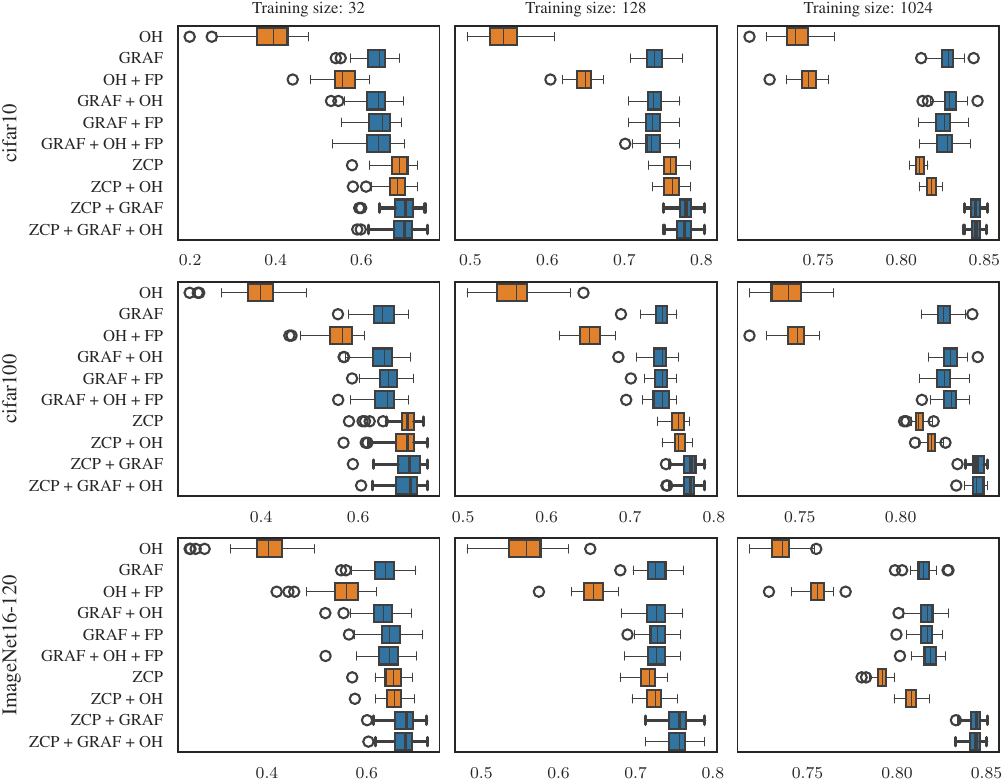}
\caption{NAS-Bench201, accuracy\label{nb201-boxes-full}}
\end{figure}

\begin{figure}[h]
\includegraphics{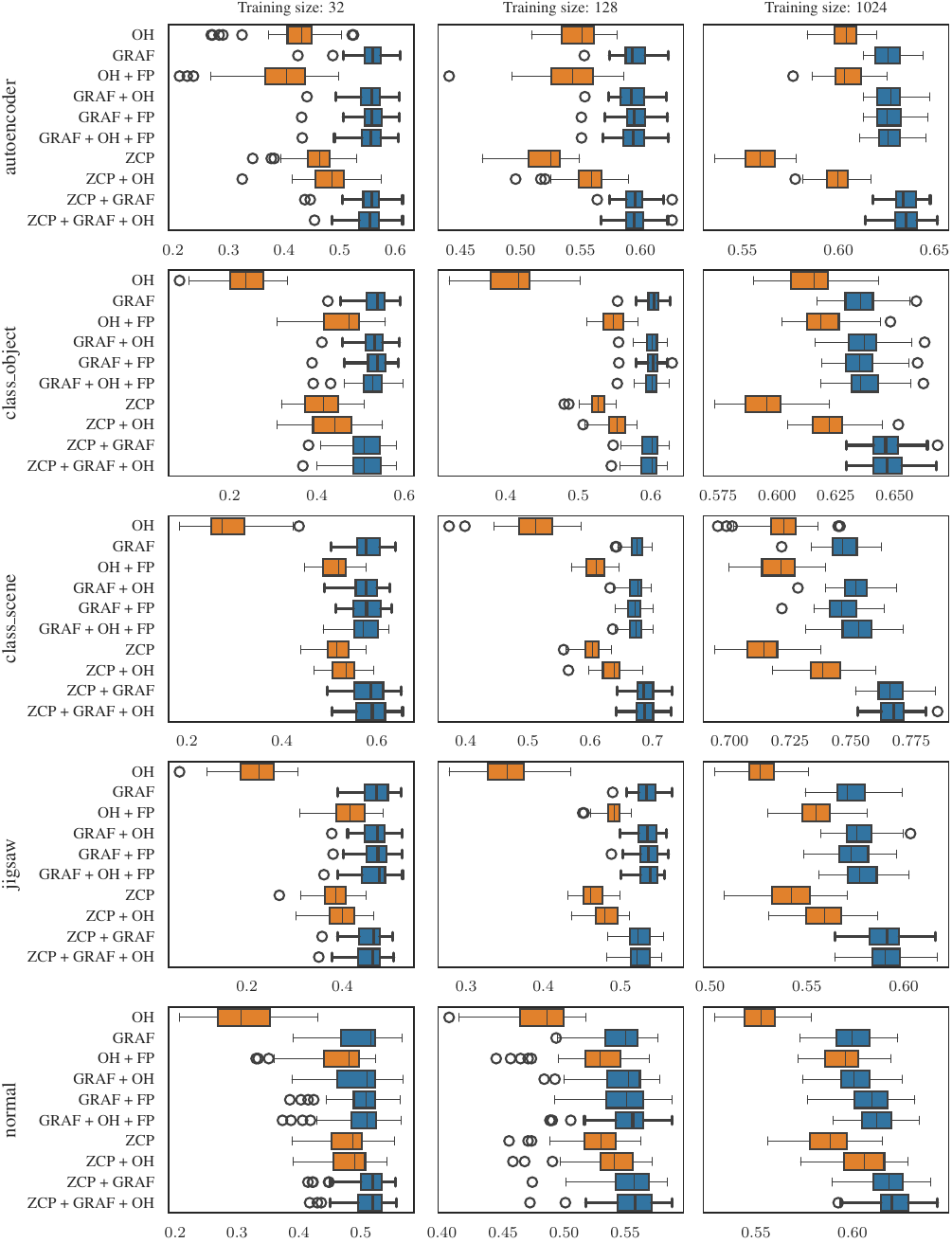}
\caption{Trans-NAS-Bench101-micro, different targets}
\label{app-tnbmicro-1}
\end{figure}

\begin{figure}[h]
\includegraphics{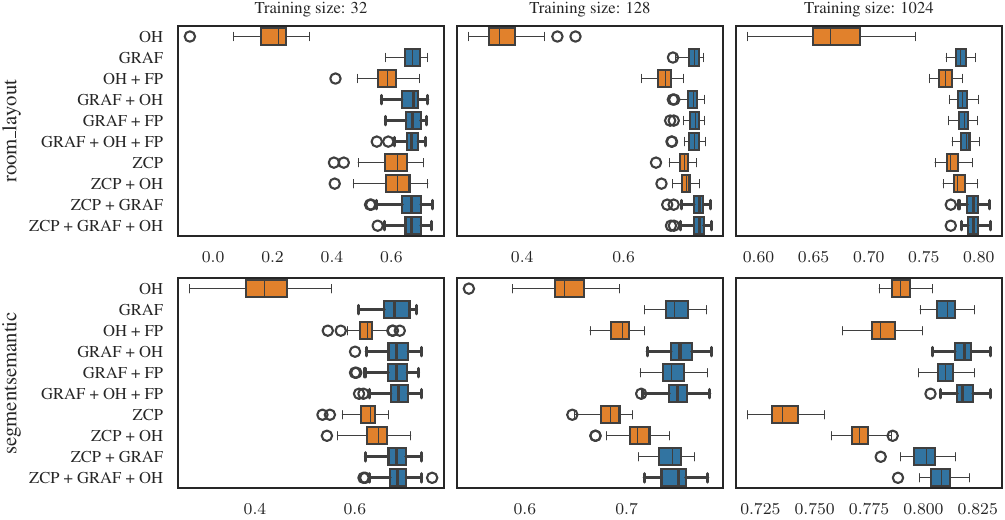}
\caption{Trans-NAS-Bench101-micro, different targets}
\label{app-tnbmicro-2}
\end{figure}

\begin{figure}[h]
\includegraphics{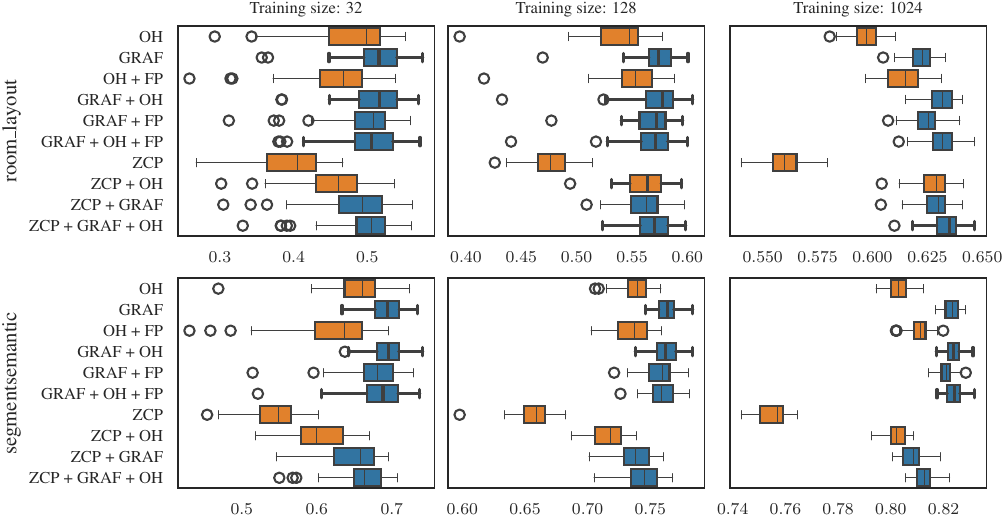}
\caption{Trans-NAS-Bench101-macro, different targets}
\label{app-tnbmacro-1}
\end{figure}

\begin{figure}[h]
\includegraphics{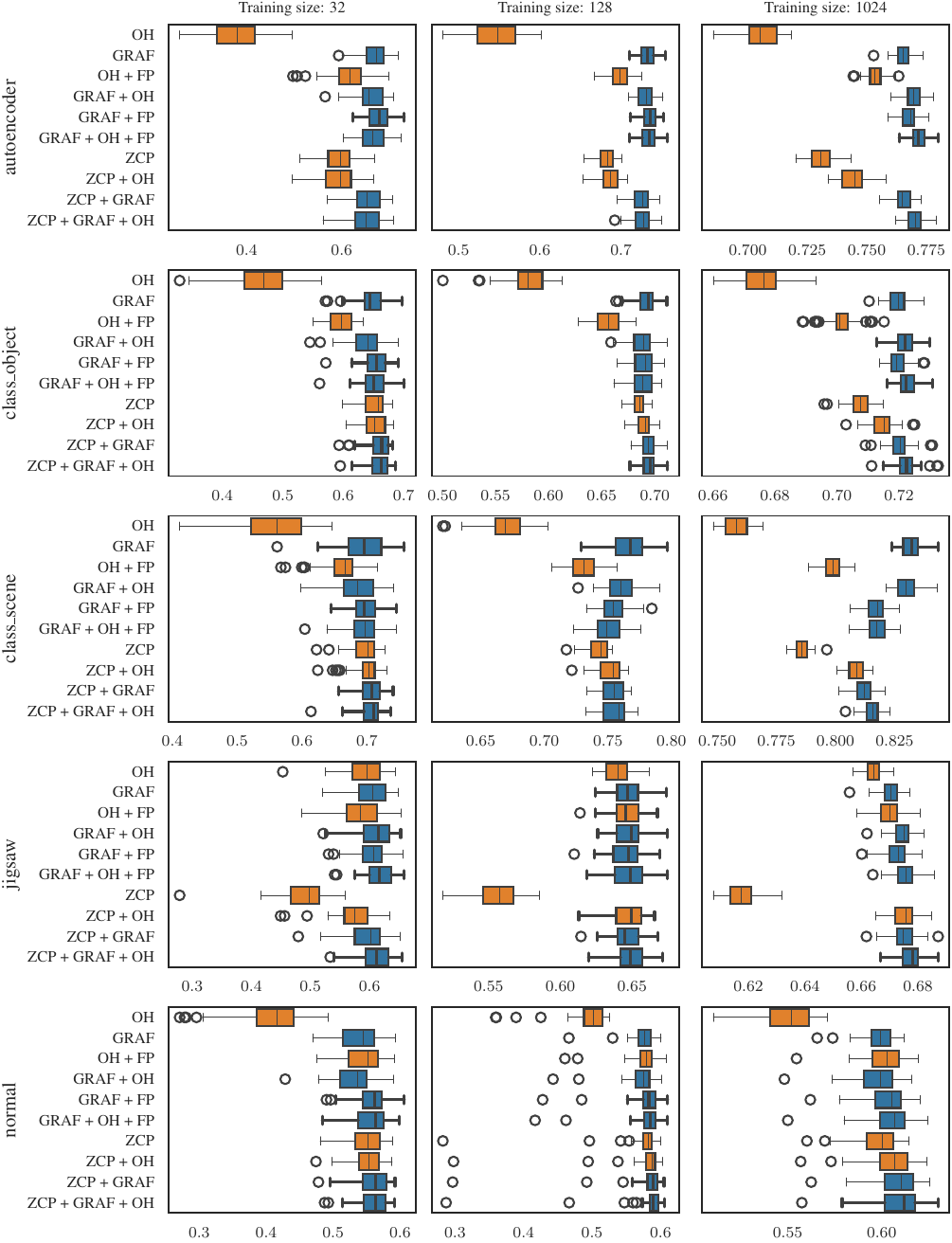}
\caption{Trans-NAS-Bench101-macro, different targets}
\label{app-tnbmacro-2}
\end{figure}

Figures \ref{nb201-boxes-full}--\ref{app-tnbmacro-2} and Tables \ref{acc-nb101-table}--\ref{acc-nb301-table} show full results of the validation accuracy prediction task.

\FloatBarrier
\begin{table}[h]
    \small
    \addtolength{\tabcolsep}{-0.2em}
    \centering
    \caption{Results for predicting validation accuracy on cifar10 on the NB101 benchmark. Average and standard deviation of Kendall tau over 50 independent runs. Bold values indicate the best results and those that are not statistically different from them.\label{acc-nb101-table}}
    \vskip 0.15in
    \begin{tabular}{r|ccc}
dataset & \multicolumn{3}{|c}{cifar10} \\
train\_size & 32 & 128 & 1024 \\
\hline
OH & $0.24^{0.09}$ & $0.40^{0.06}$ & $0.60^{0.02}$ \\
OH + FP & $0.32^{0.09}$ & $0.45^{0.06}$ & $0.60^{0.02}$ \\
GRAF & $0.54^{0.08}$ & $0.63^{0.04}$ & $0.71^{0.01}$ \\
GRAF + FP & $0.53^{0.08}$ & $0.63^{0.04}$ & $0.71^{0.01}$ \\
GRAF + OH & $0.52^{0.08}$ & $0.63^{0.04}$ & $0.71^{0.01}$ \\
GRAF + OH + FP & $0.53^{0.08}$ & $0.63^{0.04}$ & $0.71^{0.01}$ \\
\hline
ZCP & $0.48^{0.05}$ & $0.56^{0.02}$ & $0.63^{0.01}$ \\
ZCP + OH & $0.47^{0.05}$ & $0.57^{0.02}$ & $0.65^{0.01}$ \\
ZCP + GRAF & $\mathbf{0.56^{0.07}}$ & $\mathbf{0.66^{0.04}}$ & $0.73^{0.01}$ \\
ZCP + GRAF + OH & $\mathbf{0.55^{0.07}}$ & $\mathbf{0.66^{0.04}}$ & $0.73^{0.01}$ \\
\hline
OH + PE & $0.32^{0.07}$ & $0.50^{0.04}$ & $0.66^{0.01}$ \\
OH + FP + PE & $0.39^{0.06}$ & $0.52^{0.04}$ & $0.66^{0.01}$ \\
GRAF + PE & $0.54^{0.07}$ & $0.64^{0.03}$ & $0.71^{0.01}$ \\
GRAF + FP + PE & $0.54^{0.07}$ & $0.64^{0.03}$ & $0.71^{0.01}$ \\
GRAF + OH + PE & $0.52^{0.07}$ & $0.64^{0.03}$ & $0.72^{0.01}$ \\
GRAF + OH + FP + PE & $0.53^{0.07}$ & $0.63^{0.03}$ & $0.72^{0.01}$ \\
\hline
ZCP + PE & $0.49^{0.05}$ & $0.58^{0.03}$ & $0.67^{0.01}$ \\
ZCP + OH + PE & $0.48^{0.05}$ & $0.59^{0.03}$ & $0.67^{0.01}$ \\
ZCP + GRAF + PE & $\mathbf{0.56^{0.07}}$ & $\mathbf{0.67^{0.03}}$ & $\mathbf{0.74^{0.01}}$ \\
ZCP + GRAF + OH + PE & $\mathbf{0.55^{0.07}}$ & $\mathbf{0.67^{0.03}}$ & $\mathbf{0.74^{0.01}}$ \\
\end{tabular}
    \end{table}%
\begin{table}[h]
    \small
    \addtolength{\tabcolsep}{-0.2em}
    \centering
    \caption{Results for predicting validation accuracy on different datasets on the NB201 benchmark. Average and standard deviation of Kendall tau over 50 independent runs. Bold values indicate the best results and those that are not statistically different from them.\label{app-nb201-boxes}}
    \vskip 0.15in
    \begin{tabular}{r|ccc|ccc|ccc}
dataset & \multicolumn{3}{|c}{cifar10} & \multicolumn{3}{|c}{cifar100} & \multicolumn{3}{|c}{ImageNet16-120} \\
train\_size & 32 & 128 & 1024 & 32 & 128 & 1024 & 32 & 128 & 1024 \\
\hline
OH & $0.39^{0.06}$ & $0.54^{0.03}$ & $0.74^{0.01}$ & $0.39^{0.05}$ & $0.56^{0.03}$ & $0.74^{0.01}$ & $0.40^{0.06}$ & $0.56^{0.03}$ & $0.74^{0.01}$ \\
OH + FP & $0.56^{0.04}$ & $0.65^{0.02}$ & $0.74^{0.01}$ & $0.56^{0.04}$ & $0.65^{0.02}$ & $0.75^{0.01}$ & $0.55^{0.04}$ & $0.64^{0.02}$ & $0.75^{0.01}$ \\
GRAF & $0.63^{0.03}$ & $0.74^{0.01}$ & $0.83^{0.01}$ & $0.65^{0.03}$ & $0.74^{0.01}$ & $0.82^{0.01}$ & $0.63^{0.03}$ & $0.73^{0.02}$ & $0.81^{0.01}$ \\
GRAF + FP & $0.64^{0.04}$ & $0.74^{0.01}$ & $0.83^{0.01}$ & $0.66^{0.03}$ & $0.74^{0.01}$ & $0.82^{0.01}$ & $0.64^{0.03}$ & $0.73^{0.02}$ & $0.82^{0.01}$ \\
GRAF + OH & $0.63^{0.04}$ & $0.74^{0.01}$ & $0.83^{0.01}$ & $0.65^{0.03}$ & $0.74^{0.01}$ & $0.82^{0.01}$ & $0.63^{0.03}$ & $0.73^{0.02}$ & $0.82^{0.01}$ \\
GRAF + OH + FP & $0.63^{0.04}$ & $0.74^{0.02}$ & $0.83^{0.01}$ & $0.65^{0.03}$ & $0.74^{0.01}$ & $0.82^{0.00}$ & $0.64^{0.03}$ & $0.73^{0.02}$ & $0.82^{0.01}$ \\
\hline
ZCP & $0.69^{0.03}$ & $0.76^{0.01}$ & $0.81^{0.00}$ & $\mathbf{0.69^{0.03}}$ & $0.76^{0.01}$ & $0.81^{0.00}$ & $0.65^{0.02}$ & $0.72^{0.01}$ & $0.79^{0.00}$ \\
ZCP + OH & $0.68^{0.03}$ & $0.76^{0.01}$ & $0.82^{0.00}$ & $\mathbf{0.69^{0.04}}$ & $0.76^{0.01}$ & $0.82^{0.00}$ & $0.65^{0.02}$ & $0.72^{0.01}$ & $0.81^{0.00}$ \\
ZCP + GRAF & $\mathbf{0.70^{0.03}}$ & $\mathbf{0.78^{0.01}}$ & $0.84^{0.00}$ & $\mathbf{0.70^{0.03}}$ & $\mathbf{0.77^{0.01}}$ & $0.84^{0.00}$ & $\mathbf{0.67^{0.03}}$ & $\mathbf{0.75^{0.02}}$ & $\mathbf{0.84^{0.00}}$ \\
ZCP + GRAF + OH & $\mathbf{0.69^{0.04}}$ & $\mathbf{0.78^{0.01}}$ & $0.84^{0.00}$ & $\mathbf{0.70^{0.03}}$ & $\mathbf{0.77^{0.01}}$ & $0.84^{0.00}$ & $\mathbf{0.67^{0.02}}$ & $0.75^{0.02}$ & $0.84^{0.00}$ \\
\hline
OH + PE & $0.42^{0.06}$ & $0.57^{0.03}$ & $0.75^{0.01}$ & $0.42^{0.05}$ & $0.58^{0.03}$ & $0.75^{0.01}$ & $0.40^{0.05}$ & $0.56^{0.03}$ & $0.74^{0.01}$ \\
OH + FP + PE & $0.56^{0.04}$ & $0.66^{0.01}$ & $0.76^{0.01}$ & $0.57^{0.04}$ & $0.66^{0.02}$ & $0.76^{0.01}$ & $0.55^{0.04}$ & $0.65^{0.02}$ & $0.76^{0.01}$ \\
GRAF + PE & $0.63^{0.03}$ & $0.74^{0.01}$ & $0.83^{0.01}$ & $0.65^{0.03}$ & $0.74^{0.01}$ & $0.82^{0.01}$ & $0.63^{0.04}$ & $0.73^{0.02}$ & $0.82^{0.01}$ \\
GRAF + FP + PE & $0.64^{0.03}$ & $0.74^{0.01}$ & $0.83^{0.01}$ & $0.66^{0.03}$ & $0.74^{0.01}$ & $0.82^{0.01}$ & $0.64^{0.03}$ & $0.73^{0.02}$ & $0.82^{0.01}$ \\
GRAF + OH + PE & $0.63^{0.04}$ & $0.74^{0.02}$ & $0.83^{0.01}$ & $0.65^{0.03}$ & $0.74^{0.01}$ & $0.83^{0.01}$ & $0.63^{0.04}$ & $0.72^{0.02}$ & $0.82^{0.01}$ \\
GRAF + OH + FP + PE & $0.64^{0.04}$ & $0.74^{0.02}$ & $0.83^{0.01}$ & $0.66^{0.03}$ & $0.74^{0.01}$ & $0.82^{0.01}$ & $0.64^{0.03}$ & $0.73^{0.02}$ & $0.82^{0.01}$ \\
\hline
ZCP + PE & $0.68^{0.03}$ & $0.76^{0.01}$ & $0.82^{0.00}$ & $\mathbf{0.69^{0.03}}$ & $0.76^{0.01}$ & $0.82^{0.00}$ & $0.65^{0.02}$ & $0.72^{0.01}$ & $0.80^{0.00}$ \\
ZCP + OH + PE & $0.68^{0.03}$ & $0.76^{0.01}$ & $0.82^{0.00}$ & $\mathbf{0.69^{0.04}}$ & $0.76^{0.01}$ & $0.82^{0.00}$ & $0.65^{0.02}$ & $0.72^{0.01}$ & $0.81^{0.00}$ \\
ZCP + GRAF + PE & $\mathbf{0.69^{0.04}}$ & $\mathbf{0.78^{0.01}}$ & $\mathbf{0.85^{0.00}}$ & $\mathbf{0.70^{0.03}}$ & $\mathbf{0.77^{0.01}}$ & $\mathbf{0.84^{0.00}}$ & $\mathbf{0.67^{0.03}}$ & $\mathbf{0.75^{0.02}}$ & $\mathbf{0.84^{0.00}}$ \\
ZCP + GRAF + OH + PE & $\mathbf{0.69^{0.03}}$ & $\mathbf{0.78^{0.01}}$ & $\mathbf{0.85^{0.00}}$ & $\mathbf{0.70^{0.03}}$ & $\mathbf{0.77^{0.01}}$ & $\mathbf{0.84^{0.00}}$ & $\mathbf{0.67^{0.02}}$ & $0.75^{0.02}$ & $\mathbf{0.84^{0.00}}$ \\
\end{tabular}
    \end{table}%
\begin{table}[h]
    \small
    \addtolength{\tabcolsep}{-0.2em}
    \centering
    \caption{Results for predicting validation accuracy on cifar10 on the NB301 benchmark. Average and standard deviation of Kendall tau over 50 independent runs. Bold values indicate the best results and those that are not statistically different from them.\label{acc-nb301-table}}
    \vskip 0.15in
    \begin{tabular}{r|ccc}
dataset & \multicolumn{3}{|c}{cifar10} \\
train\_size & 32 & 128 & 1024 \\
\hline
OH & $0.19^{0.05}$ & $0.32^{0.03}$ & $0.48^{0.02}$ \\
OH + FP & $0.27^{0.06}$ & $0.38^{0.02}$ & $0.48^{0.01}$ \\
GRAF & $0.38^{0.05}$ & $0.52^{0.03}$ & $0.61^{0.01}$ \\
GRAF + FP & $0.39^{0.05}$ & $0.53^{0.03}$ & $0.63^{0.01}$ \\
GRAF + OH & $0.39^{0.05}$ & $0.52^{0.03}$ & $0.61^{0.01}$ \\
GRAF + OH + FP & $0.39^{0.05}$ & $0.53^{0.03}$ & $0.63^{0.01}$ \\
\hline
ZCP & $0.28^{0.06}$ & $0.34^{0.02}$ & $0.38^{0.00}$ \\
ZCP + OH & $0.31^{0.06}$ & $0.40^{0.02}$ & $0.48^{0.01}$ \\
ZCP + GRAF & $\mathbf{0.40^{0.05}}$ & $0.53^{0.03}$ & $0.64^{0.01}$ \\
ZCP + GRAF + OH & $\mathbf{0.40^{0.05}}$ & $\mathbf{0.53^{0.03}}$ & $0.64^{0.01}$ \\
\hline
OH + PE & $0.23^{0.05}$ & $0.40^{0.04}$ & $0.55^{0.02}$ \\
OH + FP + PE & $0.31^{0.07}$ & $0.47^{0.04}$ & $0.59^{0.01}$ \\
GRAF + PE & $0.38^{0.05}$ & $0.52^{0.03}$ & $0.61^{0.01}$ \\
GRAF + FP + PE & $0.39^{0.05}$ & $0.53^{0.03}$ & $0.64^{0.01}$ \\
GRAF + OH + PE & $0.39^{0.05}$ & $0.52^{0.03}$ & $0.61^{0.01}$ \\
GRAF + OH + FP + PE & $0.39^{0.05}$ & $0.53^{0.03}$ & $0.64^{0.01}$ \\
\hline
ZCP + PE & $0.33^{0.08}$ & $0.46^{0.04}$ & $0.58^{0.00}$ \\
ZCP + OH + PE & $0.33^{0.07}$ & $0.47^{0.04}$ & $0.59^{0.00}$ \\
ZCP + GRAF + PE & $\mathbf{0.40^{0.05}}$ & $\mathbf{0.53^{0.03}}$ & $\mathbf{0.64^{0.01}}$ \\
ZCP + GRAF + OH + PE & $\mathbf{0.40^{0.05}}$ & $\mathbf{0.53^{0.03}}$ & $\mathbf{0.64^{0.01}}$ \\
\end{tabular}
    \end{table}%

\FloatBarrier
\section{Feature Importances}
\label{app-featimp}

Now we list additional results for Section~\ref{sec:featimp}. The Tables~\ref{tab:featimp1}--\ref{tab:featimpx} list the top ten important features for a given task and benchmark. We can observe a great variability across tasks and search space.

For the macro search spaces (Tables \ref{featimp-first-macro}--\ref{tab:featimpx}), an interesting observation is that the number of strides until pos. 0 (effectively \emph{at} position 0) is often the most important feature. This indicates that the predictor benefits from the information about which operation is strided and which increases channels.

\begin{table*}[h]
\caption{The ten most important features on NB201 \texttt{ImageNet16-120} and TNB101-micro \texttt{class\_scene}  from ranking based on Shapley values computed from random forest trained on 1024 training samples with ZCP+GRAF.}
\label{featimp-tnb101img}
\vskip 0.1in
\centering
\small
\begin{tabular}{lrllr}
\hline
\multicolumn{2}{c}{NB201 - 
  \texttt{ImageNet16-120}} && \multicolumn{2}{c}{TNB101-micro - \texttt{class\_scene}} \\

             Feature name &  Mean rank & $\quad$ &                         Feature name &  Mean rank \\
\hline
                              jacov &    0.00  &  &                               params &       0.28 \\
     min path over [skip,C3x3,C1x1] &    2.30  &  &                                flops &       0.72 \\
                               nwot &    2.30  &  &              Average out deg. - C3x3 &       2.48 \\
                             params &    3.56  &  &  Average out deg. - [zero,skip,C1x1] &       2.52 \\
                            synflow &    4.08  &  &   Average in deg. - [zero,skip,C1x1] &       4.32 \\
                              flops &    4.30  &  &               Average in deg. - C3x3 &       4.68 \\
           min path over [skip,C1x1] &    5.86  &  &                       number of C3x3 &       6.00 \\
                                zen &    6.56  &  &                              synflow &       7.30 \\
                             fisher &    9.40  &  &      Input node degree - [zero,C1x1] &       9.34 \\
                          grad\_norm &   11.56  & &  Input node degree - [zero,C1x1,C3x3] &       9.42 \\
\hline
\end{tabular}
\end{table*}

\begin{table}[hb]
\caption{The ten most important features on NB101 and NB301 (\texttt{cifar-10} dataset) from ranking based on Shapley values computed from random forest trained on 1024 training samples with ZCP+GRAF. \label{tab:featimp1}}
\tabcolsep=2pt
\begin{tabular}{lrllr}
\hline
  \multicolumn{2}{c}{NB101 - \texttt{cifar-10}} && \multicolumn{2}{c}{NB301 - \texttt{cifar-10}} \\
                  Feature name &  Mean rank & $\quad$ &                                          Feature name &  Mean rank \\
\hline
     min path over [C1x1,C3x3] &       2.16 &        &                                                  nwot &       2.02 \\
                       l2\_norm &       2.16 &        &                                                params &       5.16 \\
                           zen &       2.38 &        &     max path over [MP3x3,AP3x3]from input 2  (normal) &       7.68 \\
                        fisher &       3.72 &        &                   Input node 2 degree - skip (normal) &       9.00 \\
 Average input node degreeC3x3 &       5.42 &        &         Average in deg. - [skip,SC3x3,SC5x5] (normal) &       9.70 \\
Average output node degreeC3x3 &       5.48 &        &                                                fisher &      10.76 \\
                number of C3x3 &       6.06 &        &          Input 1 degree - [MP3x3,skip,SC3x3] (normal) &      11.58 \\
                         grasp &       8.26 &        &                                               l2\_norm &      11.58 \\
    Output node degree - MP3x3 &       9.72 &        &        Average out deg. - [skip,SC3x3,SC5x5] (normal) &      11.60 \\
                         plain &       9.88 &        & Average out deg. - [MP3x3,AP3x3,DC3x3,DC5x5] (normal) &      12.10 \\
\hline
\end{tabular}
\end{table}

\begin{table}[hb]
\caption{The ten most important features on TNB101-micro (\texttt{class\_object} and \texttt{normal} datasets) from ranking based on Shapley values computed from random forest trained on 1024 training samples with ZCP+GRAF.}
\centering
\begin{tabular}{lrllr}
\hline
  \multicolumn{2}{c}{TNB101-micro - \texttt{class\_object}} && \multicolumn{2}{c}{TNB301-micro - \texttt{normal}} \\
                        Feature name &  Mean rank & $\quad$ &                         Feature name &  Mean rank \\
\hline
                              params &       0.36 &        &                                jacov &       0.02 \\
                               flops &       0.78 &        &                                flops &       0.98 \\
 Average out deg. - [zero,skip,C1x1] &       2.68 &        &   Average in deg. - [zero,skip,C1x1] &       4.22 \\
             Average out deg. - C3x3 &       2.72 &        &  Average out deg. - [zero,skip,C1x1] &       4.64 \\
  Average in deg. - [zero,skip,C1x1] &       4.58 &        &              Average out deg. - C3x3 &       4.72 \\
              Average in deg. - C3x3 &       5.00 &        &                               params &       5.18 \\
                  min path over skip &       5.02 &        &                       number of C3x3 &       5.34 \\
                      number of C3x3 &       6.88 &        &               Average in deg. - C3x3 &       5.44 \\
            Input node degree - skip &       9.08 &        &             Input node degree - C3x3 &       8.68 \\
Input node degree - [zero,C1x1,C3x3] &       9.32 &        & Input node degree - [zero,skip,C1x1] &       8.92 \\
\hline
\end{tabular}
\end{table}

\begin{table}[hb]
\caption{The ten most important features on TNB101-micro (\texttt{jigsaw} and \texttt{room\_layout} datasets) from ranking based on Shapley values computed from random forest trained on 1024 training samples with ZCP+GRAF.}
\centering
\begin{tabular}{lrllr}
\hline
  \multicolumn{2}{c}{TNB101-micro - \texttt{jigsaw}} && \multicolumn{2}{c}{TNB301-micro - \texttt{room\_layout}} \\

                       Feature name &  Mean rank & $\quad$ &                        Feature name &  Mean rank \\
\hline
    Input node degree - [zero,C1x1] &       1.84 &        &                              params &       1.10 \\
                              flops &       2.34 &        &                              fisher &       1.70 \\
                             params &       2.48 &        &                               flops &       1.82 \\
    Input node degree - [skip,C3x3] &       3.74 &        & Average out deg. - [zero,skip,C1x1] &       3.74 \\
            Average out deg. - C3x3 &       6.56 &        &             Average out deg. - C3x3 &       3.86 \\
                              jacov &       6.86 &        &                               jacov &       5.84 \\
Average out deg. - [zero,skip,C1x1] &       7.22 &        &  Average in deg. - [zero,skip,C1x1] &       6.08 \\
 Average in deg. - [zero,skip,C1x1] &       7.46 &        &              Average in deg. - C3x3 &       6.38 \\
           Input node degree - skip &       7.48 &        &                      number of C3x3 &       8.52 \\
             Average in deg. - C3x3 &       7.96 &        &                  min path over skip &       8.56 \\
\hline
\end{tabular}
\end{table}

\begin{table}[hb]
\caption{The ten most important features on TNB101-micro (\texttt{segmentsemantic} dataset) from ranking based on Shapley values computed from random forest trained on 1024 training samples with ZCP+GRAF.}
\centering
\begin{tabular}{lrl}
\hline
  \multicolumn{2}{c}{TNB101-micro - \texttt{segmentsemantic}}  \\
                       Feature name &  Mean rank & $\quad$ \\
\hline
                              flops &       0.60 &        \\
Average out deg. - [zero,skip,C1x1] &       0.60 &        \\
             Average in deg. - C3x3 &       1.96 &        \\
 Average in deg. - [zero,skip,C1x1] &       3.12 &        \\
            Average out deg. - C3x3 &       4.36 &        \\
                             params &       4.60 &        \\
                     number of C3x3 &       5.76 &        \\
                              jacov &       7.68 &        \\
                               snip &       9.54 &        \\
    Input node degree - [skip,C3x3] &       9.86 &        \\
\hline
\end{tabular}
\end{table}

\begin{table}
\caption{The ten most important features on TNB101-macro (\texttt{autoencoder} and \texttt{class\_scene} datasets) from ranking based on Shapley values computed from random forest trained on 1024 training samples with ZCP+GRAF.\label{featimp-first-macro}}
\centering 
\begin{tabular}{lrllr}
\hline
    \multicolumn{2}{c}{TNB101-macro - \texttt{autoencoder}} && \multicolumn{2}{c}{TNB101-macro - \texttt{class\_scene}} \\
                       Feature name &  Mean rank & $\quad$ &                             Feature name &  Mean rank \\
\hline
     Number of strides until pos. 4 &       0.04 &        &                                     nwot &       0.00 \\
     Number of strides until pos. 5 &       1.04 &        &           Number of strides until pos. 0 &       2.10 \\
                              flops &       1.92 &        &                                grad\_norm &       2.36 \\
                             params &       3.00 &        &                                    flops &       3.14 \\
                               nwot &       4.72 &        &                                    grasp &       4.04 \\
number of convs - channel increased &       5.86 &        &                                    jacov &       5.52 \\
     Number of strides until pos. 3 &       5.92 &        &                                     snip &       5.58 \\
             number of simple convs &       8.96 &        & Number of channel increases until pos. 1 &       9.32 \\
                                zen &       9.86 &        &                                  l2\_norm &       9.34 \\
                            l2\_norm &      10.68 &        &                                   fisher &      10.38 \\
\hline
\end{tabular}
\end{table}

\begin{table}[hb]
\caption{The ten most important features on TNB101-macro (\texttt{class\_object} and \texttt{normal} datasets) from ranking based on Shapley values computed from random forest trained on 1024 training samples with ZCP+GRAF.}
\centering
\begin{tabular}{lrllr}
\hline
 \multicolumn{2}{c}{TNB101-macro - \texttt{class\_object}} && \multicolumn{2}{c}{TNB101-macro - \texttt{normal}} \\
                  Feature name &  Mean rank & $\quad$ &                   Feature name &  Mean rank \\
\hline
                          nwot &       0.00 &        &                           nwot &       0.46 \\
Number of strides until pos. 5 &       1.50 &        &                          flops &       0.54 \\
                         grasp &       1.62 &        &                         params &       3.44 \\
Number of strides until pos. 4 &       4.08 &        &                        l2\_norm &       3.72 \\
                         flops &       4.22 &        &                          jacov &       4.76 \\
                     grad\_norm &       4.22 &        & Number of strides until pos. 5 &       6.58 \\
                         jacov &       7.38 &        &                          plain &       8.74 \\
     number of convs - strided &       7.70 &        &                         fisher &       9.14 \\
Number of strides until pos. 0 &       8.62 &        &                            zen &       9.56 \\
                       epe\_nas &      11.22 &        &                      grad\_norm &       9.78 \\
\hline
\end{tabular}
\end{table}

\begin{table}
\caption{The ten most important features on TNB101-macro (\texttt{jigsaw} and \texttt{room\_layout} datasets) from ranking based on Shapley values computed from random forest trained on 1024 training samples with ZCP+GRAF.}
\centering
\begin{tabular}{lrllr}
\hline
 \multicolumn{2}{c}{TNB101-macro - \texttt{jigsaw}} && \multicolumn{2}{c}{TNB101-macro - \texttt{room\_layout}} \\
                            Feature name &  Mean rank & $\quad$ &                             Feature name &  Mean rank \\
\hline
          Number of strides until pos. 0 &       0.00 &        &           Number of strides until pos. 0 &       0.00 \\
Number of channel increases until pos. 0 &       1.00 &        & Number of channel increases until pos. 0 &       1.12 \\
                                    nwot &       2.26 &        &                                     nwot &       2.98 \\
                                   flops &       4.06 &        &                                      zen &       5.66 \\
Number of channel increases until pos. 1 &       5.48 &        &                                   fisher &       5.90 \\
                                   grasp &       5.78 &        &           Number of strides until pos. 1 &       6.14 \\
                                   jacov &       6.86 &        &                                grad\_norm &       6.50 \\
                                  fisher &       8.18 &        &                                  synflow &       8.20 \\
                                   plain &       8.52 &        &                                    jacov &      10.02 \\
               number of convs - strided &       9.54 &        &           Number of strides until pos. 4 &      10.18 \\
\hline
\end{tabular}
\end{table}

\begin{table}[hb]
\caption{The ten most important features on TNB101-macro (\texttt{segmentsemantic} dataset) from ranking based on Shapley values computed from random forest trained on 1024 training samples with ZCP+GRAF. \label{tab:featimpx}}
\centering
\begin{tabular}{lrl}
\hline
\multicolumn{2}{c}{TNB101-macro - \texttt{segmentsemantic}}  \\
                            Feature name &  Mean rank & $\quad$ \\
\hline
          Number of strides until pos. 0 &       0.00 &        \\
                                    nwot &       1.02 &        \\
          Number of strides until pos. 1 &       2.00 &        \\
Number of channel increases until pos. 0 &       2.98 &        \\
               number of convs - strided &       4.00 &        \\
                                   flops &       5.52 &        \\
          Number of strides until pos. 2 &       5.64 &        \\
                                  params &       7.30 &        \\
Number of channel increases until pos. 1 &       8.58 &        \\
          Number of strides until pos. 5 &       9.58 &        \\
\hline
\end{tabular}
\end{table}

\FloatBarrier
\section{Feature Redundancy}
\label{app-featredund}

For the group redundancy, we used 2 CPU hours on Intel Xeon CPU E5-2620.

Table~\ref{ap:tab:redundancy} lists average Kendall tau values comparing accuracy prediction using Random Forest with all GRAF and 
selected linearly independent GRAF. We compare cases with only GRAF and GRAF with ZCP. We denote the independent set of GRAF as sel. GRAF.

When using only GRAF, GRAF outperforms sel. GRAF in all cases except for NB201 and sample size 1024. When ZCP are also included, GRAF and sel. GRAF perform on par on NB101 and NB201 while being worse on NB301. This suggests that ZCP capture the same information as some of the redundant features, but cannot capture all important properties on NB301.

\begin{table}[hb]
\small
  \caption{Comparison of prediction with all GRAF features 
    and selected (linearly independent) GRAF features. Average of Kendall tau over 50 runs. All benchmarks evaluated for \texttt{CIFAR-10}}
  \label{ap:tab:redundancy}
  \centering
\begin{tabular}{l|rrr|rrr|rrr}
benchmark & \multicolumn{3}{c}{NB101} & \multicolumn{3}{c}{NB201} & \multicolumn{3}{c}{NB301} \\
train size &      32   &      128  &      1024 &      32   &      128  &      1024 &      32   &      128  &      1024 \\
features        &           &           &           &           &           &           &           &           &           \\
\hline
GRAF                    &  $0.54^{ 0.08 }$ &  $0.63^{ 0.04 }$ &  $0.71^{ 0.01 }$ &  $0.63^{ 0.03 }$ &  $0.74^{ 0.01 }$ &  $0.83^{ 0.01 }$ &  $0.38^{ 0.05 }$ &  $0.52^{ 0.03 }$ &  $0.61^{ 0.01 }$ \\
sel. GRAF               &  $0.53^{ 0.07 }$ &  $0.62^{ 0.04 }$ &   $0.7^{ 0.01 }$ &  $0.61^{ 0.05 }$ &  $0.72^{ 0.02 }$ &   $0.83^{ 0.0 }$ &  $0.31^{ 0.06 }$ &  $0.44^{ 0.02 }$ &  $0.57^{ 0.01 }$ \\
\hline
GRAF + ZCP              &  $0.56^{ 0.07 }$ &  $0.66^{ 0.04 }$ &  $0.73^{ 0.01 }$ &   $0.7^{ 0.03 }$ &  $0.78^{ 0.01 }$ &   $0.84^{ 0.0 }$ &   $0.4^{ 0.05 }$ &  $0.53^{ 0.03 }$ &  $0.64^{ 0.01 }$ \\
sel. GRAF + ZCP         &  $0.56^{ 0.06 }$ &  $0.66^{ 0.03 }$ &  $0.73^{ 0.01 }$ &   $0.7^{ 0.04 }$ &  $0.78^{ 0.01 }$ &   $0.84^{ 0.0 }$ &  $0.35^{ 0.06 }$ &  $0.48^{ 0.03 }$ &  $0.61^{ 0.01 }$ \\
\hline
\end{tabular} 
\end{table}

We include an additional experiment in Table \ref{data-dep-prox} -- we study how the results differ when GRAF is used only with data-dependent proxies (as per \citep{nbsuitezero}). They seem to capture most of the performance gain, and including \texttt{params} leads to the same performance as with all ZCP.

\begin{table}[h]
\small
    \centering
    \caption{Results of GRAF used along with data-dependent ZCP (DD-ZCP), params (P) and all ZCP (ZCP).\label{data-dep-prox}}
\begin{tabular}{llllllllll}

benchmark & \multicolumn{3}{l}{nb101} & \multicolumn{3}{l}{nb201} & \multicolumn{3}{l}{nb301} \\
train size &             32   &             128  &             1024 &             32   &             128  &             1024 &             32   &             128  &             1024 \\
features                &                  &                  &                  &                  &                  &                  &                  &                  &                  \\
\midrule
GRAF                    &  $0.54^{ 0.08 }$ &  $0.63^{ 0.04 }$ &  $0.71^{ 0.01 }$ &  $0.63^{ 0.03 }$ &  $0.74^{ 0.01 }$ &  $0.83^{ 0.01 }$ &  $0.38^{ 0.05 }$ &  $0.52^{ 0.03 }$ &  $0.61^{ 0.01 }$ \\
GRAF + DD\_ZCP          &  $0.56^{ 0.07 }$ &  $0.66^{ 0.04 }$ &  $0.73^{ 0.01 }$ &  $0.69^{ 0.03 }$ &  $0.78^{ 0.01 }$ &   $0.84^{ 0.0 }$ &   $0.4^{ 0.05 }$ &  $0.53^{ 0.03 }$ &  $0.63^{ 0.01 }$ \\
GRAF + DD\_ZCP + P      &  $0.55^{ 0.07 }$ &  $0.66^{ 0.04 }$ &  $0.73^{ 0.01 }$ &   $0.7^{ 0.04 }$ &  $0.78^{ 0.01 }$ &   $0.84^{ 0.0 }$ &   $0.4^{ 0.05 }$ &  $0.53^{ 0.03 }$ &  $0.64^{ 0.01 }$ \\
GRAF + ZCP              &  $0.56^{ 0.07 }$ &  $0.66^{ 0.04 }$ &  $0.73^{ 0.01 }$ &   $0.7^{ 0.03 }$ &  $0.78^{ 0.01 }$ &   $0.84^{ 0.0 }$ &   $0.4^{ 0.05 }$ &  $0.53^{ 0.03 }$ &  $0.64^{ 0.01 }$ \\
\bottomrule
\end{tabular}
\end{table}

\begin{figure}[h]
\includegraphics[width=0.92\textwidth]{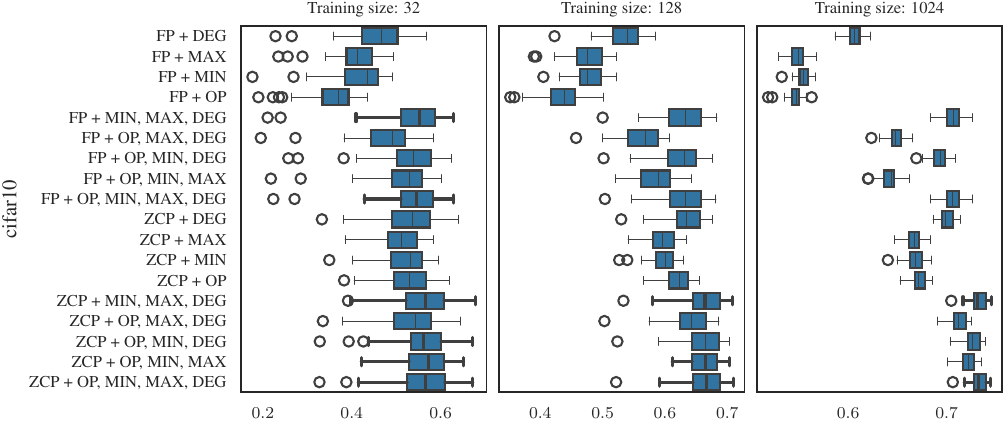}
\caption{NB101 feature group ablation for \texttt{cifar-10}.}
\label{app-abl-nb101}
\end{figure}

\begin{figure}[h]
\includegraphics{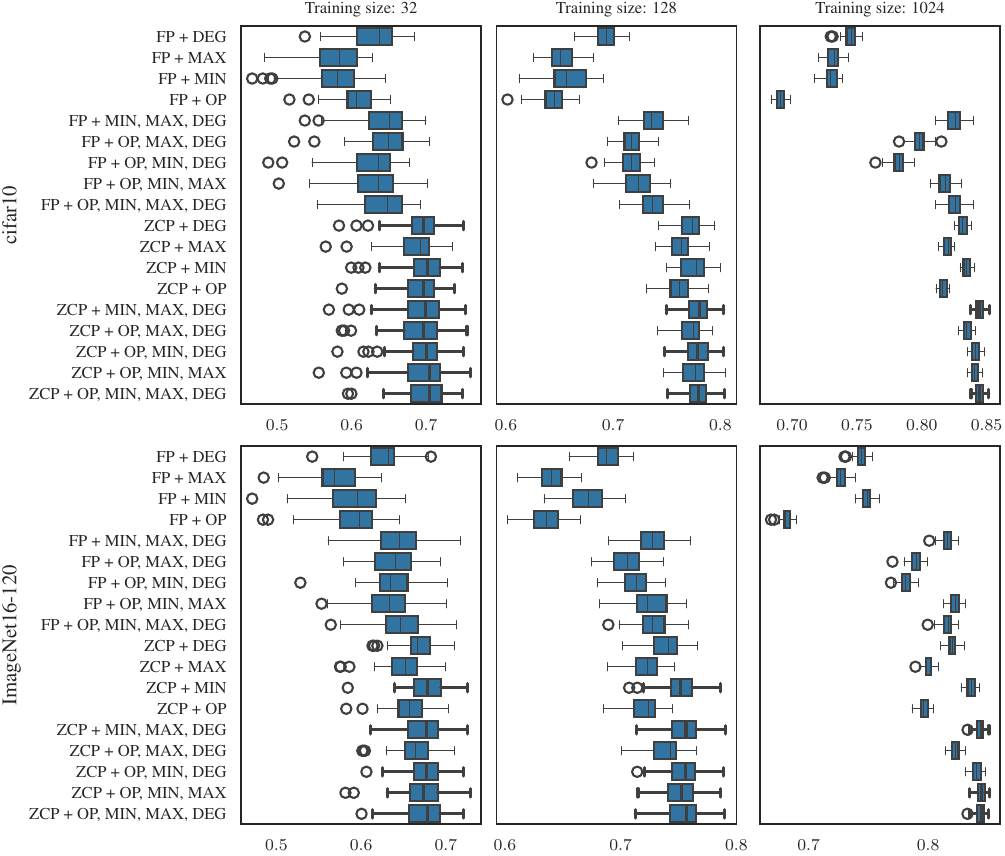}
\caption{NB201 feature group ablation for \texttt{cifar-10} and \texttt{ImageNet16-120}.}
\label{app-abl-nb201}
\end{figure}

We list additional plots of the feature group ablation in Figures \ref{app-abl-nb101} and \ref{app-abl-nb201}.

\FloatBarrier    
\section{Other Encoding Types}
In this section, we list the full results of other encoding types (listed in Section \ref{app-enc-list}) in Tables \ref{app-nasbowl-nb201} and \ref{app-nasbowl-nb101}. It is important to note that NASBOWL uses the WL features in a Gaussian process predictor, and their performance in a random forest might be limited.

\begin{table}
    \small
    \addtolength{\tabcolsep}{-0.2em}
    \centering
    \caption{Results on the validation accuracy prediction tasks for different datasets on NB201 using Arch2vec and WL embedding. Average and standard deviation of Kendall tau over 50 independent runs. Bold values indicate the best results and those that are not statistically different from them.\label{app-nasbowl-nb201}}
    \vskip 0.15in
    \begin{tabular}{r|ccc|ccc|ccc}
dataset & \multicolumn{3}{|c}{cifar10} & \multicolumn{3}{|c}{cifar100} & \multicolumn{3}{|c}{ImageNet16-120} \\
train\_size & 32 & 128 & 1024 & 32 & 128 & 1024 & 32 & 128 & 1024 \\
\hline
OH + WL & $0.54^{0.04}$ & $0.67^{0.02}$ & $0.78^{0.01}$ & $0.55^{0.04}$ & $0.67^{0.02}$ & $0.78^{0.00}$ & $0.55^{0.05}$ & $0.66^{0.02}$ & $0.77^{0.01}$ \\
OH + FP + WL & $0.60^{0.04}$ & $0.68^{0.01}$ & $0.77^{0.01}$ & $0.60^{0.03}$ & $0.68^{0.01}$ & $0.78^{0.01}$ & $0.59^{0.04}$ & $0.67^{0.02}$ & $0.77^{0.01}$ \\
GRAF + WL & $0.63^{0.04}$ & $0.74^{0.01}$ & $0.83^{0.01}$ & $0.65^{0.03}$ & $0.73^{0.01}$ & $0.82^{0.01}$ & $0.63^{0.03}$ & $0.73^{0.02}$ & $0.81^{0.01}$ \\
GRAF + FP + WL & $0.64^{0.03}$ & $0.74^{0.01}$ & $0.82^{0.01}$ & $0.66^{0.03}$ & $0.73^{0.01}$ & $0.82^{0.01}$ & $0.64^{0.03}$ & $0.73^{0.02}$ & $0.82^{0.01}$ \\
GRAF + OH + WL & $0.63^{0.04}$ & $0.74^{0.01}$ & $0.83^{0.01}$ & $0.65^{0.03}$ & $0.73^{0.01}$ & $0.82^{0.01}$ & $0.63^{0.03}$ & $0.72^{0.02}$ & $0.82^{0.01}$ \\
GRAF + OH + FP + WL & $0.63^{0.04}$ & $0.73^{0.01}$ & $0.83^{0.01}$ & $0.65^{0.03}$ & $0.73^{0.01}$ & $0.82^{0.01}$ & $0.64^{0.03}$ & $0.73^{0.02}$ & $0.82^{0.01}$ \\
\hline
ZCP + WL & $\mathbf{0.69^{0.03}}$ & $0.76^{0.01}$ & $0.82^{0.00}$ & $\mathbf{0.70^{0.03}}$ & $0.76^{0.01}$ & $0.82^{0.00}$ & $0.66^{0.02}$ & $0.72^{0.01}$ & $0.81^{0.00}$ \\
ZCP + OH + WL & $0.69^{0.03}$ & $0.76^{0.01}$ & $0.83^{0.00}$ & $\mathbf{0.69^{0.03}}$ & $0.76^{0.01}$ & $0.82^{0.00}$ & $0.66^{0.02}$ & $0.73^{0.01}$ & $0.82^{0.00}$ \\
ZCP + GRAF + WL & $\mathbf{0.69^{0.03}}$ & $\mathbf{0.78^{0.01}}$ & $\mathbf{0.85^{0.00}}$ & $\mathbf{0.70^{0.03}}$ & $\mathbf{0.77^{0.01}}$ & $\mathbf{0.84^{0.00}}$ & $\mathbf{0.67^{0.02}}$ & $\mathbf{0.75^{0.02}}$ & $\mathbf{0.84^{0.00}}$ \\
ZCP + GRAF + OH + WL & $\mathbf{0.69^{0.04}}$ & $0.78^{0.01}$ & $\mathbf{0.84^{0.00}}$ & $\mathbf{0.70^{0.04}}$ & $0.77^{0.01}$ & $\mathbf{0.84^{0.00}}$ & $\mathbf{0.67^{0.03}}$ & $\mathbf{0.75^{0.02}}$ & $0.84^{0.00}$ \\
\hline
OH + A2V & $0.56^{0.05}$ & $0.65^{0.02}$ & $0.72^{0.00}$ & $0.57^{0.04}$ & $0.66^{0.02}$ & $0.73^{0.00}$ & $0.56^{0.04}$ & $0.64^{0.02}$ & $0.72^{0.01}$ \\
OH + FP + A2V & $0.58^{0.05}$ & $0.66^{0.01}$ & $0.73^{0.00}$ & $0.60^{0.04}$ & $0.67^{0.01}$ & $0.74^{0.00}$ & $0.59^{0.04}$ & $0.66^{0.01}$ & $0.73^{0.01}$ \\
GRAF + A2V & $0.62^{0.05}$ & $0.71^{0.02}$ & $0.81^{0.01}$ & $0.64^{0.04}$ & $0.72^{0.02}$ & $0.80^{0.01}$ & $0.62^{0.03}$ & $0.71^{0.02}$ & $0.80^{0.01}$ \\
GRAF + FP + A2V & $0.62^{0.04}$ & $0.72^{0.02}$ & $0.81^{0.01}$ & $0.65^{0.03}$ & $0.72^{0.01}$ & $0.80^{0.01}$ & $0.63^{0.03}$ & $0.71^{0.02}$ & $0.80^{0.01}$ \\
GRAF + OH + A2V & $0.62^{0.04}$ & $0.71^{0.02}$ & $0.81^{0.01}$ & $0.64^{0.04}$ & $0.72^{0.02}$ & $0.81^{0.01}$ & $0.62^{0.03}$ & $0.71^{0.02}$ & $0.80^{0.01}$ \\
GRAF + OH + FP + A2V & $0.62^{0.04}$ & $0.72^{0.02}$ & $0.81^{0.01}$ & $0.64^{0.03}$ & $0.72^{0.01}$ & $0.81^{0.01}$ & $0.63^{0.03}$ & $0.71^{0.02}$ & $0.80^{0.01}$ \\
\hline
ZCP + A2V & $0.67^{0.04}$ & $0.76^{0.01}$ & $0.82^{0.00}$ & $\mathbf{0.69^{0.03}}$ & $0.75^{0.01}$ & $0.81^{0.00}$ & $0.65^{0.03}$ & $0.72^{0.01}$ & $0.80^{0.00}$ \\
ZCP + OH + A2V & $0.67^{0.04}$ & $0.76^{0.01}$ & $0.82^{0.00}$ & $0.68^{0.03}$ & $0.76^{0.01}$ & $0.82^{0.00}$ & $0.65^{0.03}$ & $0.72^{0.01}$ & $0.81^{0.00}$ \\
ZCP + GRAF + A2V & $0.68^{0.04}$ & $0.77^{0.01}$ & $0.84^{0.00}$ & $\mathbf{0.69^{0.03}}$ & $0.76^{0.01}$ & $0.83^{0.00}$ & $\mathbf{0.67^{0.03}}$ & $0.75^{0.02}$ & $0.84^{0.00}$ \\
ZCP + GRAF + OH + A2V & $0.68^{0.04}$ & $0.77^{0.01}$ & $0.84^{0.00}$ & $\mathbf{0.69^{0.03}}$ & $0.76^{0.01}$ & $0.83^{0.00}$ & $\mathbf{0.67^{0.03}}$ & $0.75^{0.02}$ & $0.84^{0.00}$ \\
\end{tabular}
    \end{table}

\begin{table}
    \small
    \addtolength{\tabcolsep}{-0.2em}
    \centering
    \caption{Results on the validation accuracy prediction tasks for cifar10 on NB101 using Arch2vec and WL embedding. Average and standard deviation of Kendall tau over 50 independent runs. Bold values indicate the best results and those that are not statistically different from them.\label{app-nasbowl-nb101}}
    \vskip 0.15in
    \begin{tabular}{r|ccc}
dataset & \multicolumn{3}{|c}{cifar10} \\
train\_size & 32 & 128 & 1024 \\
\hline
GRAF + WL & $0.54^{0.06}$ & $0.63^{0.03}$ & $0.71^{0.01}$ \\
GRAF + FP + WL & $0.53^{0.07}$ & $0.63^{0.03}$ & $0.71^{0.01}$ \\
ZCP + WL & $0.51^{0.05}$ & $0.62^{0.02}$ & $0.68^{0.01}$ \\
ZCP + GRAF + WL & $\mathbf{0.55^{0.07}}$ & $\mathbf{0.67^{0.03}}$ & $\mathbf{0.73^{0.01}}$ \\
GRAF + A2V & $0.49^{0.09}$ & $0.60^{0.04}$ & $0.68^{0.01}$ \\
GRAF + FP + A2V & $0.49^{0.09}$ & $0.60^{0.04}$ & $0.68^{0.01}$ \\
ZCP + A2V & $0.45^{0.07}$ & $0.56^{0.04}$ & $0.63^{0.01}$ \\
ZCP + GRAF + A2V & $0.52^{0.09}$ & $0.64^{0.04}$ & $0.72^{0.01}$ \\
\end{tabular}
    \end{table}

\FloatBarrier
\section{Hardware Metrics Results}
\label{ap:hwexp}
In this section, we present the complete results from hardware metrics predictions. The experiment was realized using HW-NAS-Bench~\citep{hwnasbench}. This benchmark contains networks from NB201 and provides ten hardware statistics on \texttt{cifar-10}, \texttt{cifar-100} and \texttt{ImageNet16-120}. The prediction tasks are of varying difficulty.

\begin{figure}[h!]
\includegraphics{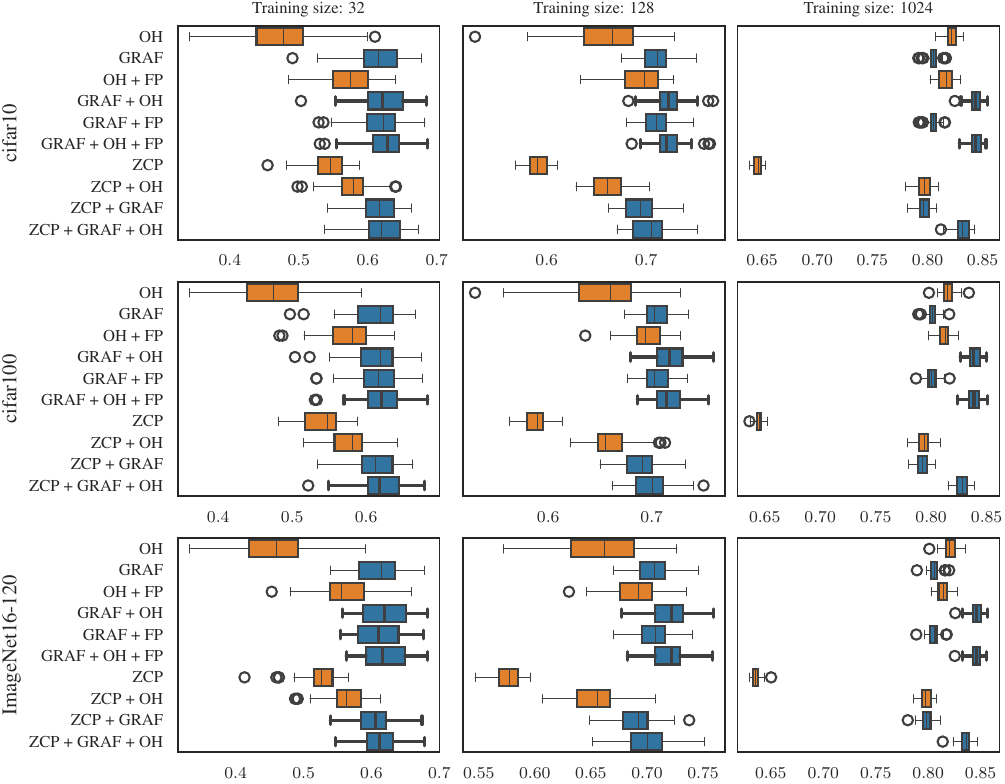}
\caption{Random forest \texttt{edgegpu\_energy} prediction  on NB201 \texttt{cifar-10}, \texttt{cifar-100} and \texttt{ImageNet16-120} (different training sample sizes).}
\label{ap:fig-nb201-edgegpu-img}
\end{figure}

Figure~\ref{ap:fig-nb201-edgegpu-img} shows the results for \texttt{edgegpu\_energy} prediction.
The following Tables~\ref{ap:tab:eyerissar}, \ref{ap:tab:pixlat} and~\ref{ap:tab:raspilat} list average Kendall tau for selected nontrivial tasks and various predictors. In all cases, the best results are produced with settings containing GRAF among predictors. 
The similar results were obtained also on the rest of the tasks from HW-NAS-Bench, the only exceptions are 5 tasks, where onehot + FP produced the best result (from total 90 tasks; three datasets, three different training sizes, 10 hardware metrics).

In general, onehot is a good predictor for HW tasks (as shown in~\citep{hwpredictors}), but on the majority of tasks, the prediction can be improved by adding GRAF to the predictors. 

The computational cost of the experiment (50 evaluations for ten prediction tasks, three datasets and 10 settings, resulting in 15000 runs)  was 1577~s for 32 training samples, 3087~s for 128 training samples, and 16 348~s for 1024 training samples on AMD Ryzen 7 3800X.

\begin{table}
    \small
    \addtolength{\tabcolsep}{-0.2em}
    \centering
    \caption{Results on the {\tt eyeriss\_arithmetic\_intensity} task for various datasets. Average and standard deviation of Kendall tau over 50 independent runs. Bold values indicate the best results and those that are not statistically different from them. \label{ap:tab:eyerissar}}
    \vskip 0.15in
    \begin{tabular}{r|ccc|ccc|ccc}
dataset & \multicolumn{3}{|c}{cifar10} & \multicolumn{3}{|c}{cifar100} & \multicolumn{3}{|c}{ImageNet16-120} \\
train\_size & 32 & 128 & 1024 & 32 & 128 & 1024 & 32 & 128 & 1024 \\
\hline
OH & $0.53^{0.08}$ & $0.73^{0.04}$ & $0.88^{0.01}$ & $0.53^{0.08}$ & $0.73^{0.04}$ & $0.88^{0.01}$ & $0.53^{0.08}$ & $0.73^{0.04}$ & $0.88^{0.01}$ \\
OH + FP & $0.65^{0.04}$ & $0.77^{0.02}$ & $0.94^{0.01}$ & $0.65^{0.04}$ & $0.77^{0.02}$ & $0.94^{0.01}$ & $0.64^{0.04}$ & $0.76^{0.03}$ & $0.92^{0.01}$ \\
GRAF & $\mathbf{0.78^{0.05}}$ & $\mathbf{0.92^{0.01}}$ & $\mathbf{0.99^{0.00}}$ & $\mathbf{0.78^{0.05}}$ & $0.92^{0.01}$ & $\mathbf{0.99^{0.00}}$ & $0.78^{0.05}$ & $\mathbf{0.93^{0.01}}$ & $\mathbf{1.00^{0.00}}$ \\
GRAF + FP & $\mathbf{0.78^{0.05}}$ & $0.91^{0.02}$ & $0.99^{0.00}$ & $\mathbf{0.78^{0.05}}$ & $0.91^{0.02}$ & $0.99^{0.00}$ & $\mathbf{0.79^{0.05}}$ & $0.93^{0.02}$ & $\mathbf{1.00^{0.00}}$ \\
GRAF + OH & $\mathbf{0.78^{0.05}}$ & $\mathbf{0.92^{0.02}}$ & $\mathbf{0.99^{0.00}}$ & $\mathbf{0.78^{0.05}}$ & $\mathbf{0.92^{0.02}}$ & $0.99^{0.00}$ & $0.78^{0.05}$ & $\mathbf{0.93^{0.01}}$ & $\mathbf{1.00^{0.00}}$ \\
GRAF + OH + FP & $\mathbf{0.78^{0.05}}$ & $0.91^{0.02}$ & $0.99^{0.00}$ & $\mathbf{0.78^{0.05}}$ & $0.91^{0.02}$ & $0.99^{0.00}$ & $\mathbf{0.79^{0.05}}$ & $0.93^{0.02}$ & $1.00^{0.00}$ \\
\hline
ZCP & $0.61^{0.04}$ & $0.72^{0.02}$ & $0.88^{0.00}$ & $0.61^{0.04}$ & $0.72^{0.02}$ & $0.88^{0.01}$ & $0.62^{0.03}$ & $0.70^{0.01}$ & $0.85^{0.01}$ \\
ZCP + OH & $0.63^{0.04}$ & $0.74^{0.02}$ & $0.89^{0.00}$ & $0.63^{0.04}$ & $0.74^{0.02}$ & $0.89^{0.00}$ & $0.64^{0.03}$ & $0.73^{0.02}$ & $0.88^{0.01}$ \\
ZCP + GRAF & $0.75^{0.05}$ & $0.89^{0.02}$ & $0.99^{0.00}$ & $0.75^{0.05}$ & $0.89^{0.02}$ & $0.99^{0.00}$ & $\mathbf{0.78^{0.04}}$ & $0.91^{0.02}$ & $0.99^{0.00}$ \\
ZCP + GRAF + OH & $0.75^{0.05}$ & $0.89^{0.02}$ & $0.99^{0.00}$ & $0.75^{0.05}$ & $0.89^{0.02}$ & $0.99^{0.00}$ & $\mathbf{0.78^{0.04}}$ & $0.91^{0.02}$ & $0.99^{0.00}$ \\
\end{tabular}
    \end{table}\begin{table}
    \small
    \addtolength{\tabcolsep}{-0.2em}
    \centering
    \caption{Results on the {\tt pixel3\_latency} task for various datasets. Average and standard deviation of Kendall tau over 50 independent runs. Bold values indicate the best results and those that are not statistically different from them. \label{ap:tab:pixlat}}
    \vskip 0.15in
    \begin{tabular}{r|ccc|ccc|ccc}
dataset & \multicolumn{3}{|c}{cifar10} & \multicolumn{3}{|c}{cifar100} & \multicolumn{3}{|c}{ImageNet16-120} \\
train\_size & 32 & 128 & 1024 & 32 & 128 & 1024 & 32 & 128 & 1024 \\
\hline
OH & $0.56^{0.07}$ & $0.72^{0.03}$ & $0.83^{0.00}$ & $0.56^{0.08}$ & $0.72^{0.03}$ & $0.83^{0.00}$ & $0.57^{0.08}$ & $0.72^{0.03}$ & $0.83^{0.00}$ \\
OH + FP & $0.76^{0.02}$ & $0.81^{0.01}$ & $0.86^{0.00}$ & $0.76^{0.02}$ & $0.81^{0.01}$ & $0.86^{0.00}$ & $0.78^{0.02}$ & $0.82^{0.01}$ & $0.86^{0.00}$ \\
GRAF & $\mathbf{0.80^{0.03}}$ & $\mathbf{0.85^{0.01}}$ & $0.87^{0.00}$ & $\mathbf{0.79^{0.03}}$ & $\mathbf{0.85^{0.01}}$ & $0.86^{0.00}$ & $\mathbf{0.81^{0.03}}$ & $\mathbf{0.85^{0.01}}$ & $0.87^{0.00}$ \\
GRAF + FP & $\mathbf{0.80^{0.02}}$ & $0.85^{0.01}$ & $0.87^{0.00}$ & $\mathbf{0.80^{0.02}}$ & $0.84^{0.01}$ & $0.86^{0.00}$ & $\mathbf{0.81^{0.02}}$ & $0.85^{0.01}$ & $0.87^{0.00}$ \\
GRAF + OH & $0.79^{0.03}$ & $\mathbf{0.85^{0.01}}$ & $0.87^{0.00}$ & $\mathbf{0.79^{0.03}}$ & $\mathbf{0.85^{0.01}}$ & $0.87^{0.00}$ & $\mathbf{0.80^{0.03}}$ & $\mathbf{0.85^{0.01}}$ & $0.87^{0.00}$ \\
GRAF + OH + FP & $\mathbf{0.80^{0.02}}$ & $0.85^{0.01}$ & $0.87^{0.00}$ & $\mathbf{0.80^{0.03}}$ & $0.84^{0.01}$ & $0.87^{0.00}$ & $\mathbf{0.81^{0.03}}$ & $0.85^{0.01}$ & $0.87^{0.00}$ \\
\hline
ZCP & $0.75^{0.02}$ & $0.80^{0.01}$ & $0.86^{0.00}$ & $0.75^{0.02}$ & $0.80^{0.01}$ & $0.85^{0.00}$ & $0.76^{0.02}$ & $0.80^{0.01}$ & $0.86^{0.00}$ \\
ZCP + OH & $0.75^{0.02}$ & $0.81^{0.01}$ & $0.86^{0.00}$ & $0.75^{0.02}$ & $0.81^{0.01}$ & $0.85^{0.00}$ & $0.77^{0.02}$ & $0.81^{0.01}$ & $0.86^{0.00}$ \\
ZCP + GRAF & $0.79^{0.02}$ & $0.84^{0.01}$ & $0.87^{0.00}$ & $\mathbf{0.79^{0.02}}$ & $0.84^{0.01}$ & $\mathbf{0.87^{0.00}}$ & $\mathbf{0.80^{0.03}}$ & $0.85^{0.01}$ & $0.87^{0.00}$ \\
ZCP + GRAF + OH & $0.79^{0.02}$ & $0.84^{0.01}$ & $\mathbf{0.87^{0.00}}$ & $\mathbf{0.79^{0.03}}$ & $0.84^{0.01}$ & $\mathbf{0.87^{0.00}}$ & $0.80^{0.03}$ & $0.85^{0.01}$ & $\mathbf{0.87^{0.00}}$ \\
\end{tabular}
    \end{table}\begin{table}
    \small
    \addtolength{\tabcolsep}{-0.2em}
    \centering
    \caption{Results on the {\tt raspi4\_latency} task for various datasets. Average and standard deviation of Kendall tau over 50 independent runs. Bold values indicate the best results and those that are not statistically different from them. \label{ap:tab:raspilat}}
    \vskip 0.15in
    \begin{tabular}{r|ccc|ccc|ccc}
dataset & \multicolumn{3}{|c}{cifar10} & \multicolumn{3}{|c}{cifar100} & \multicolumn{3}{|c}{ImageNet16-120} \\
train\_size & 32 & 128 & 1024 & 32 & 128 & 1024 & 32 & 128 & 1024 \\
\hline
OH & $0.58^{0.08}$ & $0.73^{0.03}$ & $0.84^{0.00}$ & $0.56^{0.08}$ & $0.72^{0.03}$ & $0.83^{0.00}$ & $0.46^{0.09}$ & $0.61^{0.02}$ & $0.69^{0.00}$ \\
OH + FP & $0.80^{0.01}$ & $0.84^{0.01}$ & $0.87^{0.00}$ & $0.80^{0.02}$ & $0.84^{0.01}$ & $0.87^{0.00}$ & $0.65^{0.03}$ & $0.69^{0.01}$ & $0.72^{0.00}$ \\
GRAF & $0.80^{0.04}$ & $\mathbf{0.86^{0.01}}$ & $0.88^{0.00}$ & $\mathbf{0.81^{0.04}}$ & $\mathbf{0.86^{0.01}}$ & $0.88^{0.00}$ & $0.63^{0.04}$ & $\mathbf{0.70^{0.01}}$ & $0.73^{0.00}$ \\
GRAF + FP & $\mathbf{0.81^{0.02}}$ & $\mathbf{0.86^{0.01}}$ & $0.88^{0.00}$ & $\mathbf{0.81^{0.02}}$ & $\mathbf{0.86^{0.00}}$ & $0.88^{0.00}$ & $\mathbf{0.66^{0.03}}$ & $0.70^{0.01}$ & $0.73^{0.00}$ \\
GRAF + OH & $0.80^{0.03}$ & $\mathbf{0.86^{0.01}}$ & $\mathbf{0.88^{0.00}}$ & $0.80^{0.03}$ & $\mathbf{0.86^{0.01}}$ & $0.88^{0.00}$ & $0.63^{0.04}$ & $\mathbf{0.70^{0.01}}$ & $0.73^{0.00}$ \\
GRAF + OH + FP & $\mathbf{0.81^{0.03}}$ & $\mathbf{0.86^{0.01}}$ & $0.88^{0.00}$ & $\mathbf{0.81^{0.02}}$ & $\mathbf{0.86^{0.00}}$ & $0.88^{0.00}$ & $\mathbf{0.66^{0.02}}$ & $\mathbf{0.70^{0.01}}$ & $0.73^{0.00}$ \\
\hline
ZCP & $0.79^{0.02}$ & $0.84^{0.01}$ & $0.87^{0.00}$ & $0.79^{0.02}$ & $0.83^{0.01}$ & $0.87^{0.00}$ & $0.64^{0.03}$ & $0.68^{0.01}$ & $0.72^{0.00}$ \\
ZCP + OH & $0.79^{0.02}$ & $0.84^{0.01}$ & $0.87^{0.00}$ & $0.79^{0.02}$ & $0.84^{0.01}$ & $0.87^{0.00}$ & $0.64^{0.03}$ & $0.68^{0.01}$ & $0.72^{0.00}$ \\
ZCP + GRAF & $\mathbf{0.81^{0.02}}$ & $\mathbf{0.86^{0.01}}$ & $0.88^{0.00}$ & $\mathbf{0.81^{0.02}}$ & $0.86^{0.00}$ & $\mathbf{0.88^{0.00}}$ & $\mathbf{0.66^{0.03}}$ & $\mathbf{0.70^{0.01}}$ & $0.73^{0.00}$ \\
ZCP + GRAF + OH & $\mathbf{0.81^{0.02}}$ & $\mathbf{0.86^{0.01}}$ & $\mathbf{0.88^{0.00}}$ & $\mathbf{0.81^{0.02}}$ & $0.86^{0.00}$ & $\mathbf{0.88^{0.00}}$ & $\mathbf{0.66^{0.02}}$ & $\mathbf{0.70^{0.01}}$ & $\mathbf{0.73^{0.00}}$ \\
\end{tabular}
    \end{table}

\FloatBarrier
\section{Robustness tasks}
\label{app_robustness}
\begin{figure*}[ht]
\begin{center}
        \includegraphics[width=0.8\textwidth]{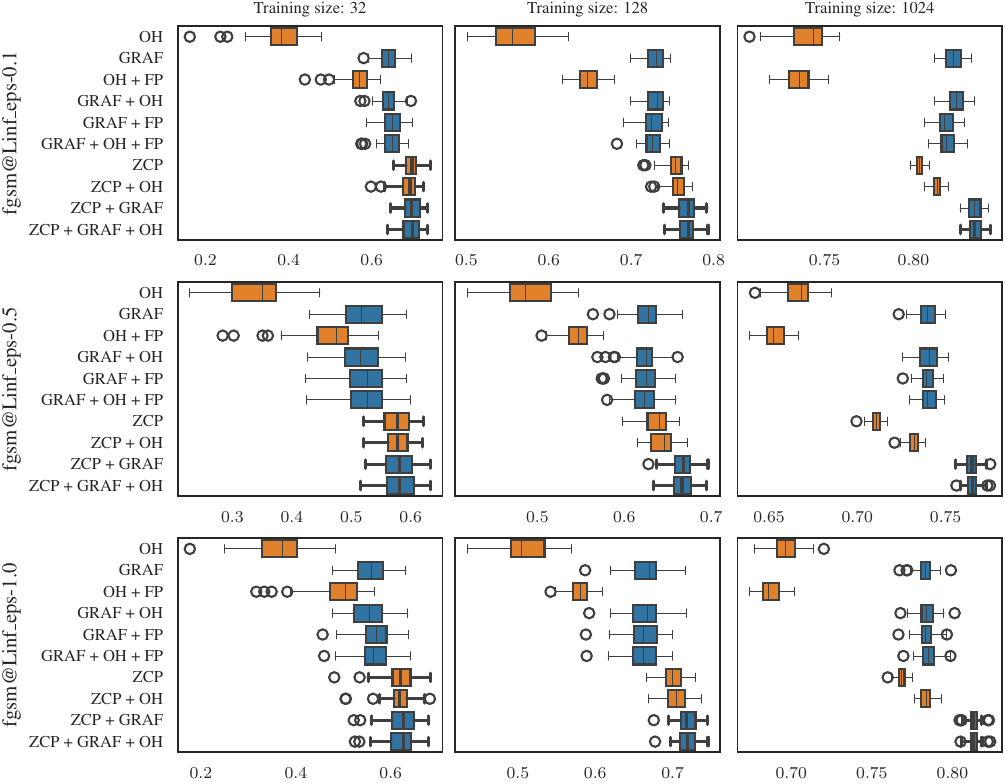}
        \includegraphics[width=0.8\textwidth]{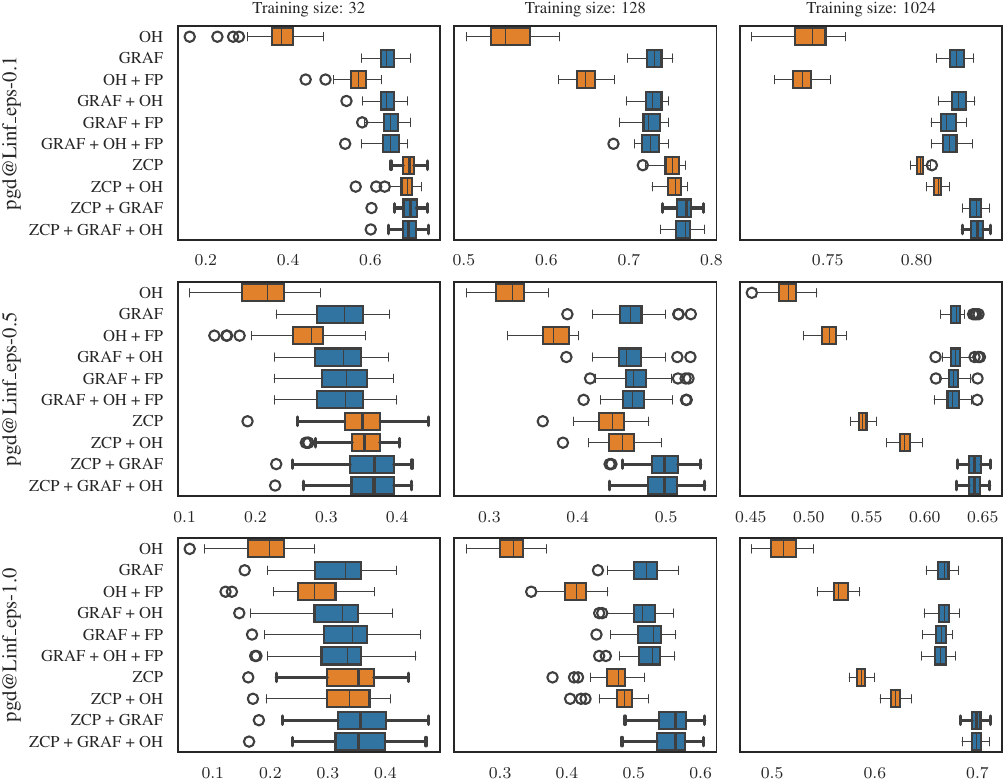}
\caption{Results on predicting the robust accuracy on cifar-10 for the two different adversarial attacks (\textbf{FGSM, PGD}) and different strengths of perturbations from \citet{robustness}.}
\label{fig-nb201-adv_1}
\end{center}
\end{figure*}

Here we present the evaluations on the robustness task on \texttt{cifar-10} on a subset of results for clarity. The Figure \ref{fig-nb201-adv_1} 
presents the results on the FGSM and PGD attacks for the perturbation strengths $\epsilon =\{0.1/255, 0.5/255, 1.0/255 \}$ and the training sizes $32, 128, 1024$ for the first task, predicting only the robust accuracy. The combination of all ZCP, the proposed graph features, and additionally onehot encoding leads to the highest Kendall tau in almost all cases. We present the results on the PGD attack also in Tab. \ref{tab_pgd}.

\begin{figure*}[ht]
\begin{center}
        \includegraphics[width=0.85\textwidth]{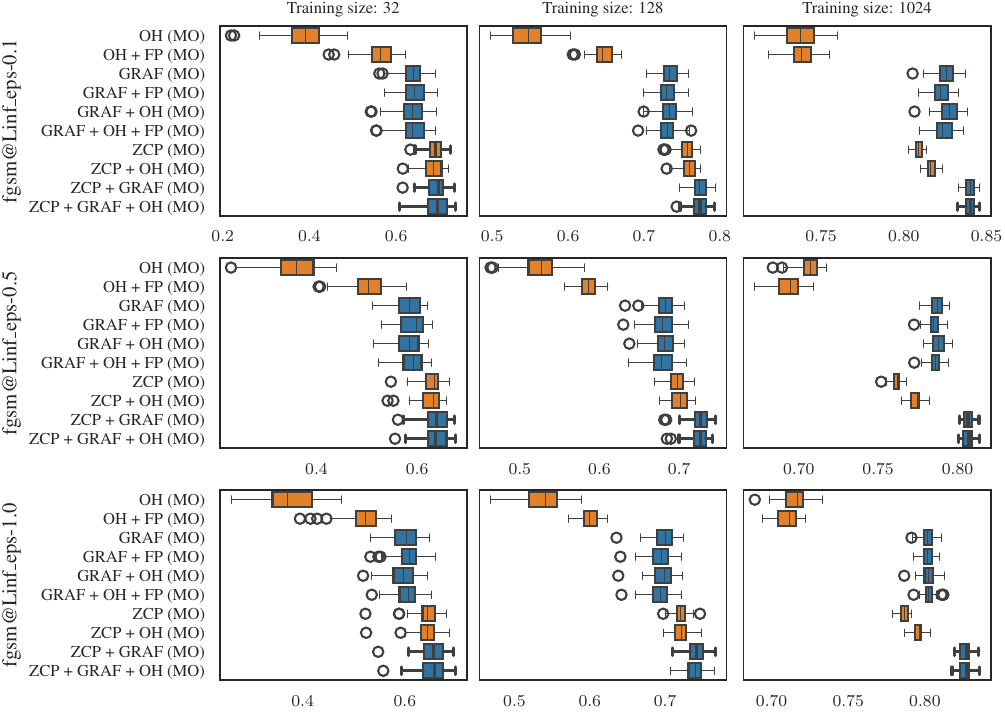}
        \includegraphics[width=0.85\textwidth]{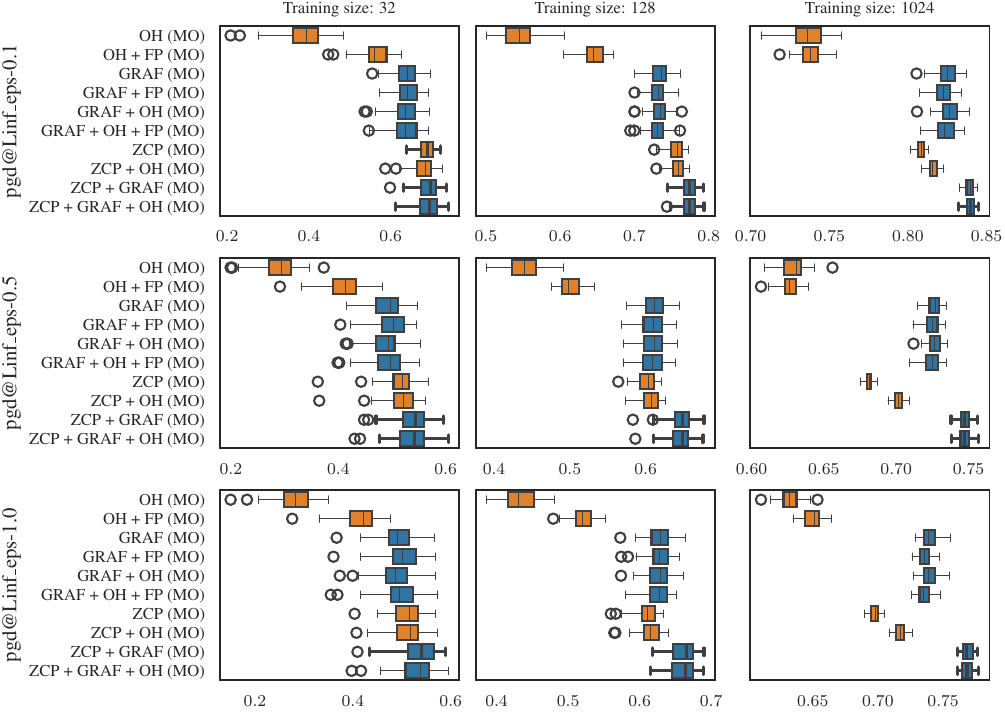}
\caption{Results on the multi-objective prediction tasks for the validation \textbf{and} robust accuracy on cifar-10 for the two different adversarial attacks (\textbf{FGSM, PGD}) and different strengths of perturbations from \citet{robustness}.}
\label{fig-nb201-mo_adv_1}
\end{center}
\end{figure*}

If we turn to the multi-objective case, in which the prediction targets are both the validation \textbf{and} robust accuracy, we see the same conclusions (see Figs. \ref{fig-nb201-mo_adv_1} for visualizations of fgsm and pgd, and Tab. \ref{tab_pgd_mo}.) 

Furthermore, we can see that the multi-objective task has in general a higher Kendall tau than the single robust accuracy prediction task. Interestingly, the prediction task for the attack strength $\epsilon = 0.5/255 $, shows in most cases the lowest correlation, though $\epsilon = 1.0/255 $ is the stronger perturbation attack. 

The computational cost of the experiment (50 evaluations for two adversarial prediction tasks (single and multi-objectives), two adversarial attacks,  three perturbations strengths and three training sizes resulting in 18 000 runs) was 677 s for 32 training samples, 1 537 s for 128 training samples, and 11 404 s for 1024 training samples.

    \begin{table}[h]
    \small
    \addtolength{\tabcolsep}{-0.2em}
    \centering
    \caption{Results on predicting the {\tt pgd} adversarial accuracy on cifar10 for various values of $\epsilon$ on the NB201 benchmark. Average and standard deviation of Kendall tau over 50 independent runs. Bold values indicate the best results and those that are not statistically different from them. \label{tab_pgd}}
    \vskip 0.15in
    \begin{tabular}{r|ccc|ccc|ccc}
dataset & \multicolumn{3}{|c}{pgd@Linf\_eps-0.1} & \multicolumn{3}{|c}{pgd@Linf\_eps-0.5} & \multicolumn{3}{|c}{pgd@Linf\_eps-1.0} \\
train\_size & 32 & 128 & 1024 & 32 & 128 & 1024 & 32 & 128 & 1024 \\
\hline
OH & $0.38^{0.06}$ & $0.55^{0.03}$ & $0.74^{0.01}$ & $0.21^{0.04}$ & $0.32^{0.02}$ & $0.48^{0.01}$ & $0.19^{0.05}$ & $0.31^{0.03}$ & $0.51^{0.02}$ \\
OH + FP & $0.57^{0.03}$ & $0.65^{0.01}$ & $0.74^{0.01}$ & $0.27^{0.04}$ & $0.37^{0.02}$ & $0.52^{0.01}$ & $0.28^{0.05}$ & $0.41^{0.03}$ & $0.57^{0.01}$ \\
GRAF & $0.64^{0.03}$ & $0.73^{0.01}$ & $0.82^{0.01}$ & $0.32^{0.04}$ & $0.46^{0.03}$ & $0.63^{0.01}$ & $0.32^{0.06}$ & $0.52^{0.03}$ & $0.67^{0.01}$ \\
GRAF + FP & $0.65^{0.03}$ & $0.73^{0.01}$ & $0.82^{0.01}$ & $0.32^{0.05}$ & $0.47^{0.02}$ & $0.63^{0.01}$ & $0.33^{0.06}$ & $0.52^{0.03}$ & $0.66^{0.01}$ \\
GRAF + OH & $0.64^{0.03}$ & $0.73^{0.01}$ & $0.82^{0.01}$ & $0.32^{0.04}$ & $0.46^{0.02}$ & $0.63^{0.01}$ & $0.31^{0.06}$ & $0.51^{0.03}$ & $0.67^{0.01}$ \\
GRAF + OH + FP & $0.65^{0.03}$ & $0.73^{0.01}$ & $0.82^{0.01}$ & $0.32^{0.05}$ & $0.46^{0.02}$ & $0.62^{0.01}$ & $0.33^{0.06}$ & $0.52^{0.03}$ & $0.66^{0.01}$ \\
\hline
ZCP & $\mathbf{0.69^{0.02}}$ & $0.75^{0.01}$ & $0.80^{0.00}$ & $\mathbf{0.35^{0.05}}$ & $0.44^{0.02}$ & $0.55^{0.00}$ & $\mathbf{0.34^{0.06}}$ & $0.47^{0.03}$ & $0.59^{0.01}$ \\
ZCP + OH & $0.68^{0.03}$ & $0.75^{0.01}$ & $0.81^{0.00}$ & $\mathbf{0.35^{0.03}}$ & $0.45^{0.02}$ & $0.58^{0.01}$ & $0.33^{0.05}$ & $0.48^{0.02}$ & $0.62^{0.01}$ \\
ZCP + GRAF & $\mathbf{0.69^{0.02}}$ & $\mathbf{0.77^{0.01}}$ & $0.83^{0.00}$ & $\mathbf{0.36^{0.04}}$ & $\mathbf{0.50^{0.02}}$ & $\mathbf{0.64^{0.01}}$ & $\mathbf{0.36^{0.06}}$ & $\mathbf{0.56^{0.03}}$ & $\mathbf{0.70^{0.01}}$ \\
ZCP + GRAF + OH & $\mathbf{0.69^{0.03}}$ & $0.77^{0.01}$ & $\mathbf{0.83^{0.00}}$ & $\mathbf{0.36^{0.04}}$ & $\mathbf{0.50^{0.02}}$ & $\mathbf{0.64^{0.01}}$ & $\mathbf{0.35^{0.06}}$ & $\mathbf{0.56^{0.03}}$ & $0.70^{0.01}$ \\
\end{tabular}
    \end{table}
    \begin{table}[h]
    \small
    \addtolength{\tabcolsep}{-0.2em}
    \centering
    \caption{Results on predicting both the {\tt pgd} adversarial accuracy and clean validation accuracy on cifar10 for various values of $\epsilon$ on the NB201 benchmark. Average and standard deviation of Kendall tau over 50 independent runs. Bold values indicate the best results and those that are not statistically different from them. \label{tab_pgd_mo}}
    \vskip 0.15in
    \begin{tabular}{r|ccc|ccc|ccc}
dataset & \multicolumn{3}{|c}{pgd@Linf\_eps-0.1} & \multicolumn{3}{|c}{pgd@Linf\_eps-0.5} & \multicolumn{3}{|c}{pgd@Linf\_eps-1.0} \\
train\_size & 32 & 128 & 1024 & 32 & 128 & 1024 & 32 & 128 & 1024 \\
\hline
OH (MO) & $0.39^{0.06}$ & $0.55^{0.03}$ & $0.74^{0.01}$ & $0.29^{0.04}$ & $0.44^{0.02}$ & $0.63^{0.01}$ & $0.28^{0.04}$ & $0.43^{0.02}$ & $0.63^{0.01}$ \\
OH + FP (MO) & $0.56^{0.04}$ & $0.65^{0.01}$ & $0.74^{0.01}$ & $0.41^{0.04}$ & $0.50^{0.01}$ & $0.63^{0.01}$ & $0.41^{0.04}$ & $0.52^{0.02}$ & $0.65^{0.01}$ \\
GRAF (MO) & $0.64^{0.03}$ & $0.73^{0.01}$ & $0.82^{0.01}$ & $0.49^{0.03}$ & $0.61^{0.02}$ & $0.73^{0.00}$ & $0.49^{0.04}$ & $0.63^{0.02}$ & $0.74^{0.01}$ \\
GRAF + FP (MO) & $0.64^{0.03}$ & $0.73^{0.01}$ & $0.82^{0.01}$ & $0.50^{0.03}$ & $0.61^{0.02}$ & $0.72^{0.00}$ & $0.50^{0.04}$ & $0.63^{0.02}$ & $0.74^{0.00}$ \\
GRAF + OH (MO) & $0.63^{0.03}$ & $0.73^{0.01}$ & $0.83^{0.01}$ & $0.49^{0.03}$ & $0.61^{0.02}$ & $0.73^{0.01}$ & $0.49^{0.04}$ & $0.62^{0.02}$ & $0.74^{0.01}$ \\
GRAF + OH + FP (MO) & $0.64^{0.04}$ & $0.73^{0.01}$ & $0.82^{0.01}$ & $0.49^{0.03}$ & $0.61^{0.02}$ & $0.72^{0.00}$ & $0.50^{0.04}$ & $0.62^{0.02}$ & $0.74^{0.01}$ \\
\hline
ZCP (MO) & $\mathbf{0.69^{0.02}}$ & $0.76^{0.01}$ & $0.81^{0.00}$ & $0.51^{0.03}$ & $0.60^{0.01}$ & $0.68^{0.00}$ & $0.51^{0.03}$ & $0.61^{0.02}$ & $0.70^{0.00}$ \\
ZCP + OH (MO) & $0.68^{0.03}$ & $0.76^{0.01}$ & $0.82^{0.00}$ & $0.52^{0.03}$ & $0.61^{0.01}$ & $0.70^{0.00}$ & $0.51^{0.03}$ & $0.61^{0.02}$ & $0.72^{0.00}$ \\
ZCP + GRAF (MO) & $\mathbf{0.69^{0.03}}$ & $\mathbf{0.77^{0.01}}$ & $0.84^{0.00}$ & $\mathbf{0.54^{0.03}}$ & $\mathbf{0.64^{0.02}}$ & $\mathbf{0.75^{0.00}}$ & $\mathbf{0.53^{0.04}}$ & $\mathbf{0.66^{0.02}}$ & $\mathbf{0.77^{0.00}}$ \\
ZCP + GRAF + OH (MO) & $\mathbf{0.69^{0.03}}$ & $\mathbf{0.77^{0.01}}$ & $\mathbf{0.84^{0.00}}$ & $\mathbf{0.53^{0.04}}$ & $\mathbf{0.64^{0.02}}$ & $\mathbf{0.75^{0.00}}$ & $0.53^{0.04}$ & $\mathbf{0.66^{0.02}}$ & $\mathbf{0.77^{0.00}}$ \\
\end{tabular}
    \end{table}

\FloatBarrier
\section{BRP-NAS full results}
\label{sec:appbrpnas}

The BRP-NAS model~\citep{brpnas} is a graph convolutional network connected to a single-layer MLP, i.e. a linear regressor on the embedding of the architecture. The additional features are added using so-called \emph{augments}, also supported by the original BRP-NAS model. These get appended to the embedding produced by the GCN. We standardize these features before appending them. 

The BRP-NAS model can be used in two modes -- either as a binary classifier on two inputs or as a regressor that predicts the given target. We use the latter approach for better comparison with the random forest models used in the other experiments.

\begin{table}[h]
\caption{BRP-NAS hyper-parameters}
\label{brpnas-hyper}
\centering
\begin{tabular}{l|l}
\multicolumn{2}{c}{Model Parameters} \\
\hline
Parameter & Value \\
\hline
    num\_layers & 4 \\
    num\_hidden & 600 \\
    dropout\_ratio & 0.002 \\
    weight\_init & thomas \\
    bias\_init &  thomas \\
\hline
\multicolumn{2}{c}{Training Parameters} \\
\hline
Parameter & Value \\
\hline
    epochs & 128 \\
    learning\_rate & $4\times 10^{-4}$ \\
    weight\_decay & $5\times 10^{-5}$ \\
    lr\_patience & 10 \\
    batch\_size & 1 (train = 32) or 16 (otherwise) \\
    optim\_name & adamw \\
    lr\_scheduler & plateau \\
\end{tabular}
\end{table}

The hyper-parameters of the BRP-NAS model are in Table~\ref{brpnas-hyper}. These are based on the settings published together with the original model, but to make the experiments more comparable to the other experiments in this paper, we do not use any validation set and early stopping, and always run the experiments for 128 epochs. The number of epochs was set during preliminary experiments, when we noticed that the early stopping typically stops the training after roughly this number of epochs. We use different batch size for the experiments with only 32 training networks, as for this setting we observed much worse performance with the larger bach size used in the other experiments. Experiments with 128 and 1024 training networks use batch size of 16 -- we did not observe any difference in the performance between the batch size 1 and 16 in these experiments, and the experiments with larger step size run faster.

Every setting of BRP-NAS was trained 50 times with different, independently sampled training sets. All the results were first evaluated using the Kruskal test to assess statistical significance, then the Mann-Whitney rank sum test was used as a posthoc test with significance level $\alpha=0.05$ and Holm correction for multiple testing. The statistical treatment is different for BRP-NAS compared to the other experiments, as with BRP-NAS, we do not use the same training sets for runs with different sets of features. Therefore, we use unpaired statistical tests instead of paired ones.

Training the BRP-NAS models is much slower than training the tree-based models. We run the experiments in a heterogeneous cluster environment. All the experiments always run on a single CPU and a single GPU. A single training and evaluation run on the NB201 benchmark with 32 training networks takes roughly between 35s on an NVIDIA A100-SXM4-40GB and 70s on an NVIDIA GeForce GTX 1080 Ti. With 128 training networks, it takes similar time between 40s and 70s thanks to the larger batch size. With 1024 training networks, the runs take between 3 minutes on an NVIDIA A100-SXM4-40GB and around 6.5 minutes on an NVIDIA Tesla T4.

The experiments on the NB301 dataset are even slower, mainly due to the large number of features. While without the additional features the experiment times are similar to those on NB201, with the full set of features, they take between 4 and 6 minutes on an NVIDIA GeForce GTX 1080 Ti even for the smallest training set. For 128 training networks with all features each run takes between 6 and 7 minutes on an NVIDIA Tesla T4 and for the largest training set, they take between 20 minutes on an NVIDIA A40 and 37 minutes on an NVIDIA GeForce GTX 1080 Ti.

Overall, we needed around 100 GPU days to finish all these experiments (including initial testing, development, and results not included in the paper for brevity).

\begin{table}[h]
    \small
    \addtolength{\tabcolsep}{-0.2em}
    \centering
    \caption{BRP-NAS model results for validation accuracy prediction on cifar10 and ImageNet16-120 on NB201. Average and standard deviation of Kendall tau over 50 independent runs. Bold values indicate the best results and those that are not statistically different from them.}
    \label{tab:brp:acc}
    \vskip 0.15in
    \begin{tabular}{r|ccc|ccc|ccc}
dataset & \multicolumn{3}{|c}{NB201: cifar10} & \multicolumn{3}{|c}{NB201: ImageNet16-120} & \multicolumn{3}{|c}{NB301: cifar10} \\
train\_size & 32 & 128 & 1024 & 32 & 128 & 1024 & 32 & 128 & 1024 \\
\hline
None (BRP-NAS) & $0.48^{0.05}$ & $0.58^{0.04}$ & $0.80^{0.02}$ & $0.48^{0.06}$ & $0.61^{0.03}$ & $0.84^{0.01}$ & $0.37^{0.05}$ & $0.48^{0.02}$ & $0.58^{0.01}$ \\
OH & $0.53^{0.05}$ & $0.63^{0.02}$ & $0.77^{0.02}$ & $0.54^{0.04}$ & $0.63^{0.02}$ & $0.83^{0.01}$ & $0.21^{0.08}$ & $0.46^{0.03}$ & $0.59^{0.01}$ \\
OH + FP & $0.57^{0.04}$ & $0.64^{0.02}$ & $0.77^{0.02}$ & $0.54^{0.04}$ & $0.64^{0.02}$ & $0.83^{0.02}$ & $0.28^{0.06}$ & $0.47^{0.02}$ & $0.58^{0.01}$ \\
GRAF & $0.62^{0.04}$ & $0.69^{0.01}$ & $0.79^{0.01}$ & $0.62^{0.04}$ & $0.69^{0.02}$ & $0.83^{0.01}$ & $\mathbf{0.38^{0.05}}$ & $\mathbf{0.54^{0.03}}$ & $0.67^{0.01}$ \\
GRAF + FP & $0.62^{0.03}$ & $0.69^{0.02}$ & $0.78^{0.01}$ & $0.62^{0.03}$ & $0.69^{0.02}$ & $0.83^{0.01}$ & $\mathbf{0.37^{0.07}}$ & $\mathbf{0.55^{0.02}}$ & $\mathbf{0.67^{0.01}}$ \\
GRAF + OH & $0.59^{0.05}$ & $0.69^{0.02}$ & $0.79^{0.01}$ & $0.61^{0.04}$ & $0.69^{0.02}$ & $0.83^{0.01}$ & $\mathbf{0.35^{0.09}}$ & $\mathbf{0.54^{0.02}}$ & $0.67^{0.01}$ \\
GRAF + OH + FP & $0.61^{0.04}$ & $0.69^{0.02}$ & $0.79^{0.01}$ & $0.61^{0.04}$ & $0.68^{0.02}$ & $0.83^{0.02}$ & $0.35^{0.07}$ & $\mathbf{0.54^{0.03}}$ & $0.66^{0.01}$ \\
\hline
ZCP & $\mathbf{0.63^{0.03}}$ & $0.66^{0.02}$ & $0.80^{0.02}$ & $0.56^{0.04}$ & $0.63^{0.03}$ & $\mathbf{0.85^{0.01}}$ & $\mathbf{0.40^{0.06}}$ & $0.51^{0.02}$ & $0.61^{0.00}$ \\
ZCP + OH & $\mathbf{0.64^{0.04}}$ & $0.68^{0.02}$ & $0.79^{0.01}$ & $0.58^{0.03}$ & $0.66^{0.02}$ & $0.83^{0.02}$ & $0.32^{0.06}$ & $0.50^{0.02}$ & $0.61^{0.01}$ \\
ZCP + GRAF & $\mathbf{0.65^{0.05}}$ & $\mathbf{0.72^{0.02}}$ & $\mathbf{0.81^{0.01}}$ & $\mathbf{0.64^{0.03}}$ & $\mathbf{0.71^{0.02}}$ & $0.84^{0.01}$ & $0.35^{0.06}$ & $\mathbf{0.55^{0.02}}$ & $\mathbf{0.67^{0.01}}$ \\
ZCP + GRAF + OH & $\mathbf{0.64^{0.04}}$ & $\mathbf{0.72^{0.02}}$ & $\mathbf{0.81^{0.01}}$ & $\mathbf{0.63^{0.03}}$ & $\mathbf{0.71^{0.02}}$ & $0.84^{0.01}$ & $\mathbf{0.37^{0.07}}$ & $\mathbf{0.55^{0.02}}$ & $0.67^{0.01}$ \\
\end{tabular}
    \end{table}
\begin{table}[h]
    \small
    \addtolength{\tabcolsep}{-0.2em}
    \centering
    \caption{BRP-NAS model results for various hardware tasks on cifar10 and NB201. Average and standard deviation of Kendall tau over 50 independent runs. Bold values indicate the best results and those that are not statistically different from them.}
    \label{ap:brp:hw1}
    \vskip 0.15in
    \begin{tabular}{r|ccc|ccc|ccc}
dataset & \multicolumn{3}{|c}{edgegpu\_energy} & \multicolumn{3}{|c}{edgegpu\_latency} & \multicolumn{3}{|c}{edgetpu\_latency} \\
train\_size & 32 & 128 & 1024 & 32 & 128 & 1024 & 32 & 128 & 1024 \\
\hline
None (BRP-NAS) & $0.59^{0.05}$ & $0.71^{0.01}$ & $0.80^{0.00}$ & $0.59^{0.05}$ & $0.72^{0.01}$ & $0.80^{0.00}$ & $\mathbf{0.59^{0.08}}$ & $0.69^{0.04}$ & $\mathbf{0.84^{0.02}}$ \\
OH & $0.67^{0.05}$ & $0.77^{0.01}$ & $0.87^{0.00}$ & $\mathbf{0.73^{0.04}}$ & $\mathbf{0.83^{0.01}}$ & $\mathbf{0.87^{0.00}}$ & $\mathbf{0.55^{0.13}}$ & $\mathbf{0.73^{0.04}}$ & $\mathbf{0.83^{0.02}}$ \\
OH + FP & $0.69^{0.04}$ & $0.78^{0.01}$ & $\mathbf{0.87^{0.00}}$ & $\mathbf{0.74^{0.03}}$ & $\mathbf{0.83^{0.01}}$ & $\mathbf{0.87^{0.00}}$ & $\mathbf{0.56^{0.12}}$ & $\mathbf{0.73^{0.04}}$ & $\mathbf{0.83^{0.02}}$ \\
GRAF & $0.75^{0.03}$ & $0.81^{0.01}$ & $0.86^{0.00}$ & $\mathbf{0.73^{0.04}}$ & $0.82^{0.01}$ & $0.87^{0.00}$ & $0.49^{0.16}$ & $\mathbf{0.73^{0.04}}$ & $0.81^{0.02}$ \\
GRAF + FP & $0.75^{0.02}$ & $0.82^{0.01}$ & $0.86^{0.00}$ & $0.73^{0.03}$ & $0.82^{0.01}$ & $0.87^{0.00}$ & $0.44^{0.16}$ & $\mathbf{0.73^{0.03}}$ & $0.81^{0.02}$ \\
GRAF + OH & $\mathbf{0.77^{0.03}}$ & $\mathbf{0.83^{0.01}}$ & $\mathbf{0.87^{0.00}}$ & $\mathbf{0.75^{0.03}}$ & $\mathbf{0.83^{0.01}}$ & $\mathbf{0.87^{0.00}}$ & $0.41^{0.15}$ & $\mathbf{0.72^{0.04}}$ & $0.81^{0.02}$ \\
GRAF + OH + FP & $\mathbf{0.77^{0.02}}$ & $\mathbf{0.83^{0.01}}$ & $\mathbf{0.87^{0.00}}$ & $\mathbf{0.75^{0.04}}$ & $\mathbf{0.83^{0.01}}$ & $\mathbf{0.87^{0.00}}$ & $0.42^{0.14}$ & $\mathbf{0.72^{0.04}}$ & $0.81^{0.02}$ \\
\hline
ZCP & $0.63^{0.05}$ & $0.72^{0.01}$ & $0.81^{0.00}$ & $0.62^{0.05}$ & $0.72^{0.01}$ & $0.81^{0.00}$ & $\mathbf{0.60^{0.09}}$ & $0.70^{0.04}$ & $\mathbf{0.83^{0.02}}$ \\
ZCP + OH & $0.71^{0.04}$ & $0.78^{0.01}$ & $0.87^{0.00}$ & $0.72^{0.05}$ & $0.83^{0.01}$ & $\mathbf{0.87^{0.00}}$ & $0.47^{0.13}$ & $\mathbf{0.72^{0.04}}$ & $0.82^{0.02}$ \\
ZCP + GRAF & $0.74^{0.03}$ & $0.81^{0.01}$ & $0.86^{0.00}$ & $0.69^{0.05}$ & $0.82^{0.01}$ & $0.87^{0.00}$ & $0.45^{0.12}$ & $\mathbf{0.72^{0.03}}$ & $0.81^{0.02}$ \\
ZCP + GRAF + OH & $0.75^{0.04}$ & $0.83^{0.01}$ & $\mathbf{0.87^{0.00}}$ & $0.72^{0.05}$ & $0.82^{0.01}$ & $0.87^{0.00}$ & $0.41^{0.16}$ & $0.71^{0.04}$ & $0.81^{0.02}$ \\
\end{tabular}
    \end{table}
    \begin{table}[h]
    \small
    \addtolength{\tabcolsep}{-0.2em}
    \centering
    \caption{BRP-NAS model results on the {\tt fgsm} adversarial accuracy on cifar10 for various values of $\epsilon$ on the NB201 benchmark. Average and standard deviation of Kendall tau over 50 independent runs. Bold values indicate the best results and those that are not statistically different from them.}
    \label{ap:brp:advs}
    \vskip 0.15in
    \begin{tabular}{r|ccc|ccc|ccc}
dataset & \multicolumn{3}{|c}{fgsm@Linf\_eps-0.1} & \multicolumn{3}{|c}{fgsm@Linf\_eps-0.5} & \multicolumn{3}{|c}{fgsm@Linf\_eps-1.0} \\
train\_size & 32 & 128 & 1024 & 32 & 128 & 1024 & 32 & 128 & 1024 \\
\hline
None (BRP-NAS) & $0.48^{0.05}$ & $0.58^{0.04}$ & $\mathbf{0.80^{0.01}}$ & $0.42^{0.07}$ & $0.55^{0.03}$ & $0.74^{0.01}$ & $0.44^{0.06}$ & $0.60^{0.03}$ & $0.78^{0.01}$ \\
OH & $0.53^{0.05}$ & $0.64^{0.03}$ & $0.78^{0.02}$ & $0.44^{0.05}$ & $0.55^{0.04}$ & $\mathbf{0.74^{0.01}}$ & $0.47^{0.06}$ & $0.60^{0.02}$ & $0.79^{0.01}$ \\
OH + FP & $0.56^{0.04}$ & $0.65^{0.02}$ & $0.78^{0.02}$ & $0.45^{0.05}$ & $0.57^{0.02}$ & $0.74^{0.01}$ & $0.48^{0.05}$ & $\mathbf{0.61^{0.02}}$ & $0.79^{0.01}$ \\
GRAF & $0.60^{0.04}$ & $0.69^{0.02}$ & $0.78^{0.02}$ & $\mathbf{0.50^{0.04}}$ & $\mathbf{0.58^{0.02}}$ & $0.69^{0.03}$ & $\mathbf{0.54^{0.05}}$ & $\mathbf{0.62^{0.02}}$ & $0.77^{0.03}$ \\
GRAF + FP & $\mathbf{0.61^{0.05}}$ & $0.69^{0.02}$ & $0.78^{0.01}$ & $\mathbf{0.48^{0.06}}$ & $0.58^{0.01}$ & $0.69^{0.02}$ & $\mathbf{0.55^{0.05}}$ & $\mathbf{0.62^{0.02}}$ & $0.77^{0.03}$ \\
GRAF + OH & $0.58^{0.04}$ & $0.68^{0.02}$ & $0.79^{0.01}$ & $\mathbf{0.47^{0.06}}$ & $0.58^{0.02}$ & $0.70^{0.02}$ & $\mathbf{0.54^{0.04}}$ & $\mathbf{0.62^{0.02}}$ & $0.77^{0.03}$ \\
GRAF + OH + FP & $0.59^{0.05}$ & $0.68^{0.02}$ & $0.78^{0.02}$ & $\mathbf{0.49^{0.05}}$ & $0.58^{0.02}$ & $0.69^{0.02}$ & $\mathbf{0.53^{0.05}}$ & $\mathbf{0.62^{0.02}}$ & $0.76^{0.04}$ \\
\hline
ZCP & $\mathbf{0.62^{0.04}}$ & $0.67^{0.02}$ & $0.80^{0.01}$ & $\mathbf{0.48^{0.06}}$ & $0.56^{0.03}$ & $\mathbf{0.74^{0.01}}$ & $0.50^{0.04}$ & $\mathbf{0.61^{0.03}}$ & $\mathbf{0.79^{0.01}}$ \\
ZCP + OH & $\mathbf{0.62^{0.04}}$ & $0.68^{0.02}$ & $0.79^{0.01}$ & $\mathbf{0.48^{0.04}}$ & $0.58^{0.02}$ & $\mathbf{0.74^{0.01}}$ & $0.51^{0.06}$ & $\mathbf{0.62^{0.02}}$ & $\mathbf{0.79^{0.01}}$ \\
ZCP + GRAF & $\mathbf{0.63^{0.04}}$ & $\mathbf{0.71^{0.02}}$ & $\mathbf{0.80^{0.02}}$ & $\mathbf{0.50^{0.05}}$ & $\mathbf{0.59^{0.01}}$ & $0.70^{0.02}$ & $\mathbf{0.55^{0.06}}$ & $\mathbf{0.62^{0.02}}$ & $0.77^{0.02}$ \\
ZCP + GRAF + OH & $\mathbf{0.62^{0.03}}$ & $\mathbf{0.71^{0.02}}$ & $\mathbf{0.81^{0.01}}$ & $\mathbf{0.49^{0.05}}$ & $\mathbf{0.59^{0.01}}$ & $0.70^{0.02}$ & $\mathbf{0.54^{0.04}}$ & $\mathbf{0.62^{0.02}}$ & $0.77^{0.03}$ \\
\end{tabular}
    \end{table}\begin{table}[h]
    \small
    \addtolength{\tabcolsep}{-0.2em}
    \centering
    \caption{BRP-NAS model results on the {\tt pgd} adversarial accuracy on cifar10 for various values of $\epsilon$ on the NB201 benchmark. Average and standard deviation of Kendall tau over 50 independent runs. Bold values indicate the best results and those that are not statistically different from them.}
    \label{ap:brp:adve}
    \vskip 0.15in
    \begin{tabular}{r|ccc|ccc|ccc}
dataset & \multicolumn{3}{|c}{pgd@Linf\_eps-0.1} & \multicolumn{3}{|c}{pgd@Linf\_eps-0.5} & \multicolumn{3}{|c}{pgd@Linf\_eps-1.0} \\
train\_size & 32 & 128 & 1024 & 32 & 128 & 1024 & 32 & 128 & 1024 \\
\hline
None (BRP-NAS) & $0.48^{0.05}$ & $0.58^{0.05}$ & $0.79^{0.02}$ & $0.24^{0.07}$ & $0.36^{0.04}$ & $\mathbf{0.61^{0.01}}$ & $0.21^{0.06}$ & $\mathbf{0.37^{0.06}}$ & $0.68^{0.01}$ \\
OH & $0.53^{0.04}$ & $0.64^{0.03}$ & $0.78^{0.02}$ & $0.24^{0.06}$ & $0.37^{0.03}$ & $\mathbf{0.61^{0.02}}$ & $0.22^{0.07}$ & $0.36^{0.05}$ & $\mathbf{0.68^{0.01}}$ \\
OH + FP & $0.56^{0.03}$ & $0.65^{0.02}$ & $0.78^{0.02}$ & $0.25^{0.07}$ & $0.36^{0.03}$ & $\mathbf{0.60^{0.02}}$ & $0.24^{0.05}$ & $\mathbf{0.38^{0.05}}$ & $\mathbf{0.68^{0.01}}$ \\
GRAF & $0.61^{0.04}$ & $0.69^{0.02}$ & $0.78^{0.01}$ & $\mathbf{0.28^{0.07}}$ & $\mathbf{0.39^{0.02}}$ & $0.55^{0.03}$ & $\mathbf{0.28^{0.05}}$ & $\mathbf{0.39^{0.04}}$ & $0.66^{0.04}$ \\
GRAF + FP & $0.60^{0.05}$ & $0.69^{0.01}$ & $0.79^{0.01}$ & $\mathbf{0.28^{0.08}}$ & $\mathbf{0.39^{0.02}}$ & $0.53^{0.03}$ & $\mathbf{0.27^{0.07}}$ & $\mathbf{0.39^{0.03}}$ & $0.66^{0.05}$ \\
GRAF + OH & $0.59^{0.04}$ & $0.68^{0.02}$ & $0.79^{0.01}$ & $\mathbf{0.29^{0.05}}$ & $\mathbf{0.39^{0.02}}$ & $0.54^{0.04}$ & $\mathbf{0.27^{0.06}}$ & $\mathbf{0.39^{0.04}}$ & $0.66^{0.04}$ \\
GRAF + OH + FP & $0.60^{0.05}$ & $0.68^{0.02}$ & $0.79^{0.01}$ & $\mathbf{0.31^{0.04}}$ & $\mathbf{0.39^{0.02}}$ & $0.54^{0.03}$ & $0.25^{0.09}$ & $\mathbf{0.38^{0.03}}$ & $0.65^{0.05}$ \\
\hline
ZCP & $\mathbf{0.64^{0.04}}$ & $0.67^{0.02}$ & $\mathbf{0.80^{0.02}}$ & $0.25^{0.07}$ & $0.37^{0.03}$ & $\mathbf{0.61^{0.02}}$ & $0.24^{0.06}$ & $\mathbf{0.38^{0.04}}$ & $\mathbf{0.68^{0.01}}$ \\
ZCP + OH & $\mathbf{0.63^{0.04}}$ & $0.68^{0.02}$ & $0.79^{0.01}$ & $0.26^{0.06}$ & $0.37^{0.03}$ & $\mathbf{0.61^{0.02}}$ & $0.25^{0.06}$ & $\mathbf{0.38^{0.05}}$ & $\mathbf{0.69^{0.01}}$ \\
ZCP + GRAF & $\mathbf{0.64^{0.04}}$ & $\mathbf{0.72^{0.02}}$ & $\mathbf{0.80^{0.01}}$ & $\mathbf{0.30^{0.06}}$ & $\mathbf{0.40^{0.03}}$ & $0.55^{0.03}$ & $\mathbf{0.30^{0.07}}$ & $\mathbf{0.40^{0.04}}$ & $0.66^{0.04}$ \\
ZCP + GRAF + OH & $\mathbf{0.62^{0.04}}$ & $\mathbf{0.71^{0.02}}$ & $\mathbf{0.81^{0.01}}$ & $\mathbf{0.28^{0.06}}$ & $\mathbf{0.39^{0.03}}$ & $0.55^{0.03}$ & $\mathbf{0.27^{0.07}}$ & $\mathbf{0.40^{0.04}}$ & $0.66^{0.05}$ \\
\end{tabular}
    \end{table}

\FloatBarrier
\section{Comparison with TA-GATES}
\label{app-tagates}
TA-GATES demonstrates how zero-cost proxies can improve graph neural network predictors. It is a graph neural network that mimics the forward and backward pass of information, and uses zero-cost proxies to distinguish operations that would otherwise have the same encoding due to similar contexts in the network \citep{tagates}. The model outperforms some strong existing predictors, such as XGBoost \citep{xgboost} or GCN \citep{gcnpred}.

For the TA-GATES experiments, we use the same evaluation setting as in their original paper, as they provide a fixed train and test set that cannot be mixed due to different data formats (the test set is missing some information). We evaluate only on NB201, since NB101 and NB301 are different samples than those from NB-Suite-Zero. After removing networks with unreachable branches (see Section \ref{appendix-unreachable}), we obtain 4~675 train and 4~770 test networks. For every evaluation, we use a subset of the train set for fitting, and evaluate on the whole held-out test set. We use the following fractions of the train set -- 0.01, 0.05, 0.1, 0.2 (corresponding to 46, 233, 467 and 935 networks), and evaluate each model on 50 different seeds.

\begin{figure}[h]
\centering
\includegraphics[width=0.8\textwidth]{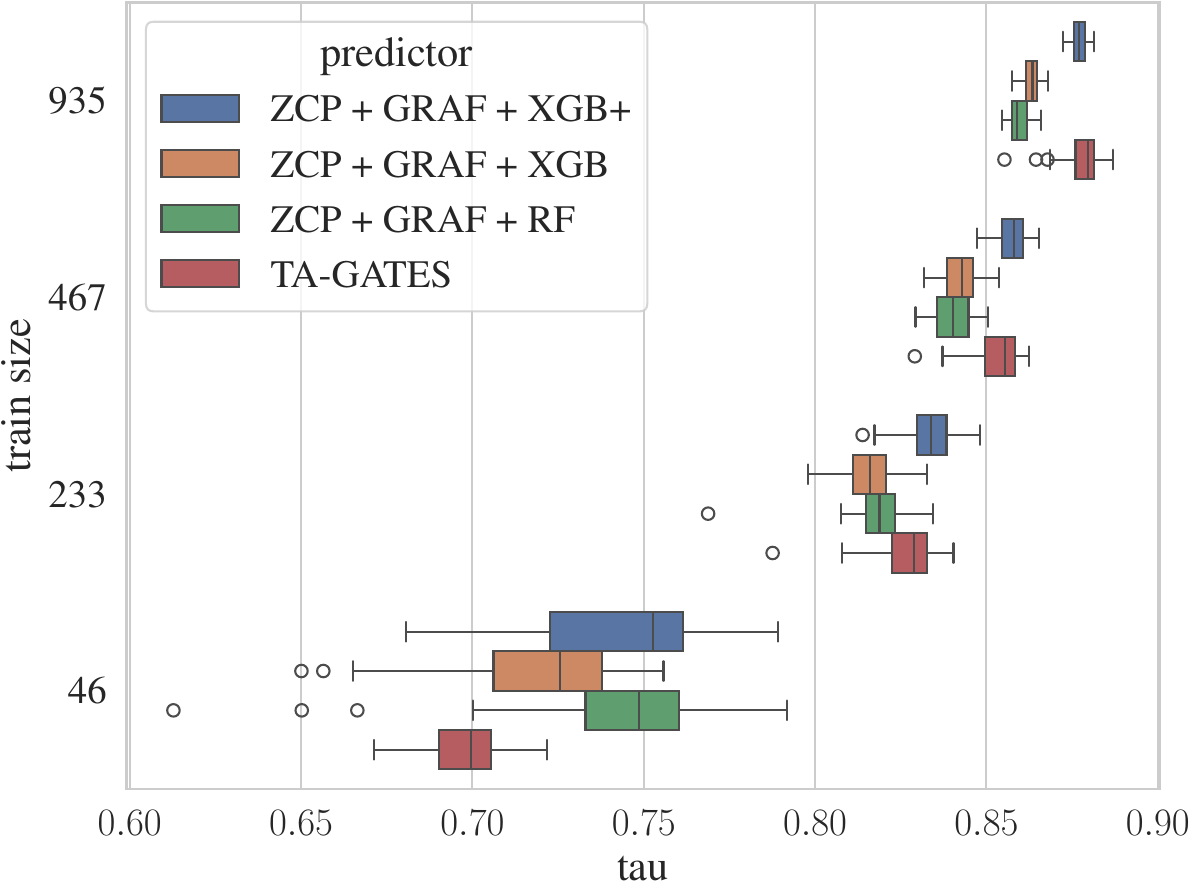}
\caption{Results on NB201 \texttt{cifar-10} -- comparison of ZCP + GRAF models and TA-GATES. \label{fig:tagates-boxes}}
\end{figure}

TA-GATES requires over 10 minutes to train on a single seed (for all sample sizes), but ZCP + GRAF + XGB+ needs only 12 seconds on the largest sample size. The total time required to run all 50 repetitions and 4 sample sizes is listed in Table \ref{app-ta-runtimes} (on Intel Xeon CPU E5-2620, NVIDIA GeForce GTX 1080 Ti, no networks from the search space were trained since we use precomputed search space data). ZCP computation time was not included for both models, as TA-GATES provides only precomputed scores. Figure \ref{fig:tagates-boxes} shows the results of all models and shows that ZCP + GRAF is competitive with TA-GATES at a fraction of the train time.

\begin{table}[h]
\centering
\small
\caption{Runtimes of different models across all experiments. ZCP + GRAF experiments take less then 1 hour on a CPU, while TA-GATES needs over 37 GPU hours.\label{app-ta-runtimes}. Time for computing ZCP is not included.}
\begin{tabular}{lr}
\toprule
predictor  & runtime \\
\midrule
ZCP + GRAF + RF & 4.9 CPU min \\
ZCP + GRAF + XGB & 0.86 CPU min \\
ZCP + GRAF + XGB+ & 42.7 CPU min \\
TA-GATES & 37.65 GPU hours \\
\bottomrule
\end{tabular}
\end{table}

\FloatBarrier
\section{NASLib Predictor Evaluation}
\label{naslib-predictors}
Unlike the original study, we evaluate our approach on the NB101 and NB301 sampled networks, and on NB201 without unreachable networks.
For predictor evaluation, we use the model-based predictors from NASLib \citep{naslib-predictors}. These are as follows:

\begin{itemize}
    \item Neural predictors -- BANANAS \citep{white2019bananas}, BONAS \citep{bonas}, GCN \citep{gcnpred}, MLP~\citep{white2019bananas}, NAO \citep{NAO}, SemiNAS~\citep{seminas}
    \item Based on Bayesian optimization -- Bayesian Linear Regression \citep{naslib-predictors}, BOHAMIANN \citep{bohamiann}, DNGO \citep{DNGO}, BONAS
    \item Tree-based methods -- LGBoost \citep{lgboost}, NGBoost \citep{ngboost}, random forest \citep{RFcite}, XGBoost (XGB)~\citep{xgboost}
    \item Gaussian process methods -- GP \citep{Rasmussen2004}, Sparse GP \citep{sparsegp}, Variational Sparse GP~\citep{varsparsegp}
\end{itemize}

The methods have up to 15 minutes of hyperparameter tuning available. For more details, refer to the original study \citep{naslib-predictors}. For ZCP + GRAF evaluation, we use 5 different models -- random forest (no tuning), XGB (no tuning), random forest and XGB with hyperparameter tuning, and the XGB+ model (see Section \ref{app-grafdetails} for details).

We evaluate the quality of the predictors based on the size of the training sets. The plots show the mean Kendall tau over 100 runs with the shaded areas showing the standard deviation of the results. All the methods use the same seeds for generating their sets in the 100 runs. The methods connected by the vertical lines in the critical difference plot on the left-hand side are not statistically significantly different for the largest training size. The script to create these plots is a part of the MCBO library \citep{dreczkowski2023mcbo}.

All the results were first evaluated using the Friedmann test to assess statistical significance, then the Wilcoxon signed-rank test was used as a posthoc test with significance level $\alpha=0.05$ and Holm correction for multiple testing.

In total, we used 21 CPU days per each of the 4 tasks, with 32GB RAM and 8 cores allocated per run, on Intel Xeon Gold 6242 CPU. No networks from the search space were trained, as we queried the accuracies.

\begin{figure}[h]
\includegraphics{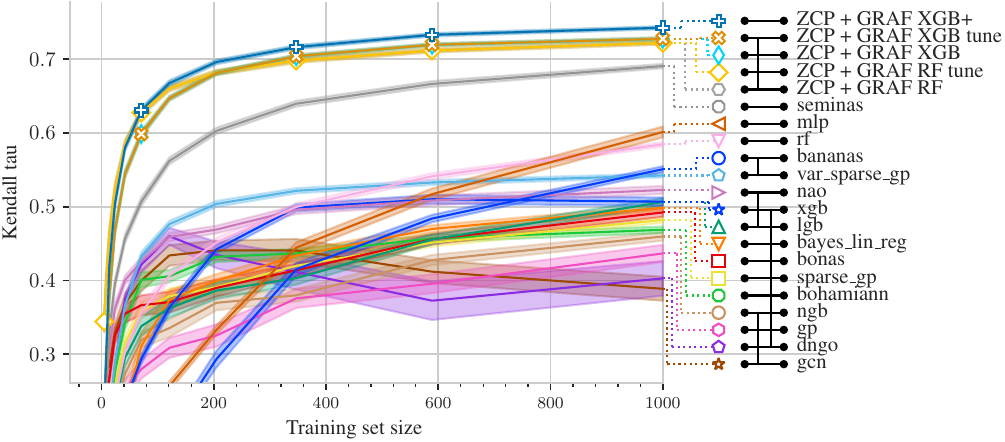}
\caption{NASLib predictor evaluation, NAS-Bench-101 and \texttt{cifar-10}.\label{fig:naslib101}}
\end{figure}

Figures \ref{fig:naslib101}--\ref{fig:naslib101} show results on NB101, NB201, and NB301 for \texttt{cifar-10}, and NB201 for \texttt{ImageNet16-120}. ZCP + GRAF + XGB+ is the overall best predictor, and all our ZCP + GRAF predictors outperform existing models.

\begin{figure}[h]
\includegraphics{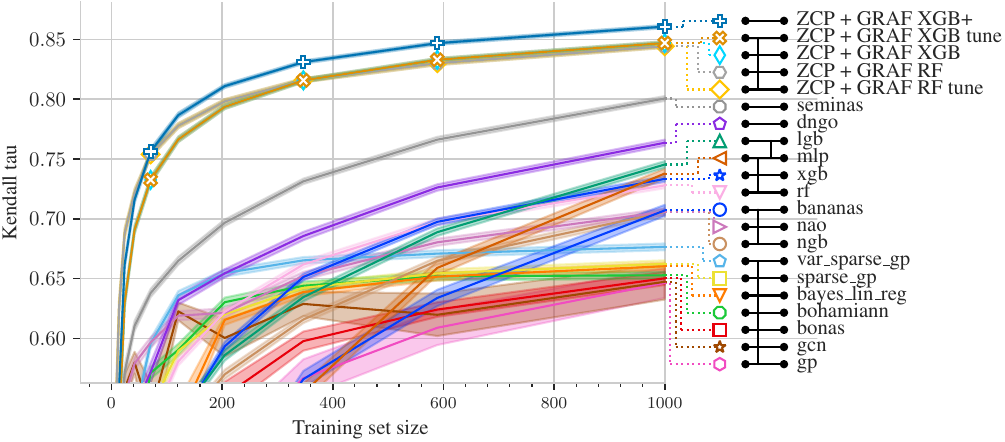}
\caption{NASLib predictor evaluation, NAS-Bench-201 and \texttt{cifar-10}.}
\label{naslib-nb201-eval}
\end{figure}

\begin{figure}[h]
\includegraphics{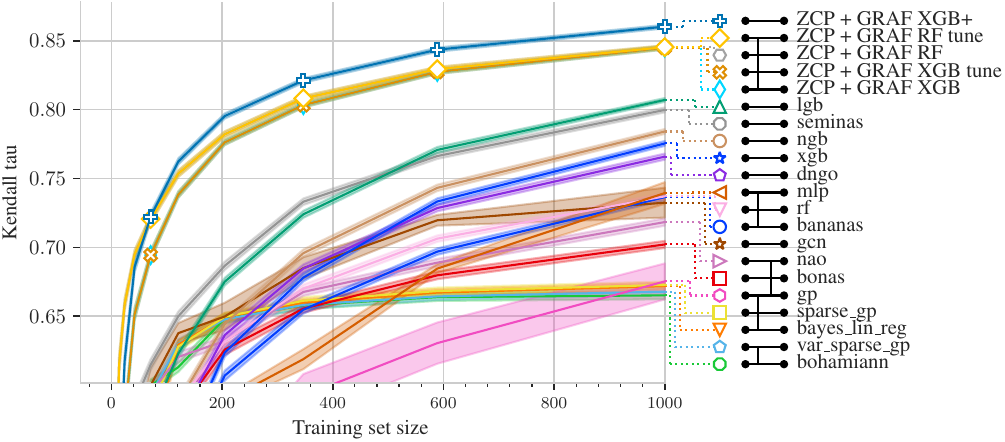}
\caption{NASLib predictor evaluation, NAS-Bench-201 and \texttt{ImageNet16-120}}
\end{figure}

\begin{figure}[h]
\includegraphics{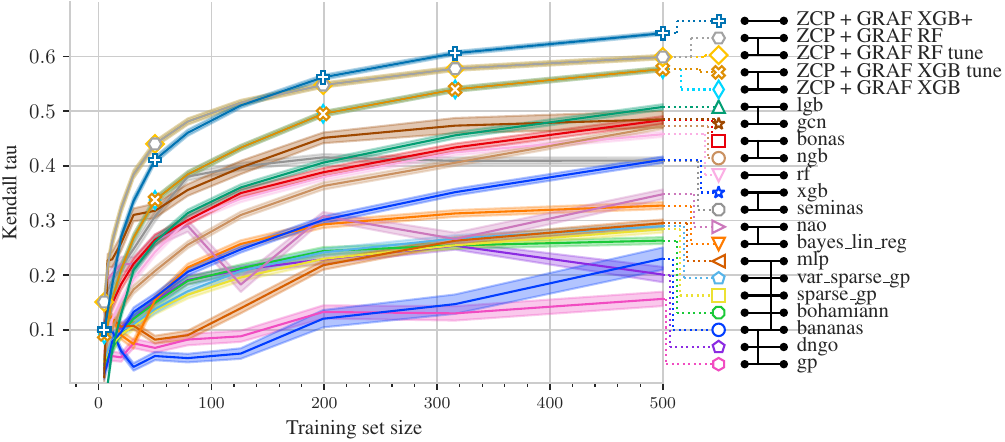}
\caption{NASLib predictor evaluation, NAS-Bench-301 and \texttt{cifar-10}.\label{fig:naslib301}}
\end{figure}

\FloatBarrier
\section{NAS Search Evaluation}
\label{app-naslib-search}

For NAS search experiments, we use the same Bayesian optimization + predictor framework \cite{nasbot, 8909403, bonas} as in the NASLib study \citep{naslib-predictors}. We compare the runs with a random search run.
In total, we used 21 CPU days for each of the 2 tasks, with 32GB RAM and 8 cores allocated per run, on Intel Xeon Gold 6242 CPU. No networks from the search space were trained, as we queried the accuracies.

In the BO + predictor framework, an ensemble of three performance predictors is used
for uncertainty estimates for each prediction. In each of the 25 iterations, 200 architectures are randomly sampled and out of these, 20 candidate architectures whose predictions maximize an acquisition function are evaluated. \citet{naslib-predictors} use independent Thompson sampling \citep{white2019bananas} as the acquisition function.

\begin{figure}[h]
\includegraphics[width=\textwidth]{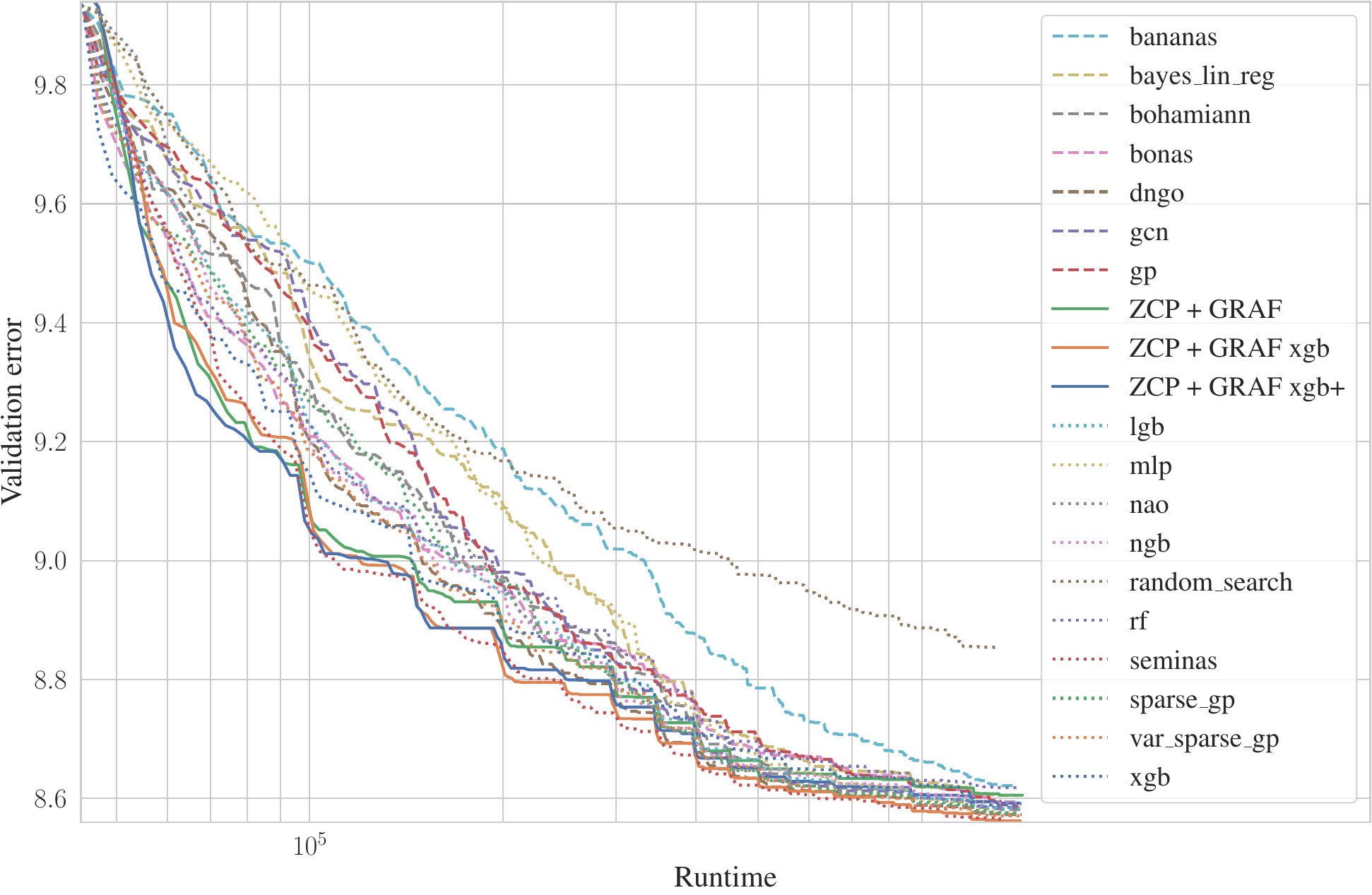}
\caption{NASLib search experiment, NAS-Bench-201 and \texttt{cifar-10}.\label{fig:cifarsearch}}
\end{figure}

\begin{figure}[h]
\includegraphics[width=\textwidth]{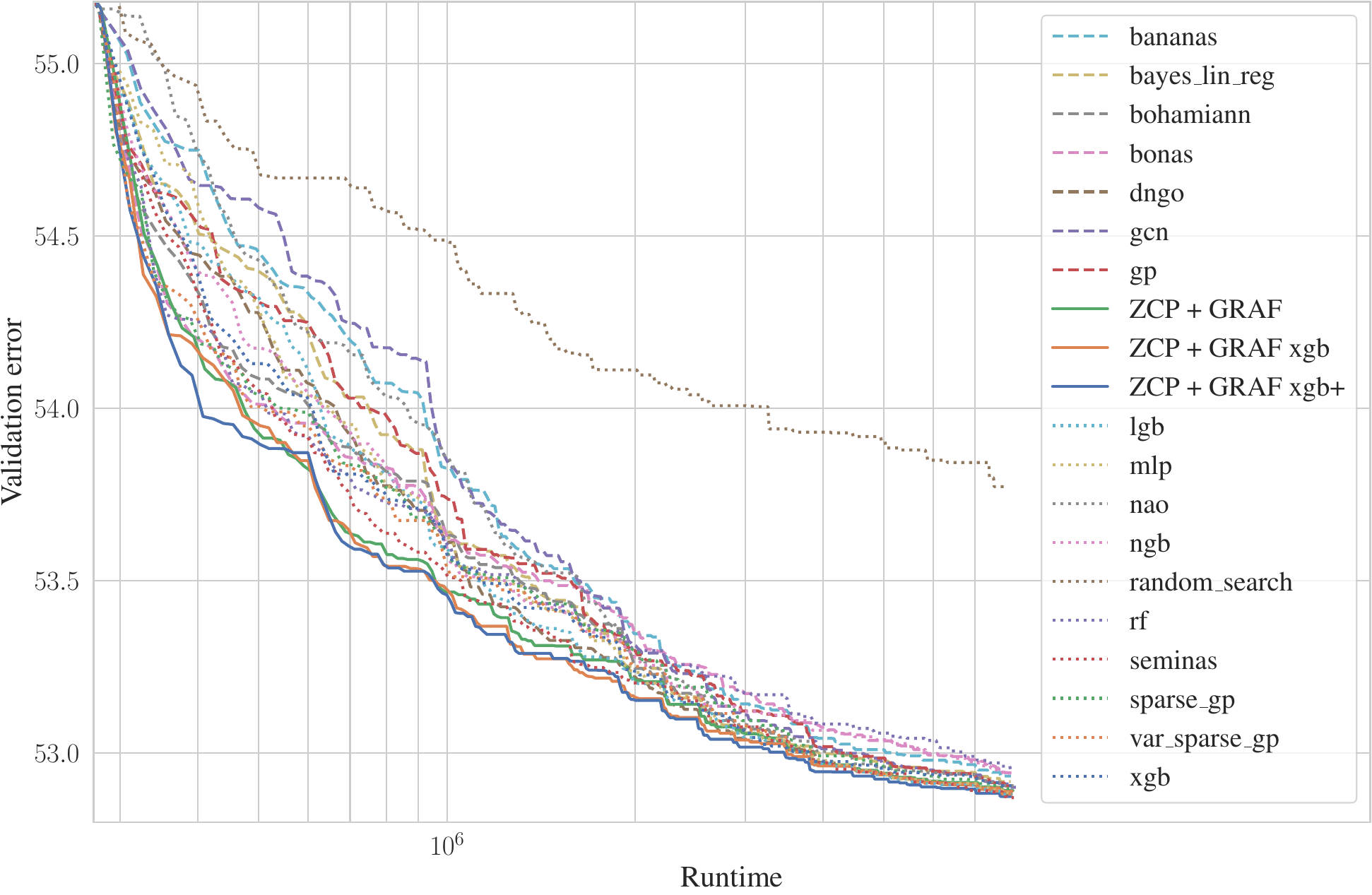}
\caption{NASLib search experiment, NAS-Bench-201 and \texttt{ImageNet16-120}.\label{fig:imgapp}}
\end{figure}

Figures \ref{fig:cifarsearch} and \ref{fig:imgapp} show results of the search on \texttt{cifar-10} and \texttt{ImageNet16-120} respectively.

\end{document}